\documentclass{article} 
\usepackage[T1]{fontenc}
\usepackage[final]{colm2026_conference}

\usepackage{microtype}
\usepackage{hyperref}
\usepackage{url}
\usepackage{booktabs}
\usepackage[most]{tcolorbox}
\usepackage[dvipsnames]{xcolor}
\usepackage{adjustbox}
\usepackage{enumitem}
\usepackage{multirow}
\usepackage{tabularx}
\usepackage{listings}
\usepackage{amsmath}
\usepackage{arydshln}
\usepackage{makecell}
\usepackage{subcaption}
\usepackage{CJKutf8}
\usepackage{fvextra}

\usepackage{lineno}

\definecolor{darkblue}{rgb}{0, 0, 0.5}
\hypersetup{colorlinks=true, citecolor=darkblue, linkcolor=darkblue, urlcolor=darkblue}

\DefineVerbatimEnvironment{exampleblock}{Verbatim}{
  breaklines=true,
  fontsize=\scriptsize
}

\DeclareCaptionType{prompt}[Prompt][List of Prompts]
\newcounter{example}

\title{Multilingual Agent-Based World Modeling for Social Science}

\author{Xuan Zhang$^{1}$ \quad Wenxuan Zhang$^{2}$ \quad Anxu Wang$^{3}$ \quad See-Kiong Ng$^{1}$ \quad Yang Deng$^{4}$\thanks{Corresponding author.} \\
$^{1}$Institute of Data Science, National University of Singapore \\
$^{2}$Singapore University of Technology and Design \\
$^{3}$National School of Development, Peking University \\
$^{4}$Singapore Management University \\
\texttt{xuanzhang@u.nus.edu}
}

\begin{document}

\ifcolmsubmission
\linenumbers
\fi

\maketitle
    
\begin{abstract}
Multi-agent role-playing has recently shown promise for studying social behavior with language agents, but existing simulations are mostly monolingual without cross-lingual interaction, an essential property of real societies. We introduce \textsc{MAWM}, the first Multilingual Agent-based World Modeling framework that supports multi-turn multilingual interactions among generative agents with diverse sociolinguistic profiles. \textsc{MAWM} enables two modes of analysis: (i) global public opinion modeling, which tracks how attitudes toward open-domain survey questions evolve across languages and cultures, and (ii) media influence and information diffusion, via autonomous news agents that dynamically generate content and shape user behavior. To ground the simulation in realistic population distributions, we construct the \textsc{MAPS} benchmark, which combines survey questions and demographic personas drawn from global population distributions. Experiments on simulation alignment, along with social science case studies on cultural assimilation and normative diffusion, show that native-language simulation better reflects real survey data than English-only simulation, and that the agent society in \textsc{MAWM} reproduces established sociocultural phenomena and highlights the value of multilingual simulation as an alternative interpretable tool for computational social science.
\end{abstract}

\section{Introduction}
Cross-cultural attitudes on global issues are shaped by multilingual discourse, yet how opinions form and shift across linguistic boundaries remains difficult to study with traditional surveys that capture only static snapshots \citep{APVS_2015, WVS_2017}. Simulating such dynamics with LLM-powered agents offers a complementary path, especially as the operating language of a model shapes its cultural cognitive style \citep{lu_cultural_2025}, making multilingual simulation essential for socially grounded modeling.

Role-playing enables language agents to adopt specific personas or roles during multi-turn interactions~\citep{chen_persona_2024}.
When scaling up to multiple role-playing agents, it becomes increasingly feasible to simulate aspects of human societies \citep{anthis_llm_2025}. A pioneering example is Smallville \citep{park_generative_2023}, which simulated a small town populated by 25 generative agents. This work was later extended in \citet{park_generative_2024}, where 1,052 digital representatives were modeled based on qualitative interviews. More recently, OpenClaw\footnote{\url{https://openclaw.ai}} lets users create digital agents of themselves to socialize on Moltbook\footnote{\url{https://www.moltbook.com}}. These initiatives rest on the assumption that \textit{if agents can take human memory and emulate human behavior, they may serve as practical proxies that complement real participants in social science research}.
More broadly, these efforts point toward agent-based world modeling of social interaction \citep{ding_understanding_2025}, computationally reproducing the dynamics of agent society that may mirror aspects of real human communities and their evolving social structures.

Building on this premise, tremendous efforts have been made in applying agent-based world modeling for diverse downstream societal studies: (1) uncovering social patterns and group behaviors \citep{zhu_multiagentbench_2025, DecodingEchoChambers-2025}, (2) interpreting social phenomena \citep{liu_skepticism_2024, tsirmpas_scalable_2025}, (3) exploring social theories \citep{dai_artificial_2024, ashery_emergent_2025}, and (4) forecasting policy outcomes \citep{li_econagent_2024, yan_opencity_2024}. These use cases offer a cost-effective complement to traditional social experiments, which typically depend on collecting human data through extensive surveys, interviews, and field studies.

Against this backdrop, two significant gaps can be identified in studying cross-cultural dynamics between the agent-based simulacra and real human communities.
(1) \textbf{Cross-lingual attitude dynamics remain unstudied in simulation}. Prior studies mainly focus on cross-cultural role-playing for data synthesis \citep{li_culturepark_2024, ki_multiple_2025}, but they neglect fine-grained agent behaviors that evolve across different languages, which constitute a pivotal component of human societies.
(2) \textbf{Limited sociolinguistic diversity of simulated environments}. Most simulations take place in monolingual or culturally homogeneous settings \citep{zhou_sotopia_2024, hou_can_2025}, restricting the emergence of diverse interaction patterns and overlooking social dynamics inherent to multilingual communities. As a result, multilingual social simulations remain largely underexplored, despite their significance for scalable, quantitative analysis of global social phenomena that are difficult to observe in traditional real-world studies or controlled survey experiments.

In light of these gaps, we propose the first \textbf{M}ultilingual \textbf{A}gent-based \textbf{W}orld \textbf{M}odeling (\textsc{MAWM}) framework, designed for societal world modeling of multi-turn interaction among generative agents with diverse sociolinguistic profiles.
Unlike prior monolingual simulations, \textsc{MAWM} allows agents to converse with, influence, and react to one another across languages.
\textsc{MAWM} supports (1) \textbf{global public opinion analysis} by simulating how attitudes toward open-domain survey topics evolve across languages and cultures. Starting from any survey question, the framework emulates user behavior on social platforms and aggregates users' responses via voting, offering a scalable and controllable alternative to global surveys \citep{APVS_2015, EVS_2017}. 
In addition, \textsc{MAWM} provides (2) \textbf{media influence and information diffusion analysis} by incorporating news agents that act as media organizations. Unlike agents in previous simulations that depend on predefined interventions \citep{li_econagent_2024, hou_can_2025}, these agents generate content conditioned on institutional profiles and evolving discourse, facilitating controlled studies of how media pierces, or reinforces, the filter bubbles that emerge in cross-cultural information networks.

To operationalize \textsc{MAWM}, we construct the \textbf{M}ultilingual \textbf{A}gent \textbf{P}erspective \textbf{S}urvey (\textsc{MAPS}), a benchmark combining survey questions from GlobalOpinionQA \citep{durmus_towards_2024} with demographic personas derived from the World Values Survey \citep{WVS_2017}, spanning 50 countries and 28 languages across geographically diverse regions.

Our evaluations, covering simulation alignment and case studies on cultural assimilation and normative diffusion, demonstrate that multilingual simulation with native-language agents estimates attitudinal patterns more closely aligned with real survey data than English-only simulation, consistent with evidence that language shapes cultural cognition in LLMs \citep{lu_cultural_2025}. Furthermore, \textsc{MAWM} can be used to reveal interpretable cross-cultural phenomena, including filter bubble formation, asymmetric media influence, and differential convergence rates that monolingual simulations cannot capture.

In summary, the primary contributions of this work are as follows:
\begin{itemize}[leftmargin=*, itemsep=0.2pt]
    \item As our primary contribution, we introduce \textsc{MAWM}, a multilingual agent-based world modeling framework that situates multi-agent society simulation within a global context. By enabling agents to communicate across languages and cultures, the framework addresses a fundamental blind spot in existing social simulations and provides a controlled environment for studying cross-cultural dynamics at a comprehensive global scale.
    \item We construct \textsc{MAPS}, an open-domain dataset that combines survey questions with user personas. This establishes a foundation for cross-cultural simulations and structured analysis of multilingual agent behavior across diverse cultural contexts.
    \item 
    Extensive experiments and case studies on cultural assimilation and normative diffusion not only validate \textsc{MAWM} as a new paradigm for multilingual society simulation, but also identify current constraints, and highlight its potential for computational social science.
\end{itemize}

\section{Related Work}

\textbf{Multi-Agent Simulation and Societal World Modeling} ~
LLM-based multi-agent systems show notable capabilities in handling complex problems \citep{zhang_ask-before-plan_2024} and sophisticated cooperation \citep{mandi_roco_2024}. 
Compared to a single agent, scaling multiple agents introduces more interactions and variables beyond isolated decision-making. This has led to world simulation as a natural application area, including society simulation \citep{park_generative_2023}, public administration \citep{xiao_simulating_2023}, and strategic games \citep{xu_language_2025}. Such frameworks extend the agent-based modeling tradition in the social sciences \citep{gao_large_2024}, creating virtual settings for examining social phenomena, such as the art of war \citep{hua_war_2024} and resource governance \citep{piatti_cooperate_2024}. However, most work assumes monolingual and culturally homogeneous environments, leaving simulations of multilingual societies largely unexplored. Even the closest attempt, CulturePark \citep{li_culturepark_2024}, focuses on data generation rather than analysis of social dynamics.
Existing surveys frame such simulation-based approaches as agent-based world modeling \citep{ding_understanding_2025}, where social simulacra complement real-world social computing systems through a world modeling process. \citet{socioverse} explicitly adopt this framing with a user pool of 10 million real-world individuals, while \citet{zhou_social_2025} formalize the representation of evolving social states and mental dynamics within such systems. Our work extends this direction to multilingual and cross-cultural social simulation settings.

\textbf{Cross-Cultural LLMs} ~ 
LLMs have made strong advancements in multilingual NLP, as shown by improvements in translation and cross-lingual question answering \citep{huang_survey_2025}. These developments build on recent multilingual LLMs \citep{xue_mt5_2021} and multilingual embedding models \citep{wang_multilingual_2024,enevoldsen_mmteb_2025}. Nonetheless, multilingual proficiency does not directly lead to cross-cultural understanding \citep{rystrom_multilingual_2025}, which requires models to handle diverse norms, values, and practices. In fact, prompting the same model in different languages yields measurably different cultural cognitive styles and social orientations \citep{lu_cultural_2025}, as language encodes culturally grounded reasoning patterns that monolingual simulation cannot activate. Cross-cultural LLMs therefore aim to represent, reason about, and adapt to cultural contexts, not just process different languages. Prior studies show measurable cultural bias toward English-centric value profiles in frontier models \citep{rao_normad_2025,myung_blend_2024}, introduce benchmarks for cultural awareness \citep{wang_cdeval_2024,chiu_culturalbench_2025}, and propose training or data-synthesis methods that improve cultural alignment \citep{nguyen_cultural_2024,li_culturellm_2024}. 
Despite these developments, cross-cultural LLM research remains largely single-agent, while little work integrates cultural modeling with multi-agent social simulations.

\section{\textsc{MAPS} Dataset}

In line with prior practices \citep{hou_can_2025, socioverse}, which pair real individuals and survey instruments for agent-based simulations, we form the Multilingual Agent Perspective Survey (\textsc{MAPS}) dataset by pairing (1) user settings sampled from multicultural populations with (2) social research questions of broad global concern.

\begin{figure*}[ht]
    \centering
    \setlength{\abovecaptionskip}{3pt}   
    \setlength{\belowcaptionskip}{0pt}
    \centering 
    \includegraphics[width=\textwidth]{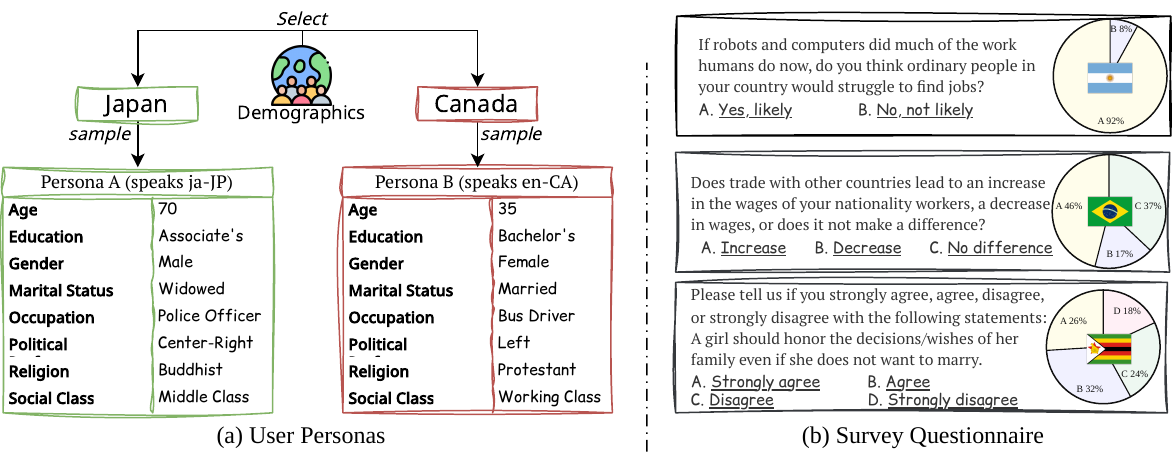}
    
    \caption{The \textsc{MAPS} dataset. (a) User personas derived from WVS, defined by eight socioeconomic attributes plus country and native language; examples from Japan and Canada. (b) Three survey questions with answer options and selected country-level response distributions, illustrated for Argentina, Brazil, and Zimbabwe as examples.}
    \label{fig:dataset}
    \vspace{-3mm}
\end{figure*}

To construct user settings across countries, we draw personas from the World Values Survey \citep{WVS_2017} given its breadth and scope. We retain eight independent attributes to characterize identity: \textit{age, education, gender, marital status, occupation, political preference, religion,} and \textit{social class}, then map these demographic variable values to human readable labels. For multilingual agent simulation, a native language is assigned to each interviewee based on country, yielding the sociolinguistic personas shown in Figure~\ref{fig:dataset}(a).

Social research questions of the \textsc{MAPS} dataset are sourced from GlobalOpinionQA \citep{durmus_towards_2024}, derived from WVS \citep{WVS_2017} and the Pew Global Attitudes Survey (GAS) \citep{pew_gas}. We keep only opinion questions whose country coverage overlaps with the user personas, and remove ambiguous or non-substantive options such as \textit{No answer}, \textit{Don't know}, and numeric placeholders. The final set includes the question, the answer options, and per-country option distributions, illustrated in Figure~\ref{fig:dataset}(b).

Full dataset statistics and annotation details are provided in Appendix~\ref{app:dataset}.

\section{The \textsc{MAWM} Framework}

\begin{figure*}[t]
    \centering
    \setlength{\abovecaptionskip}{3pt}   
    \setlength{\belowcaptionskip}{0pt}
    \centering 
    \includegraphics[width=\textwidth]{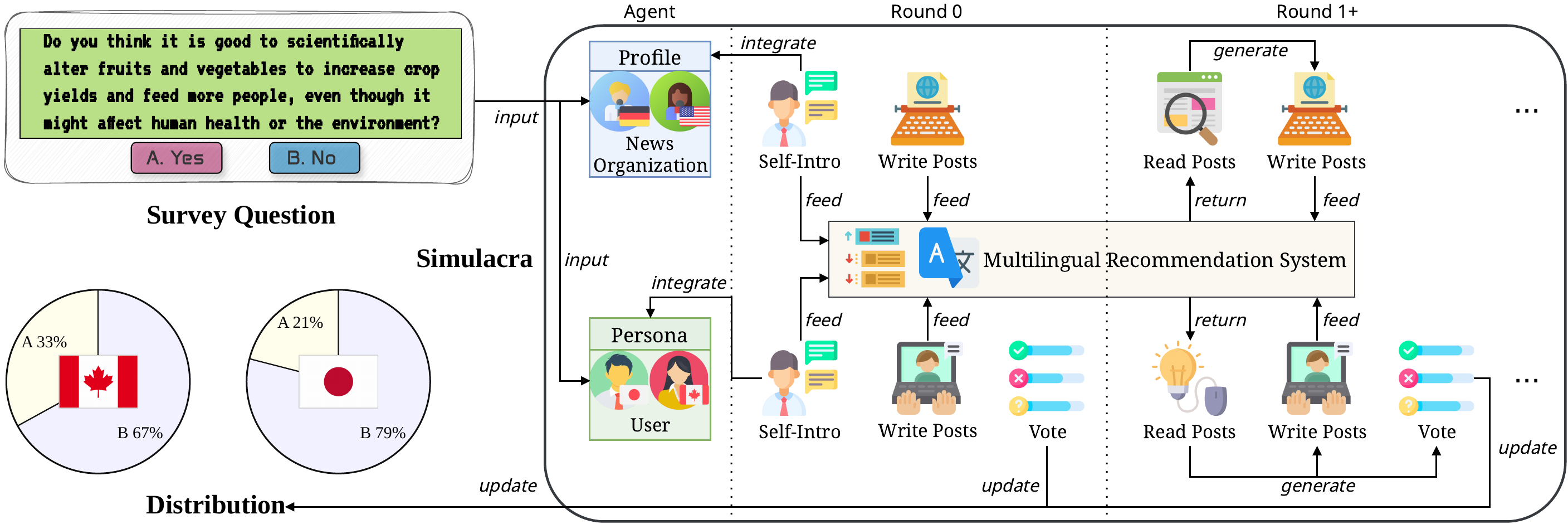}
    
    \caption{The \textsc{MAWM} framework. Starting from the survey question, in warm-up round $0$ news organization agents create self-introductions to augment their profiles and write posts stating their editorial stances, which are then fed into the recommendation system. User agents likewise create self-introductions, write posts to express their perspectives as input to the recommendation system, and participate in a multiple-choice vote. In rounds $t \geq 1$, news organization and user agents first read recommended posts, then write new posts for recommendation based on their memories and the content they read, and finally user agents vote to update the aggregated country-level attitude distribution.}
    \label{fig:framework}
    \vspace{-3mm}
\end{figure*}

\textsc{MAWM} simulates multi-turn social interactions among user and news organization agents across countries and tracks how attitudes evolve toward an opinion question, as illustrated in Figure~\ref{fig:framework}. In this system, both types of agents engage in iterative communication: news organizations express their stances on the survey question by generating news posts intended to influence users, while users read recommended content, compose their own posts, and vote to reflect their evolving attitudes toward the given survey question.
 
\subsection{Environment}
Let $(Q, C)$ denote the question-opinion pair as input, where $Q$ is the question and $C = \{c_i\}_{i=1}^{|C|}$ represents the set of possible response choices. The environment involves $N_u$ user agents with personas and $N_o$ news organization agents with profiles.
The multilingual recommendation system $R$ projects both agents and posts of different languages into a unified embedding space using a Sentence Transformer model~\citep{sentence-bert} to provide semantic post recommendations. Each agent's embedding is initialized from its self-introduction and updated each round via exponential moving average. Posts are ranked by a score combining semantic relevance with recency decay, and the top-$k_r$ posts are retrieved for each agent, excluding self-authored posts; at each round, posts are translated into the agent's language before delivery. Formal definitions are described in Appendix~\ref{app:implementation_details}.

The simulation is initialized at $t = 0$ and executed for $T$ rounds. Each agent is equipped with a persona or profile, short-term memory $\hat{M}$, long-term memory $\bar{M}$, and a set of authored posts $P$. At round $t$, the user $u_i$ produces an attitude distribution $d_{u_i}^{(t)} \in \Delta^{|C|-1}$. After the simulation completes, the framework returns $\mathcal{D} = \{ d_{u_i}^{(t)} \mid u_i = 1, \ldots, N_u; t = 0, \ldots, T \}.$

\subsection{Users}
User agents in the framework role-play social media users, each defined by a persona with characteristics such as \textit{age, education, country}, and \textit{language}. These personas allow agents to exhibit demographically grounded behaviors when interacting on the simulated platform.

Each round, the user follows a fixed sequence of actions. During the warm-up round ($t=0$), the agent generates a self-introduction, creates the first post, and casts a vote. This stage serves to initialize the agent's public identity, generate content for the recommendation system, and establish its baseline attitude toward the survey question $Q$.

In subsequent rounds ($t \ge 1$), the user first reads previous posts from the recommendation system $R$, forming long-term memories based on its understanding. Then the user agent composes a new post reflecting its experience and analysis before participating in the voting process to express its updated attitude toward the opinion question.

\textbf{Memory} ~ Inspired by agent cognitive architectures \citep{sumers_cognitive_2023}, we design short-term memory within each action and long-term memory across rounds for user agents.
\begin{itemize}[leftmargin=*, itemsep=0pt, topsep=0.5pt, parsep=0pt]
    \item \textbf{Short-term Memory}: Before each action, the user agent performs chain-of-thought reasoning \citep{cot} as part of its role-playing. Following Chain-of-Persona \citep{ji_enhancing_2025}, the agent completes three self-questioning steps before producing its final response. The resulting questions and answers serve as short-term memory.
    
    \item \textbf{Long-term Memory}: From each \textit{reading posts} action after round $0$, user agents extract weighted takeaways into long-term memory. The top-$k_m$ memories are retrieved by a score combining importance weight and temporal decay, and incorporated into subsequent action planning each round. Full definitions are provided in Appendix~\ref{app:implementation_details}.
\end{itemize}

\textbf{Actions} ~
Each user agent performs the following actions in the simulation.
\begin{itemize}[leftmargin=*, itemsep=0pt, topsep=0.5pt, parsep=0pt]
\item \textbf{Self-introduction}: At $t=0$, user agents produce self-introductions following SOTOPIA \citep{zhou_sotopia_2024}, which are incorporated into their personas.
\item \textbf{Read Posts}: For $t \ge 1$, the system $R$ provides up to $k_r$ translated posts, from which the agent extracts top-$k_m$ takeaways into long-term memory.
\item \textbf{Write Posts}: Conditioned on persona and memories, each agent creates a post $p_{u_i}^t$.

\item \textbf{Vote}: Following \citet{zhou_real_2024, hou_can_2025}, each agent outputs a probability distribution over survey options, regularized by temperature softmax (Appendix~\ref{app:implementation_details}).

\end{itemize}

\subsection{News Organizations}

News organization agents in the \textsc{MAWM} framework role-play media entities on social platforms, characterized by their \textit{editorial stance} toward $Q$ and their \textit{language}. Similar to user agents, news agents use short-term self-questioning and their recent post history as long-term memory. At $t=0$, each agent generates a self-introduction and an initial post. For $t \ge 1$, agents read recommended posts and write news in tandem, mimicking a newsroom workflow where editors review sources before producing content. To preserve asymmetry with user agents, news agents write directly from recommended content and post history rather than generating explicit takeaway memories from the content they consume.

\begin{table}[ht]
\centering
\small
\setlength{\abovecaptionskip}{3pt}   
\setlength{\belowcaptionskip}{0pt}
\begin{tabular}{p{0.947\columnwidth}}
\toprule
\textbf{Q201} (GAS): \textit{Does trade with other countries lead to an increase in the wages of your nationality’s workers, a decrease in wages, or does it not make a difference?} \\[2pt]
\makebox[4.2em][l]{\textbf{Options:}} A. Increase; B. Decrease; C. No difference. \\[1pt]
\makebox[4.2em][l]{\textbf{Cases:}} 1 – India, Japan, United States; 2 – South Korea, Brazil, Peru. \\[1pt]
\midrule
\textbf{Q278} (WVS): \textit{A girl should honor the decisions or wishes of her family even if she doesn't want to marry.} \\[2pt]
\makebox[4.2em][l]{\textbf{Options:}} A. Strongly agree; B. Agree; C. Disagree; D. Strongly disagree. \\[1pt]
\makebox[4.2em][l]{\textbf{Cases:}} 3 – Zimbabwe. \\
\bottomrule
\end{tabular}
\caption{Description of cases in the experiment.}
\label{tab:case_description}
\vspace{-3mm}
\end{table}

\section{Experiments}
\label{sec:experiments}

\subsection{Experimental Setups}
As shown in Table~\ref{tab:case_description}, we conduct experiments on three cases from the \textsc{MAPS} dataset, where the simulation initializes with 100 user agents, distributed across countries according to population proportions. The case design targets two globally consequential topics with strong cross-cultural variation. For Cases~1 and~2, we select Q201 on trade and domestic wages, a topic of heightened relevance given that new tariffs quadrupled in 2024--2025, covering 11.1\% of global imports \citep{wto_g20_2025}, and public attitudes toward trade vary sharply across countries \citep{pew_trade_2018}. Case~1 pairs India, Japan, and the United States to contrast Asian export-driven economies with an established Western market, while Case~2 pairs South Korea, Brazil, and Peru to compare one Asian and two South American economies with differing trade openness. For Case~3, we select Q278 on family-decided marriage, a gender equality issue, by pairing Zimbabwe and the Netherlands, which represent opposite ends of the Gender Inequality Index, to study normative diffusion. \citet{unicef_child_marriage_2023} estimates that, at the current pace, child marriage will take 300 years to end.

Each simulation runs for 20 rounds with an additional warm-up round. We test five LLMs as agent backbones: \texttt{GPT-5-mini}, \texttt{GPT-4o-mini}, \texttt{Gemini-2.5-Flash}, \texttt{Llama-4-Maverick}, and \texttt{Llama-4-Scout}. The multilingual recommendation system is implemented by \texttt{jina-embeddings-v3} \citep{sturua_jina-embeddings-v3_2024} and \textit{Google Translate}. To reflect real-world social platforms, we ensure that each user is exposed to at least one news item in their timeline.
Implementation details and case statistics are detailed in Appendix~\ref{app:implementation_details} and Appendix~\ref{app:case_description}.

\subsection{Reliability Analysis of \textsc{MAWM}}
\label{subsec:reliability}
We evaluate the reliability of \textsc{MAWM} from three perspectives: (1) real-world calibration, which examines whether the simulation reflects social dynamics observed in the real world; (2) global sensitivity, which tests whether the simulated society responds appropriately to external signals; and (3) local consistency, which assesses whether individual agent behaviors align with their demographic settings, cultural backgrounds, and social characteristics. Limitations of the current calibration and evaluation are discussed in Appendix~\ref{app:limitations}, and the complementary robustness analyses are further reported in Appendix~\ref{app:robustness}.

\paragraph{Real-World Calibration}
As a practical check of initial alignment, we run each case without news or cross-country communication and measure Root Mean Square Error (RMSE) between simulated and real-world survey distributions across countries and rounds, where RMSE ranges from 0 (perfect match) to $\sqrt{2/|C|}$ (maximum divergence); see Appendix~\ref{app:real_world_calibration} for the complete derivation of both country-level and overall RMSE metrics.

\begin{figure*}[t]
    \centering
    \setlength{\abovecaptionskip}{1pt}
    \setlength{\belowcaptionskip}{0pt}
    \centering 
    \includegraphics[width=\textwidth]{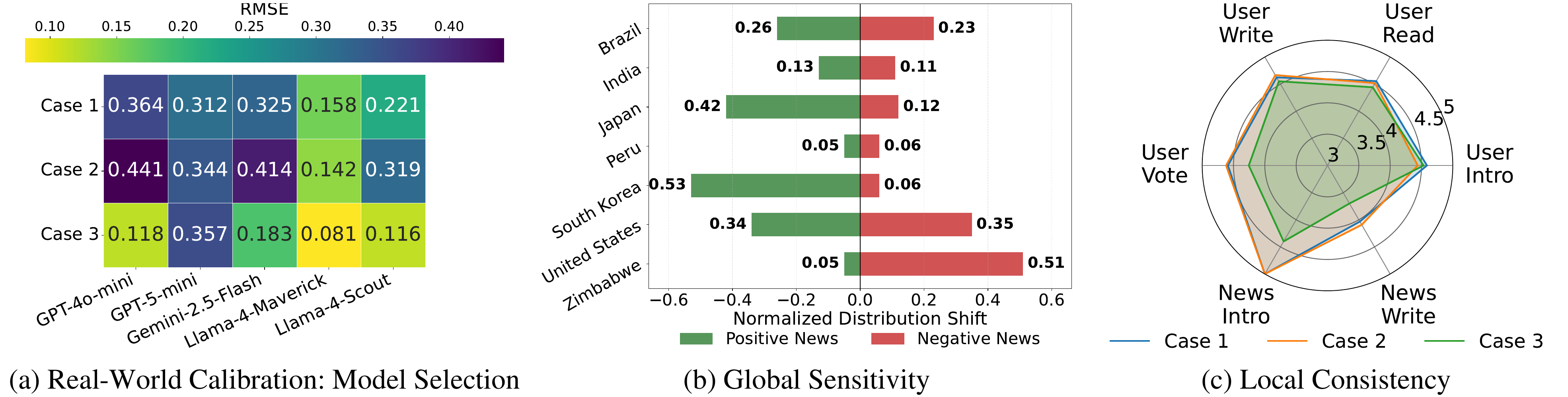}

    \caption{Experimental results. (a) Lower RMSE scores correspond to better real-world calibration. (b) Baseline-normalized distribution shifts under positive and negative news, measuring the fraction of movable opinion captured by media exposure. (c) Higher consistency scores represent better quality agent actions across all three cases.}
    \label{fig:experimental_results}
    \vspace{-1mm}
\end{figure*}

\begin{figure*}[t]
    \centering
    \setlength{\abovecaptionskip}{1pt}
    \setlength{\belowcaptionskip}{0pt}
    \includegraphics[width=\textwidth]{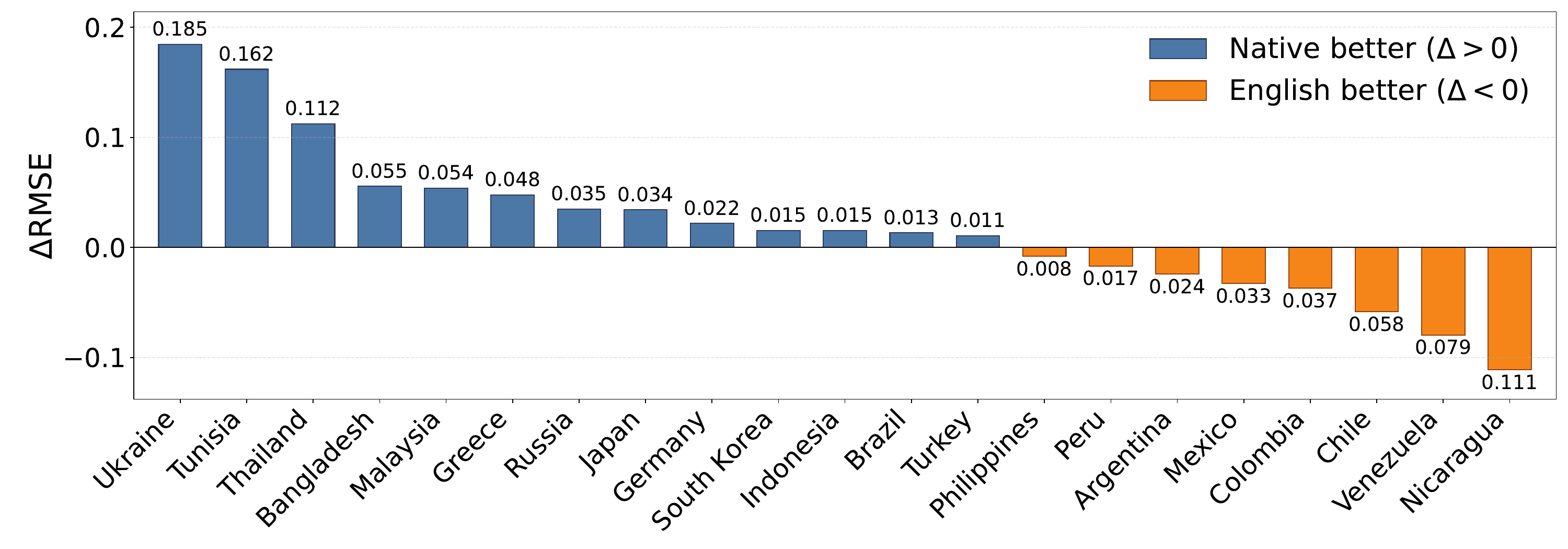}

    \caption{Real-world calibration difference $\Delta\text{RMSE} = \text{RMSE}_{\text{English}} - \text{RMSE}_{\text{native}}$ for the 21 non-English countries in \textsc{MAPS}; positive values favor native-language simulation (Japan, South Korea, Brazil, and Peru use \texttt{Llama-4-Maverick}; the rest use \texttt{GPT-4o-mini}). Native-language simulation is better calibrated in 13 of 21 countries with the largest gains in lower-resource, largely non-Latin-script languages, whereas English is favored only by small margins, almost exclusively in high-resource Spanish-speaking countries.}
    \label{fig:language_rmse}
    \vspace{-2mm}
\end{figure*}

Figure~\ref{fig:experimental_results}(a) shows that \textbf{language models can partly approximate global survey distributions, with noticeable discrepancies between models}. Specifically, \texttt{Llama-4-Maverick} and \texttt{Llama-4-Scout} perform uniformly better than others, which may be attributed to their pretraining preserving preferential rather than strictly unbiased survey question responses. Based on this observation, we choose the best-performer, \texttt{Llama-4-Maverick}, for the subsequent experiments and case studies presented below. Comparing English vs.\ native-language simulations (agents reading and writing posts in native languages) across all 21 non-English countries in \textsc{MAPS} (Figure~\ref{fig:language_rmse}), we find that \textbf{native-language simulation yields better calibration (lower RMSE) in most countries, with the largest gains concentrated in lower-resource, largely non-Latin-script languages}.

Two findings support the necessity of multilingual modeling. First, native-language simulation matches the real distributions more closely in 13 of the 21 countries, with the largest gains (up to $0.18$ RMSE) in lower-resource, largely non-Latin-script languages such as Ukrainian, Thai, Arabic, and Bengali: these languages and cultures are underrepresented in pre-training data \citep{joshi_state_2020, wang_languages_2024, nezhad_beyond_2025}, and LLM-expressed values are language-conditioned, with native-language prompting activating culture-specific perspectives that English suppresses \citep{arora_probing_2023, tao_cultural_2024, dai_word_2025}. Second, where English wins, the margin is small ($|\Delta| \leq 0.11$) and confined almost entirely to Spanish-speaking countries; as a top-tier, Western-aligned language with abundant English parallel data \citep{schwenk_ccmatrix_2021}, Spanish behaves much like English, so these small gaps are a weak test; the substantive signal lies in the large, consistent gains observed for lower-resource, non-Western languages and their cultures.

\paragraph{Global Sensitivity}
To analyze global sensitivity, we examine how news exposure affects agents across three cases. For each country, we introduce a native-language news organization agent adopting either a positive (\emph{Increase} in Q201; \emph{Strongly agree} in Q278) or negative (\emph{Decrease} in Q201; \emph{Strongly disagree} in Q278) stance, with no cross-country communication allowed. The positive-news (negative-news) effect is the increase, relative to a no-news baseline, in the share selecting the promoted (opposing) option, averaged over the final three rounds. Because media effects are bounded by the available headroom (support cannot rise far above an already-favorable baseline), we report a baseline-normalized sensitivity: $\hat{\Delta}_o = \Delta_o / (1 - \tilde{p}_o)$, where $\Delta_o$ is the raw shift toward the targeted option $o$ and $\tilde{p}_o$ is its baseline share. Figure~\ref{fig:experimental_results}(b) shows the normalized shifts for the positive (left) and negative (right) stances. Derivations and further interpretation are provided in Appendix~\ref{app:global_sensitivity}, and the results demonstrate that \textbf{user agents in the simulation respond to injected news by shifting their attitudes toward the designated editorial stance} in both the positive and the negative news exposure conditions alike across all seven studied countries.

\paragraph{Local Consistency}
We also assess whether individual agent behaviors are internally coherent with their assigned personas. We use \texttt{GPT-5} as an LLM judge to rate agent quality from 1 (poor) to 5 (excellent) over all actions. We evaluate the Global Sensitivity setting with positive news, selecting 15 user agents per case plus all news agents, and scoring their responses for each action over 5 rounds. As shown in Figure~\ref{fig:experimental_results}(c), \textbf{agents generally achieve high response quality except for news creation}; Appendix~\ref{app:local_consistency} shows that news agents tend to repeat their previous posts, reducing content diversity over rounds. 

\subsection{Case Study on Social Science}
\label{sec:case_study}
To showcase how \textsc{MAWM} can facilitate social science research, we conduct cross-cultural studies as a direct application of the framework. Multicultural discourse is substantively important yet challenging given the scarcity of suitable multinational data; \textsc{MAWM} helps overcome this limitation.
We aim to address two research questions in social science studies: 1) \textit{Does international cross-lingual communication help users from different countries converge in their opinions on a given opinion question?} (Cultural Assimilation) 2) \textit{Following international interaction, can users or news organizations from source countries promote norms in target countries, and which actor is more influential?} (Normative Diffusion) Complete results and theoretical interpretations of these dynamics across all three cases are provided in Appendix~\ref{app:case_study}.

\begin{figure*}[t]
    \centering
    \setlength{\abovecaptionskip}{1pt}
    \setlength{\belowcaptionskip}{0pt}
    \centering 
    \includegraphics[width=\textwidth]{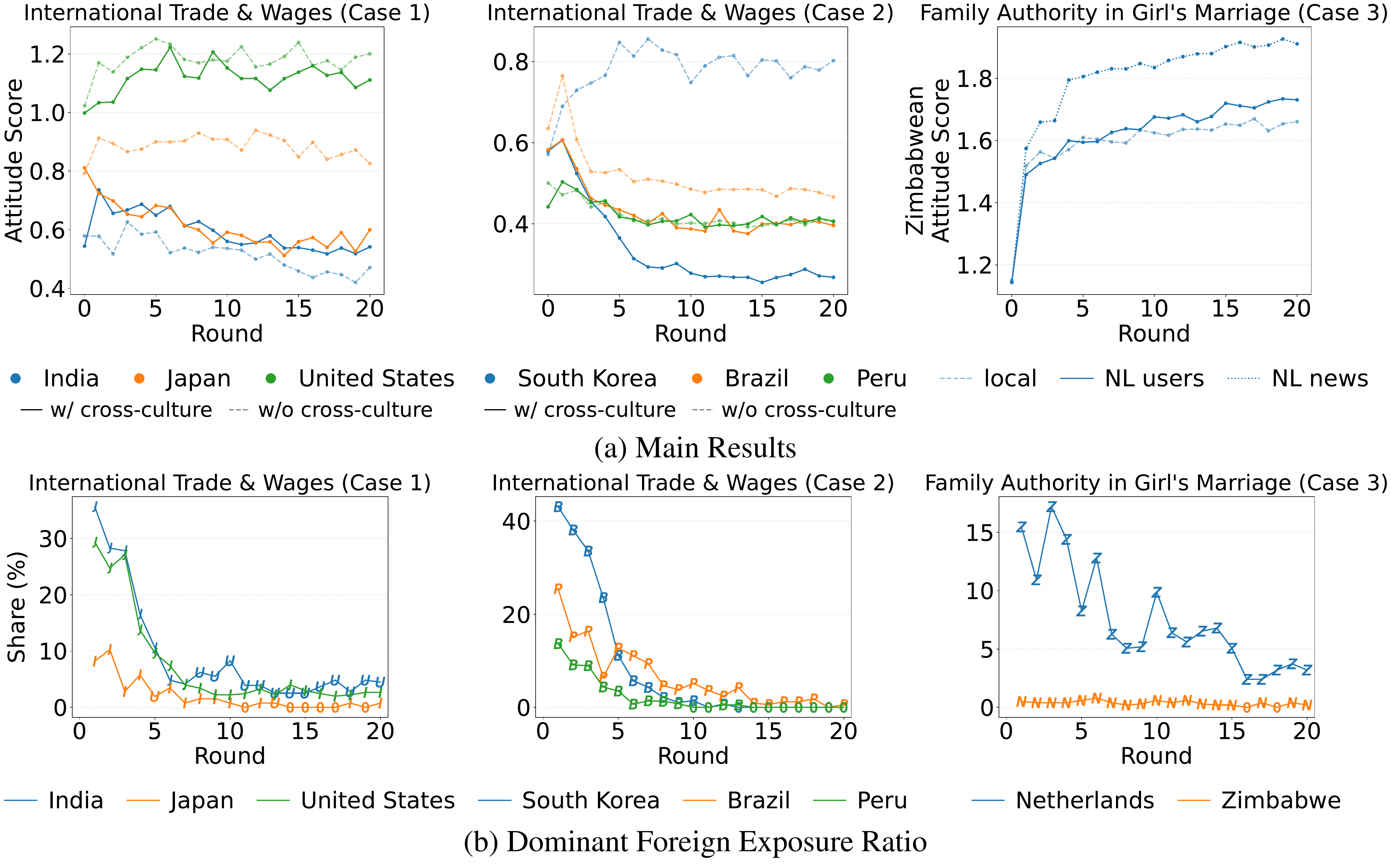}
    
    \caption{Case study. (a) Cultural assimilation attitude changes for Cases~1 and~2 and normative diffusion attitude shifts for Case~3; semi-transparent dashed lines are simulations without cross-cultural communication. (b) The maximum share of user recommendations from any single foreign country (Dominant Foreign Exposure Ratio); markers denote the largest foreign country by initial letter, and zero indicates no foreign posts are recommended.}
    \label{fig:case_study}
    \vspace{-2mm}
\end{figure*}

For each case, we run three trials and report in Figure~\ref{fig:case_study}(a) the average attitude score, where the attitude distribution $d_{u_i} \in \Delta^{|C|-1}$ is converted to a scalar score by
$
s_{\text{att}} = \sum_{c=0}^{|C|-1} c \cdot d_{u_i}^c,
$
with the $|C|$ choices reranked from most positive ($c = 0$) to most negative. Figure~\ref{fig:case_study}(b) reports the \textit{dominant foreign exposure ratio} for each country, defined as the highest proportion of recommended content for users that originates from any foreign country; across-run standard deviations of these trajectories are small relative to their round-to-round movement, as quantitatively reported in Table~\ref{tab:attitude_std} of the robustness analysis in Appendix~\ref{app:robustness}.

\subsubsection{Cultural Assimilation}
In sociology, \textit{cultural assimilation} \citep{culture_assimilation} refers to the process by which distinct groups come to share a common culture. In our setting, we conceptualize assimilation as cross-national convergence in agent attitudes on specific survey questions. Below, we analyze cultural assimilation dynamics for the question on \textit{international trade and domestic wages}, focusing on two country groups. Both cases pair two geographically proximate countries with one from a different region: Case~1 groups India and Japan (Asia) with the United States, while Case~2 groups Brazil and Peru (South America) with South Korea. This design allows us to examine how geographic and cultural proximity interacts with cross-regional communication in shaping the observed attitudinal convergence.

\noindent\textbf{Case 1: India, Japan, and the United States} 
\begin{itemize}[leftmargin=*,nosep]
    \item Simulated U.S.\ users exhibit comparatively stable attitudes, with scores ranging only from 1.0 to 1.2 (between \emph{No difference} and \emph{Decrease}) during communication. This indicates \textbf{minimal movement toward pro-trade positions}, even when exposed to Japanese content in trade discussions. This stability is notable given that LLM agents tend toward consensus in multi-turn interactions \citep{taubenfeld_systematic_2024}; here, multilingualism activates cultural orientations sustaining heterogeneous attitudes across countries.
    \item India and Japan, by contrast, have strategic incentives for openness to trade \citep{india-free_trade, japan-free_trade}. \textbf{Japan's simulated attitudes appear to converge sharply toward India's}: scores stabilize around 0.5-0.6 (between \emph{Increase} and \emph{No difference}). Japanese users receive very few foreign recommendations; by round 10, their foreign exposure approaches zero, implying the early formation of an information cluster that further limits their exposure to the broader ongoing cross-national discourse.
\end{itemize}



\noindent\textbf{Case 2: South Korea, Brazil, and Peru} 
\begin{itemize}[leftmargin=*,nosep]
    \item South Korea begins with a relatively high attitude score ($\sim$0.8), but after global communication, its score falls to $\sim$0.3, lower than those of Brazil and Peru ($\sim$0.4). This indicates that \textbf{South Korea ends up more supportive of free trade in the simulation than before.} Recommendation-flow analysis (Figure~\ref{fig:case_2_study}, Appendix~\ref{app:case_study}) reveals that Brazilian and Peruvian content, expressing similar pro-trade positions, accounts for 73.3\% of South Korea’s foreign recommendations in round 1, declining to 3.5\% by round 8. This early surge in exposure likely triggers the opinion shift, suggesting that \textbf{initial dominance of recommendations from countries with similar attitudes can strongly steer opinion convergence} in multilingual agent-based societies such as \textsc{MAWM}.
    \item Both South American countries (Peru and Brazil) exhibit \textbf{rapid convergence} within five communication rounds. Each becomes the other’s largest foreign recommendation source, consistent with \textbf{mutual influence driving assimilation}. The same filter bubble as in Japan appears in Peru, whose foreign content exposure similarly declines over rounds. 
\end{itemize}

Japan and South Korea thus reflect \textbf{two distinct mechanisms rather than one implausibly fast dynamic}. Japan is not genuinely assimilated: with the highest inbreeding-homophily index (Appendix~\ref{app:case_study}), its users are isolated by the recommender from the first round, its with- and without-communication trajectories nearly coincide (Figure~\ref{fig:case_study}(a)), and its already pro-trade initial stance makes a small self-contained drift resemble convergence, the classic recommender echo-chamber effect \citep{pariser_filter_2011} rooted in homophily \citep{mcpherson_birds_2001}. South Korea instead converges through genuine exposure to similar-stance Brazilian and Peruvian content, and such rapid shifts are realistic: the 2016--2017 Korean presidential scandal saw approval collapse to 5\% within weeks \citep{koreaherald_park_2016}. Appendix~\ref{app:culture_language} further disentangles culture from language via a controlled composition experiment across the three Spanish-speaking populations of Argentina, Bolivia, and Peru.

 

\subsubsection{Normative Diffusion}
Beyond cultural assimilation, cross-cultural communication can also produce \textit{normative diffusion} \citep{international_norm}, the directional transfer of a behavioral standard from a source population to a target population, reshaping the target’s beliefs, values, or policy preferences through sustained social interaction and repeated exposure.

Case~3 turns to the diffusion of \textit{gender equality norms} \citep{gender_norms_diffusion}. We designate \textbf{Zimbabwe}, ranked 153rd in the Gender Inequality Index \citep{undp-hdr-2025}, as the target country. Without interventions, Zimbabwean agents’ attitude scores range from about 1.5 to 1.7 (between \emph{Agree} and \emph{Disagree} on girls honoring family marriage decisions). We then introduce the source nation, the \textbf{Netherlands} (ranked 8th in the Gender Inequality Index), and compare two interventions: 1) five Dutch news organization agents supporting option~D (\emph{Strongly disagree}), and 2) 50 Dutch user agents expressing the same stance while directly interacting alongside the 100 Zimbabwean user participants.

The results in Figure~\ref{fig:case_study} indicate that \textbf{news organizations drive stronger normative shifts than ordinary users}. Dutch user posts make up less than 2\% of Zimbabwean agents’ recommended content, showing a strong filter bubble restricting cross-country user exposure. In contrast, news organizations have a larger impact, as the recommendation system emulates real platforms by including at least one news post per round. Consequently, Zimbabwean agents are exposed to Dutch news content and more likely to shift their attitudes in a pro-equality direction, highlighting the media’s role in piercing filter bubbles. Notably, as shown in Figure~\ref{fig:case_study_nl_score} in Appendix~\ref{app:case_study}, \textbf{Dutch users’ attitudes remain consistent throughout the simulation} despite sustained exposure to 100 Zimbabwean users with opposing views, confirming that the normative diffusion is asymmetric rather than mutual between the Dutch source population and the Zimbabwean target user population.

Beyond static survey snapshots, two pieces of external evidence indicate that the simulated trajectories reflect real-world change. First, real attitudes shift over time in the direction our simulation moves them: across World Values Survey waves in Zimbabwe, agreement that ``men make better political leaders than women'' went from a 52\% majority (2001), via 58\% (2012), to a 45\% minority (2020), and disagreement that ``a university education is more important for a boy than a girl'' rose from 81\% to 86\%. Second, such change is driven by information exposure, the mechanism our framework models \citep{jensen_power_2009, laferrara_soap_2012}: internet penetration, treated as a quasi-natural exposure experiment, is positively associated with gender equality across more than 50 countries \citep{worldbank_wdi_2024}, significant after controlling for GDP per capita, urbanization, enrollment, and region ($\beta = 0.0014$, $p<0.01$). As no survey tracks these questions over time, we claim only the direction and the driver of the simulated opinion change, not its exact rate of change.

\section{Conclusion}
In this work, we explore multilingual world modeling for social science, beginning with the \textsc{MAPS} dataset, which integrates open-domain survey questionnaires with user demographics from global surveys. On top of this resource, we develop \textsc{MAWM}, which models social-network interactions among cross-country user and news agents and outputs attitude dynamics for survey questions. Our evaluation examines real-world calibration, global sensitivity, and local consistency, evidencing its utility and internal consistency. We further apply \textsc{MAWM} to three \textsc{MAPS} case studies, uncovering interpretable empirical findings. These results underscore \textsc{MAWM}'s potential as a practical, complementary tool for computational social science research across diverse linguistic and cultural settings.



\section*{Reproducibility Statement}
The \textsc{MAPS} dataset is constructed from publicly available sources: user personas from the World Values Survey Wave~7 \citep{WVS_2017} and opinion questions from GlobalOpinionQA \citep{durmus_towards_2024}, both available for non-commercial research use. The five LLMs and recommendation system configurations are described in Section~\ref{sec:experiments}. Dataset construction details and persona attribute definitions are provided in Appendix~\ref{app:dataset}. All hyperparameters and implementation details of \textsc{MAWM}, including the memory scoring function, attitude regularization, embedding update, and similarity ranking, are reported in Appendix~\ref{app:implementation_details}. LLM-based evaluation prompts are listed in Appendix~\ref{app:prompts}. Since \textsc{MAWM} is a general framework, prompt templates and hyperparameters may need to be adapted for specific research scenarios; practitioners should carefully calibrate these choices for their use case. The code and data are available at \url{https://github.com/magicgh/MAWM}.

A related consideration is data contamination: the calibration targets derive from public survey resources that may appear in LLM pre-training data. Contamination would require the model to recall survey statistics rather than simulate populations, and three observations argue against this interpretation. First, our primary contribution is the dynamic interaction of multilingual agents, including round-by-round trajectories and cross-cultural assimilation, which are absent from survey data and therefore cannot be memorized. Second, the aggregate endpoint distribution is never shown to the model: opinions are generated independently for 100 persona agents and aggregated externally, making direct recall of survey percentages implausible, and the observed agent-level variance further supports genuine individual-level generation. Third, the model exhibits systematic responses to interventions not contained in survey data, including media-induced opinion shifts and assimilation under communication, which simple retrieval of memorized statistics would not reproduce. Even the static endpoint behaves inconsistently with recall: alignment RMSE varies markedly across models and between native and English prompting of the same population, which retrieval of a single public statistic could not produce.

\section*{Ethics Statement}
This work provides an alternative tool for social science research through agent-based world modeling; no new human-subject data are collected, and all persona data are derived from anonymized, publicly available surveys. The personas are role-played by large language models, and their attitudes and behaviors reflect only simulated responses under specified conditions; they should not be interpreted as empirical claims about real populations or used to draw normative conclusions about underrepresented groups. Accordingly, any applied or policy-relevant interpretation should be triangulated with empirical human-subject research. We recognize the dual-use potential of LLM-based agents and emphasize transparency, methodological clarity, and responsible collaboration with domain experts.

\section*{Acknowledgment}
This project is supported by the National Research Foundation Singapore under the AI Singapore Programme (AISG Award No: AISG3-RPGV-2025-016). Yang Deng is supported by the Lee Kong Chian Fellowship awarded by Singapore Management University.

\bibliography{references}
\bibliographystyle{colm2026_conference}

\appendix

\section{Limitations}
\label{app:limitations}
\paragraph{User Simulation} In \textsc{MAPS} and \textsc{MAWM}, we emulate users with ten persona attributes and a self-introduction through in-context learning and chain-of-thought reasoning, which are inherently approximate simulacra of real-world users rather than faithful representations. In-context prompting is a common practice in LLM-based agent simulation and is efficient to scale up to millions of agents, since in social science research the persona configuration changes with each study; alternative approaches such as fine-tuning or reinforcement learning on user trajectory data are not cost-effective at this scale. Prior work suggests that demographic attributes alone have limited power in aligning LLM role-playing with real human opinions \citep{ santurkar_whose_2023, chuang_beyond_2024}; incorporating belief-level information may improve persona fidelity, which we leave to future work. Additionally, persona-to-opinion mapping operates at the country level rather than the individual level, which may obscure within-country variation in attitudes. Since the aim of this paper is to demonstrate multilingual societal world modeling, our simulations use 100 agents due to computing resource limitations; scaling to larger populations is left to future work.
\paragraph{Social Networks} The primary goal of this work is to demonstrate multilingual simulacra as societal world modeling and present its potential applications for social science. To this end, we provide a simplified multilingual social network that preserves fundamental platform features. This abstraction is a deliberate methodological choice in line with both major traditions of social modeling: agent-based modeling \citep{schelling_dynamic_1971, axelrod_evolution_1984} follows the KISS principle \citep{axelrod_advancing_1997} and the generative program \citep{epstein_generative_2006}, placing complexity in the results rather than the assumptions so that mechanisms can be cleanly identified, while opinion-dynamics and network-formation models abstract communication even more aggressively, reducing persuasion to a weighted average of scalar opinions \citep{degroot_reaching_1974}, bounded confidence to a similarity threshold \citep{hegselmann_opinion_2002}, and strategic network formation to a cost--benefit edge \citep{jackson_strategic_1996}. Against this baseline, \textsc{MAWM} enriches rather than over-simplifies communication: interaction is grounded in real multilingual text whose influence weights are generated endogenously by the recommendation system, mirroring deployed production systems such as X's Grok-based recommender, which translates cross-language posts and ranks them by predicted user interactions \citep{xai_xalgorithm_2026}. Nevertheless, real communication involves imperfect translation, and news media may selectively report or slant content; our claims concern only the exposure structure, namely whether cross-cultural communication promotes convergence and the broad reach of news relative to dispersed peer posts, not the naturalness of translation or the veracity of content. The news agent's editorial stance is a controlled experimental variable, and modeling media bias, selective reporting, and more realistic social networks is left to future work beyond the present scope of the framework.
\paragraph{Calibration and Evaluation} Our calibration compares simulated distributions against static survey snapshots rather than longitudinal human dynamics. For open-domain questions, such dynamic ground-truth data is scarce and very hard to collect without large-scale, year-long studies. Moreover, LLM pretraining corpora may already encode outcomes of past events, making retrospective trend evaluation with historical data unreliable; the only potential dynamic validation would be predicting future events whose outcomes are not yet in the training data. We therefore position the framework as a complementary tool rather than a predictive model of human social dynamics. Besides, the convergence speed observed in our case studies may be faster than in real-world dynamics, as prior work has shown that LLM agents tend toward consensus more rapidly than humans \citep{taubenfeld_systematic_2024}. Since \textsc{MAWM} is presented as a general framework with illustrative applications, prompt templates and hyperparameters may need to be adapted for specific research scenarios; practitioners applying the framework to new domains should carefully calibrate these choices for their particular use case. Our local consistency evaluation relies on \texttt{GPT-5} as an LLM judge, which is now common practice in the evaluation of generative agents \citep{fu2024gptscore}, though it may introduce systematic biases that we have not yet quantified.

\section{Dataset}
\label{app:dataset}
\begin{table}[ht]
  \setlength{\abovecaptionskip}{3pt}   
  \setlength{\belowcaptionskip}{0pt}
  \centering
  \small
  \begin{tabular}{@{}l r@{}}
    \toprule
    \multicolumn{2}{@{}l}{\textbf{User Persona}} \\
    \midrule
    \quad \# Participants & 57,647 \\
    \quad \# Languages & 28 \\
    \quad Average Participant per Country & 1,152.94 \\
    \midrule
    \multicolumn{2}{@{}l}{\textbf{Survey Questionnaire}} \\
    \midrule
    \quad \# Question & 300 \\
    \quad Average Question Length & 191.54 characters \\
    \quad \# Country & 50 \\
    \quad Average Country per Question & 7.47 \\
    \quad Average Option per Question & 3.37 \\
    \quad Average Option Length & 14.61 characters \\
    \quad Sources & \\
    \qquad - GAS & 263 \\
    \qquad - WVS & 37 \\
    \bottomrule
  \end{tabular}
  \caption{Dataset statistics.}
  \label{tab:dataset_statistics}
\end{table}

\begin{table*}[ht]
  \centering
  \scriptsize
  \setlength{\abovecaptionskip}{1pt}   
  \setlength{\belowcaptionskip}{0pt}
  \begin{tabular}{l p{0.7\textwidth}}
  \toprule
    \textbf{Occupational Group} & \textbf{Occupations} \\ \midrule
    Professional and Technical & Doctor, Teacher, Engineer, Artist, Accountant, Nurse \\[0.1cm]
    Higher Administrative & Banker, Executive in big business, High Government Official, Union Official \\[0.1cm]
    Clerical & Secretary, Clerk, Office Manager, Civil Servant, Bookkeeper \\[0.1cm]
    Sales & Sales Manager, Shop Owner, Shop Assistant, Insurance Agent, Buyer \\[0.1cm]
    Service & Restaurant Owner, Police Officer, Waiter/Waitress, Barber, Caretaker \\[0.1cm]
    Skilled Worker & Foreman, Motor Mechanic, Printer, Seamstress, Tool and Die Maker, Electrician \\[0.1cm]
    Semi-skilled Worker & Bricklayer, Bus Driver, Cannery Worker, Carpenter, Sheet Metal Worker, Baker \\[0.1cm]
     Unskilled Worker & Laborer, Porter, Unskilled Factory Worker, Cleaner \\[0.1cm]
    Farm Worker & Farm Laborer, Tractor Driver \\[0.1cm]
    Farm Manager & Farm Proprietor, Farm Manager \\[0.1cm]
    Unemployed & Unemployed \\
    \bottomrule
  \end{tabular}
  \caption{Details of Occupational groups.}
  \label{tab:occupational_groups}
\end{table*}

The \textsc{MAPS} dataset combines user personas from WVS with questionnaires from both WVS and GAS. Table~\ref{tab:dataset_statistics} summarizes the overall statistics, and the subsections below provide further information on each individual component of the proposed benchmark.

\subsection{Personas}
We select 50 countries and 57,647 participants from WVS Wave~7, and extract 10 persona attributes from the introduction, political culture \& regimes, and demographics (shown in Table~\ref{tab:persona_schema}). The specification of occupational groups is listed in Table~\ref{tab:occupational_groups}.

\subsection{Questionnaire}
We sample 300 questions from GlobalOpinionQA, including questions, answer choices, and country-level response distributions, with 263 items from GAS and 37 from WVS. 

\section{Experiment}
\label{app:experiment}

\begin{figure*}[ht]
    \centering
    \setlength{\abovecaptionskip}{0pt}
    \setlength{\belowcaptionskip}{0pt}
    \begin{subfigure}{0.48\textwidth}
        \centering
        \includegraphics[width=\linewidth]{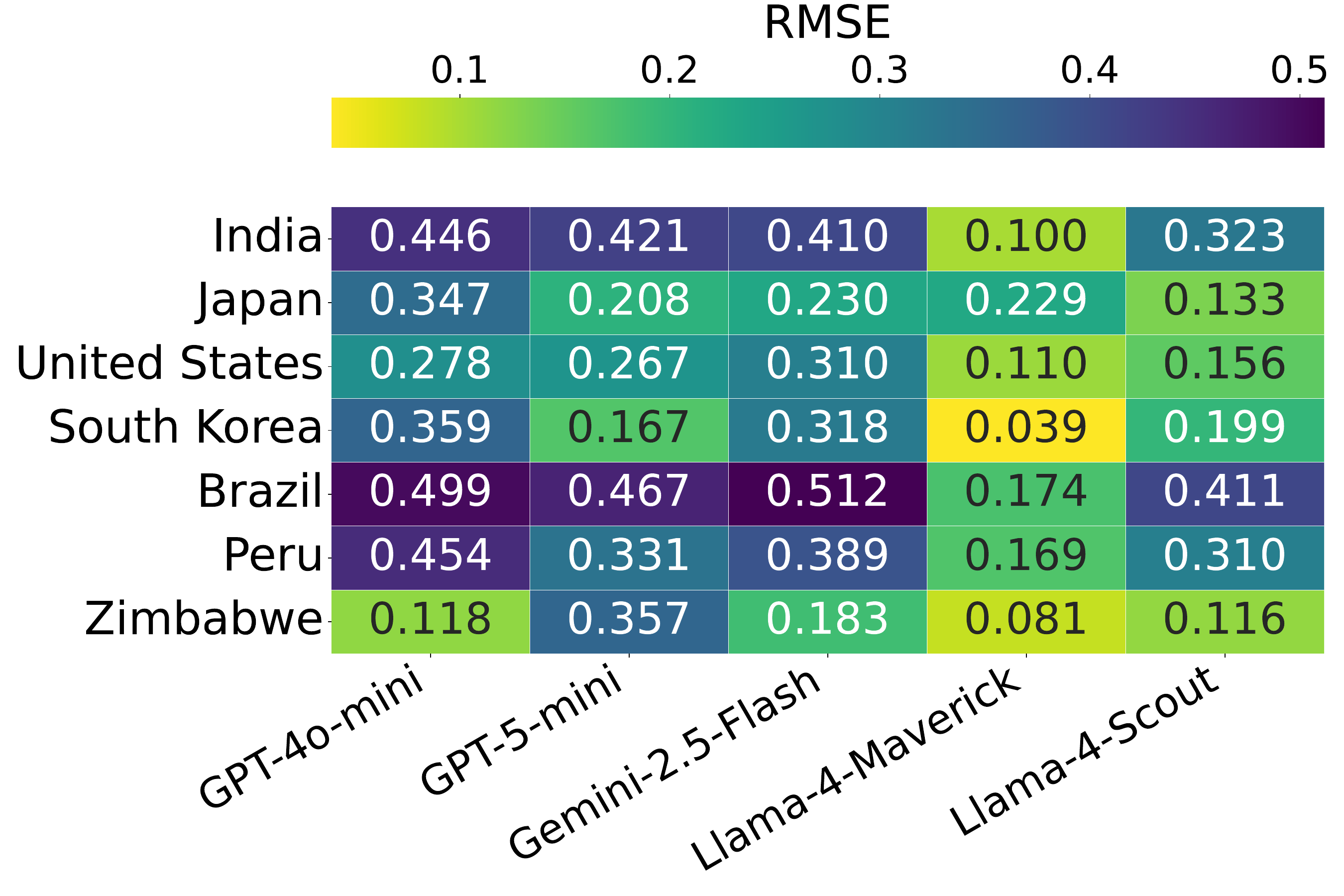}
        \caption{Real-world calibration by model and country.}
        \label{fig:real_world_calibration_by_country}
    \end{subfigure}
    \hfill
    \begin{subfigure}{0.48\textwidth}
        \centering
        \includegraphics[width=\linewidth]{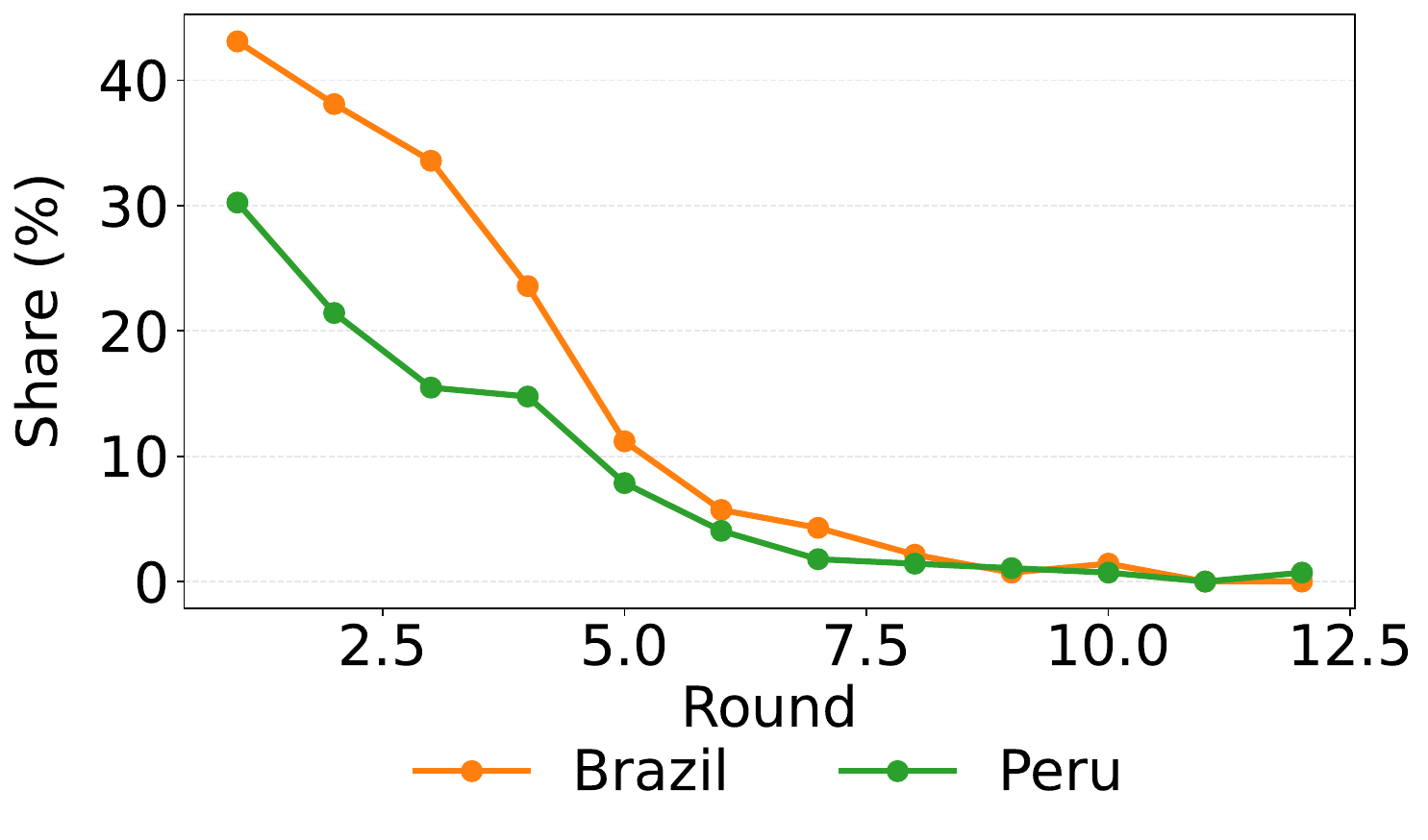}
        \caption{Foreign recommendation proportion for Case 2 (South Korea). Brazil is dominant in foreign recommendations, and in the first two rounds the ratio of recommendations from foreign countries (73.3\% to 59.5\%) surpasses that of South Korean domestic content. After round 12, a filter bubble is formed, and South Korean users no longer receive any posts from Brazil or Peru thereafter.}
        \label{fig:case_2_study}
    \end{subfigure}
    \caption{Real-world calibration and foreign recommendation analysis.}
    \label{fig:calibration_and_recommendation}
\end{figure*}

\begin{figure*}[ht]
    \centering
    \setlength{\abovecaptionskip}{3pt}
    \setlength{\belowcaptionskip}{0pt}

    \begin{subfigure}{0.32\textwidth}
        \centering
        \includegraphics[width=\linewidth]{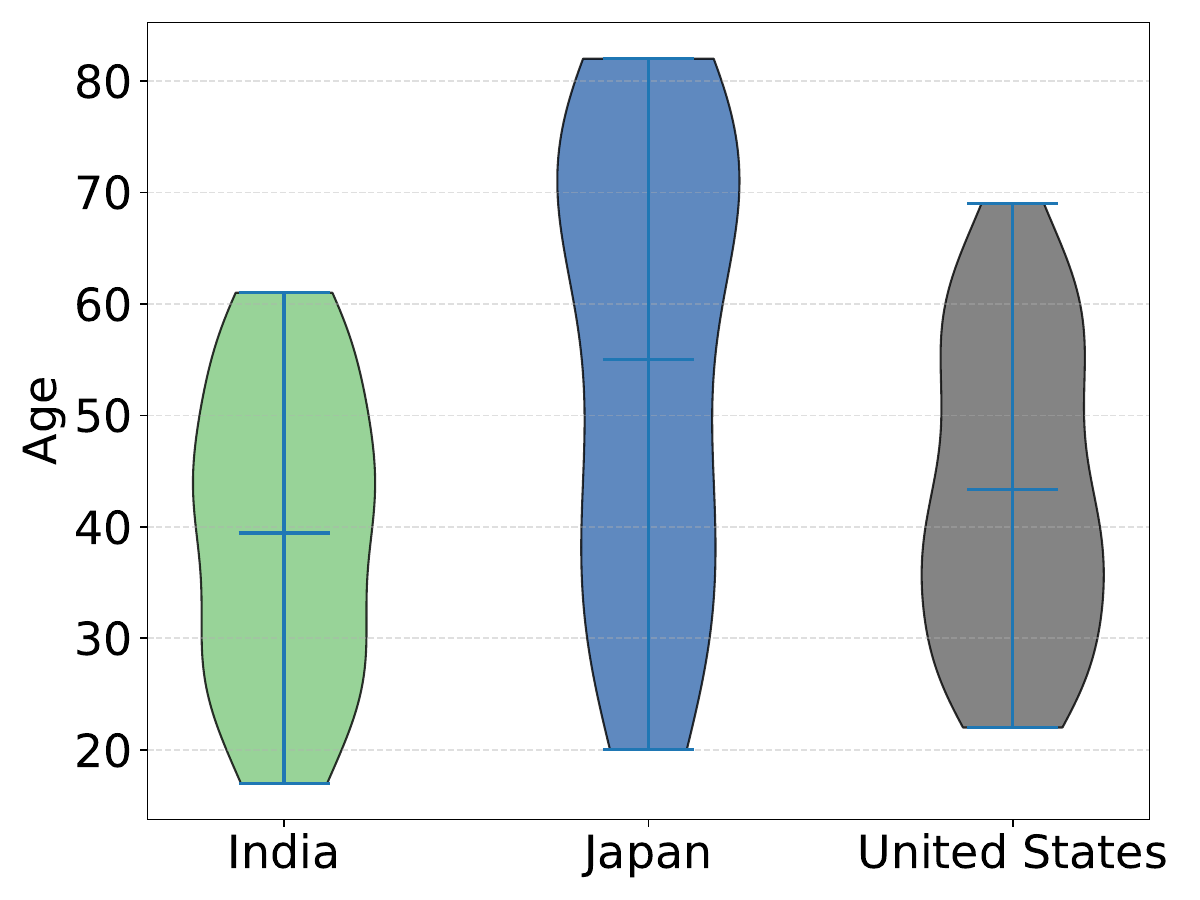}
        \caption{Case 1: Q201.}
        \label{fig:q201_age_distribution_jp}
    \end{subfigure}
    \hfill
    \begin{subfigure}{0.32\textwidth}
        \centering
        \includegraphics[width=\linewidth]{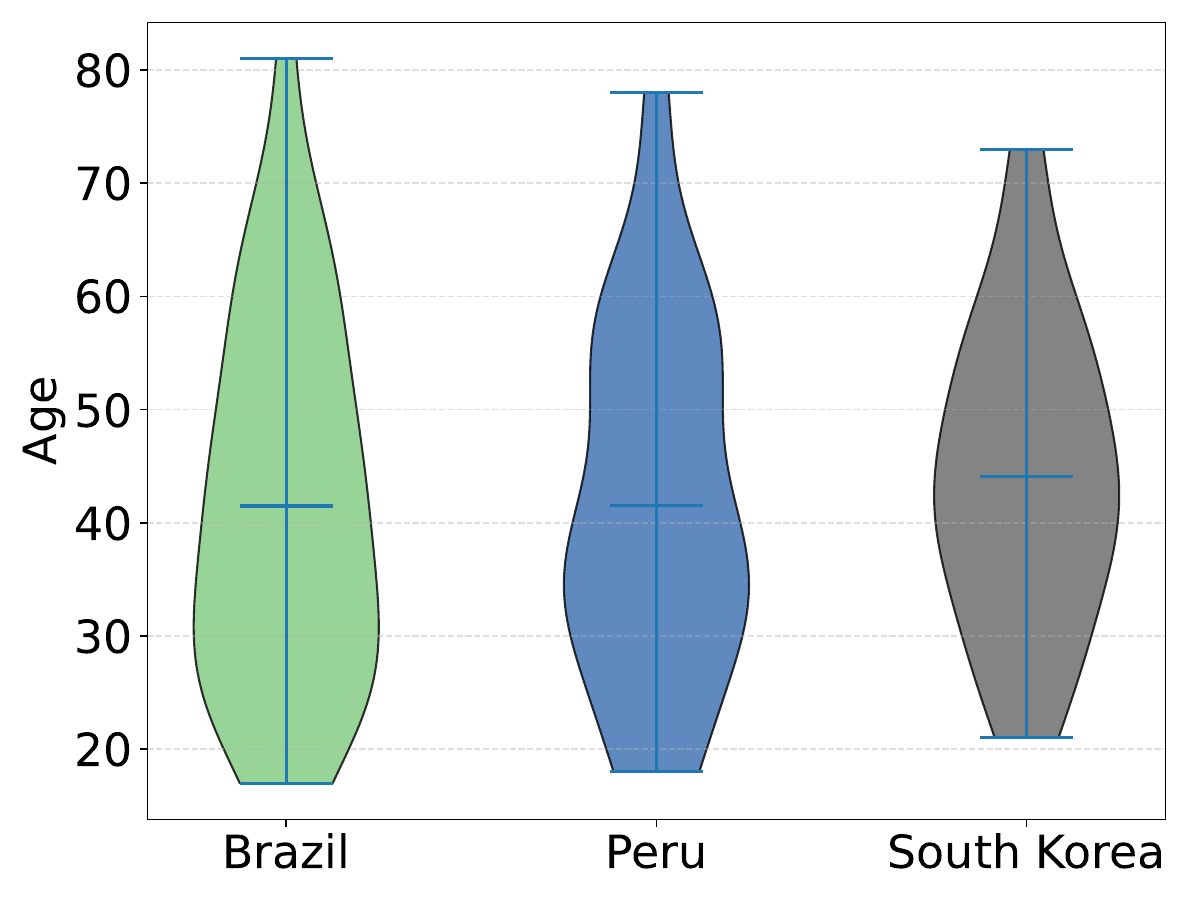}
        \caption{Case 2: Q201.}
        \label{fig:q201_age_distribution_kr}
    \end{subfigure}
    \hfill
    \begin{subfigure}{0.32\textwidth}
        \centering
        \includegraphics[width=\linewidth]{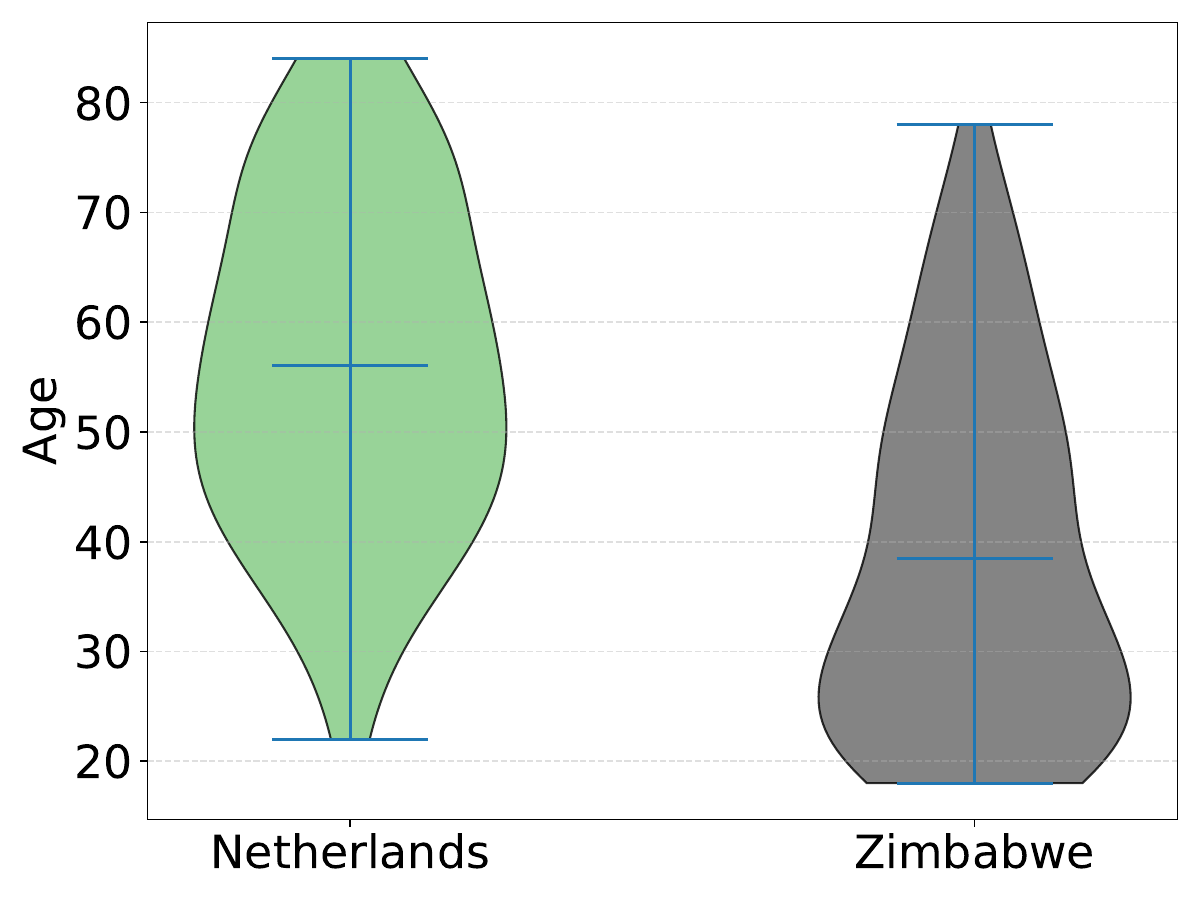}
        \caption{Case 3: Q278.}
        \label{fig:q278_age_distribution}
    \end{subfigure}

    \caption{Age distributions for the three cases.}
    \label{fig:age_distributions}
\end{figure*}

In the experiments, we assess three research cases in the \textsc{MAWM} environment, where each simulation runs 100 user agents for 21 rounds, including an initial warm-up round.
\subsection{Implementation Details}
\label{app:implementation_details}
The five LLMs were selected for their stability in multilingual settings and structured JSON outputs, both of which are required by our simulation. For the multilingual recommendation system, we use the multilingual embedding model \texttt{jina-embeddings-v3} together with \textit{Google Translate}. The hyperparameters used in our experiments are as follows: the total number of simulation rounds is $T = 20$, the maximum long-term memory retrieval size is $k_m = 3$, the number of recommended posts per round is $k_r = 5$, the temperature for attitude regularization is $\phi = 0.25$, the decay factor for long-term memory retrieval is $\lambda_m = 0.9$, the mixing coefficient for long-term memory is $\alpha_m = 0.5$, the decay factor for recommendation is $\lambda_r = 0.9$, the EMA rate for user embedding update is $\alpha_r = 0.1$, and the random seed is $42$.

\paragraph{Formal Definitions}
From each \textit{reading posts} action, the output is a set of weighted takeaways $\{(m_{\ell}, w_{\ell})\}_{\ell=1}^{L_{u_i}^t}$, where $L_{u_i}^t$ is the number of takeaways for agent $u_i$ at round $t$ and $w_{\ell} \in [0,1]$ is the importance weight of $m_{\ell}$. After min-max normalization of $w_{\ell}$, the long-term memory is updated as $\bar{M}_{u_i}^{t} \leftarrow \bar{M}_{u_i}^{t-1} \cup {(m_{\ell}, w_{\ell})}_{\ell=1}^{L_{u_i}^{t}}$. The top-$k_m$ memories are retrieved by: $\text{Score}(m, t) = (1 - \alpha_m)\, w + \alpha_m\, \lambda_m^{\delta(m, t)}$, where $\alpha_m \in [0,1]$ controls the trade-off between importance weight $w$ and temporal decay, $\lambda_m \in (0,1)$ is the decay factor, and $\delta(m, t) = \tfrac{1}{2}((t - \tau_0)_+ + (t - \tau_1)_+)$ averages the elapsed rounds since the memory's creation round $\tau_0$ and its last-use round $\tau_1$, with $(\cdot)_+$ denoting $\max(\cdot, 0)$.

The attitude distribution $d_{u_i} \in \Delta^{|C|-1}$ of user $u_i$ over $|C|$ response choices is regularized from the raw output $\tilde{d}_{u_i}$ through temperature softmax: $$d_{u_i} = \left( \frac{e^{\tilde{d}_{u_i}^c / \phi}}{\sum_{k=1}^{|C|} e^{\tilde{d}_{u_i}^k / \phi}} \right)_{c=1}^{|C|}$$ where $\phi$ is the softmax temperature that controls the sharpness of the resulting distribution.

For the recommendation system, the embedding $\bar{z}_i^{(t)}$ of agent $i$ at round $t$ is updated via Exponential Moving Average (EMA): $\bar{z}_i^{(t)} = (1 - \alpha_r)\, \bar{z}_i^{(t-1)} + \alpha_r\, \tilde{z}_{i,t-1}$, where $\tilde{z}_{i,t-1}$ is the embedding of the agent's most recent post $p_i^{t-1}$, and $\alpha_r \in (0,1)$ is the EMA update rate. Each post $p$ is encoded into an embedding $\tilde{z}_p = f_{\text{enc}}(p)$, normalized for cosine similarity. The similarity between agent $i$ and post $p$ is: $\text{S}(i, p) = \lambda_r^{(t - \mu(p))} \cdot \cos(\bar{z}_i^{t}, \tilde{z}_p)$, where $\lambda_r \in (0,1)$ is the recency decay factor and $\mu(p)$ is the round in which $p$ was posted.

\subsection{Case Description}
\label{app:case_description}
For our case design, we use Q201 from GAS, \textit{Does trade with other countries lead to an increase in the wages of your nationality’s workers, a decrease in wages, or does it not make a difference?}, for the first two cases: Case~1 includes India, Japan, and the United States, and Case~2 includes South Korea, Brazil, and Peru. For the third case, we use Q278 from WVS with Zimbabwe: \textit{A girl should honor the decisions or wishes of her family even if she does not want to marry}.

In the first two cases, we use 100 user agents in total and allocate them to each country according to that country’s population share in the \textsc{MAPS} dataset. In the third case, we sample 100 users from Zimbabwe and 50 users from the Netherlands for later case studies. The basic persona distributions for Cases 1, 2, and 3 are summarized in Tables~\ref{tab:case_1_persona_dist}, \ref{tab:case_2_persona_dist}, and \ref{tab:case_3_persona_dist}, respectively. The corresponding age distributions are showcased in Figure~\ref{fig:age_distributions}.

\subsection{Real-World Calibration}
\label{app:real_world_calibration}
To compare the simulation results with real-world survey responses, we examine (1) the difference between the distribution in the \textsc{MAPS} dataset $\mathcal{D}_{\text{real}}$ and the emulated distributions from the five LLMs $\mathcal{D}_{\text{sim}}$, and (2) whether using native language or English simulation yields results closer to the real-world survey for the best model from (1), \texttt{Llama-4-Maverick}. Without cross-cultural communication, the RMSE between the average simulated attitude distributions (over rounds and countries) and the average survey attitudes (over countries) is used as the primary evaluation criterion for each evaluated model backbone.

For each case, suppose the simulation generates attitude distributions for $N_u$ users during $T$ rounds:
$
\mathcal{D}_{\text{sim}}
= \{ d_{u_i}^{(t)} \mid u_i = 1,\ldots,N_u,\; t = 0,\ldots,T \},
$
where
$
d_{u_i}^{(t)} = \bigl(d_{u_i,1}^{(t)},\ldots,d_{u_i,|C|}^{(t)}\bigr) \in \Delta^{|C|-1}
$
is the attitude distribution of user $u_i$ at round $t$ over the $|C|$ response options.

For each model $\theta$, country $x \in X$, and round $t \in T$, the country-level distribution is
$$
d_{\theta,x}^{(t)}
= \frac{1}{|U_x|} \sum_{u_i \in U_x} d_{u_i}^{(t)},
$$
where $U_x$ is the set of users with country $x$.
The corresponding real survey distribution is
$$
d_x^{\text{real}} \in \Delta^{|C|-1}.
$$

The mean squared error (MSE) between the real-world and simulated distributions for model $\theta$, country $x$, and round $t$, taking values in the range $\bigl[0, \tfrac{2}{|C|}\bigr]$, is denoted by 
$$
\mathrm{MSE}_{\theta,x}^{(t)}
= \frac{1}{|C|} \sum_{c=0}^{|C|-1}
\bigl(d_{x,c}^{\text{real}} - d_{\theta,x,c}^{(t)}\bigr)^2
$$

The country-level RMSE in our experiments is
$$
\mathrm{RMSE}_{\theta,x}^{\text{country}}
= \sqrt{
\frac{1}{|T|}
\sum_{t=1}^T \mathrm{MSE}_{\theta,x}^{(t)}
},
$$
and the overall RMSE for model $\theta$ aggregated over all countries and simulation rounds is
$$
\mathrm{RMSE}_{\theta, x}^{\text{overall}}
= \sqrt{
\frac{1}{|X|\,|T|}
\sum_{x=1}^{X} \sum_{t=1}^T \mathrm{MSE}_{\theta,x}^{(t)}
}.
$$ 
The range of both RMSE metrics is
$
\Bigl[0,\sqrt{\tfrac{2}{|C|}}\Bigr].
$
\begin{figure*}[ht]
    \centering
    \setlength{\abovecaptionskip}{0pt}   
    \setlength{\belowcaptionskip}{0pt}
    \centering 
    \includegraphics[width=\textwidth]{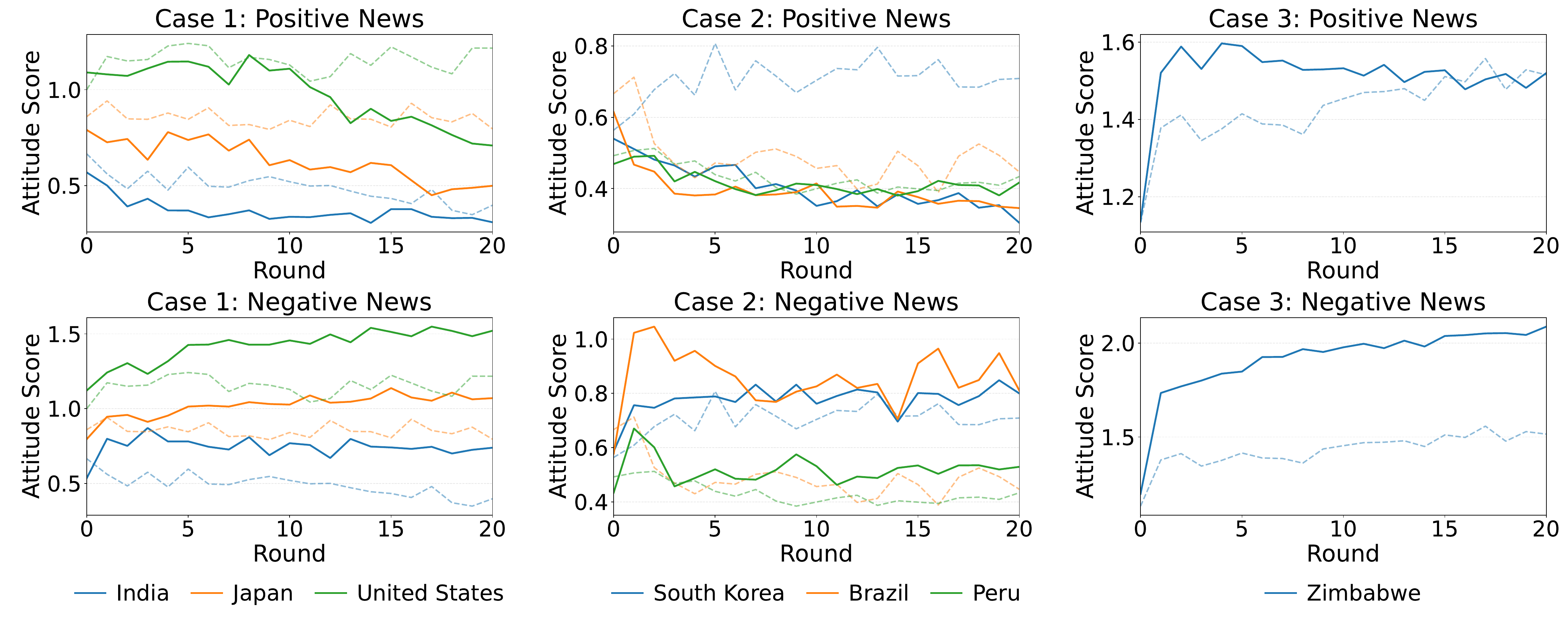}
    \caption{Global sensitivity attitude scores across countries and rounds.}
    \label{fig:global_sensitivity_by_country}
\end{figure*}

Our calibration experiments run three independent trials and report the average country-level RMSE in Figure~\ref{fig:real_world_calibration_by_country} and the overall RMSE in Figure~\ref{fig:experimental_results}(a). \texttt{Llama-4-Maverick} achieves the best performance across all three cases, outperforming other models in every country except Japan. We therefore use this model for all remaining analyses reported in this work.

To assess the effect of simulation language, we run three independent experiments in English and in each native language for the Case~1 and~2 countries, and further extend the comparison to the 17 remaining non-English countries in \textsc{MAPS} with Q201 ground truth, simulating each population with 100 agents over 20 rounds using \texttt{GPT-4o-mini}. The RMSE results appear in Figure~\ref{fig:language_rmse} and in Subsection~\ref{subsec:reliability}.

\subsection{Global Sensitivity}
\label{app:global_sensitivity}
This analysis evaluates how positive and negative news shape user agents’ attitudes across the three cases under the condition of no international discourse. In Cases 1 and 2, each country includes one additional news organization agent that adopts option A in the positive news condition and option B in the negative news condition alongside the user agents. For the Case 3 simulation, each country is assigned a news organization agent adopting option A under the positive news condition and option D under the negative news condition. 

To capture the impact of introducing news organization agents, we compute the average attitude distribution shift across the last three rounds. We define the positive news shift as:
\begin{itemize}[leftmargin=*, itemsep=6pt, topsep=2pt, parsep=1pt]
    \item \textbf{Cases 1 and 2}: Change in user attitude distribution toward option A, i.e., 
    $$
    \frac{1}{3 |U_x|} \sum_{t = T-2}^{T} \sum_{u_i \in U_x} (d_{u_i,A}^{(t)} - \tilde{d}_{u_i,A}^{(t)} )
    $$

    \item \textbf{Case 3}: Combined user attitude distribution shift toward options A and B, i.e., 
    $$
    \frac{1}{3 |U_x|} \sum_{t = T-2}^{T} \sum_{u_i \in U_x} (d_{u_i,A}^{(t)} + d_{u_i,B}^{(t)} - \tilde{d}_{u_i,A}^{(t)} - \tilde{d}_{u_i,B}^{(t)})
    $$
\end{itemize}
Here, $\tilde{d}$ denotes the pre-intervention distribution, identical to the distribution used in the real-world calibration experiments previously described in Appendix~\ref{app:real_world_calibration}.

Similarly, the negative news shift is defined as:
\begin{itemize}[leftmargin=*, itemsep=6pt, topsep=2pt, parsep=1pt]
    \item \textbf{Cases 1 and 2}: Change in user attitude distribution toward option B, i.e., 
    $$
    \frac{1}{3 |U_x|} \sum_{t = T-2}^{T} \sum_{u_i \in U_x} (d_{u_i,B}^{(t)} - \tilde{d}_{u_i,B}^{(t)} )
    $$

    \item \textbf{Case 3}: Combined user attitude distribution shift toward options C and D, i.e., 
    $$
    \frac{1}{3 |U_x|} \sum_{t = T-2}^{T} \sum_{u_i \in U_x} (d_{u_i,C}^{(t)} + d_{u_i,D}^{(t)} - \tilde{d}_{u_i,C}^{(t)} - \tilde{d}_{u_i,D}^{(t)} )
    $$
\end{itemize}

Media effects are bounded by the available headroom. To compare countries on a common scale, we therefore present normalized results, dividing each raw shift by its available headroom, $\hat{\Delta}_o = \Delta_o / (1 - \tilde{p}_o)$, where $\Delta_o$ is the raw shift defined above and $\tilde{p}_o$ denotes the baseline share of the targeted option $o$ (for the negative-news effect, the headroom is computed from the baseline share of the opposing option). This quantity measures the fraction of movable opinion captured by media exposure and distinguishes ceiling effects, genuine stability, and directional asymmetry. For example, India starts at $0.75$ support for \emph{Increase}, leaving at most $0.25$ headroom; after normalization, its near-zero raw positive effect ($+0.032$) becomes a moderate $0.13$, indicating a ceiling effect rather than insensitivity. Zimbabwe emerges as the strongest negative responder ($0.51$) despite a near-zero positive effect, revealing pronounced directional asymmetry, whereas Peru remains small in both directions ($\approx 0.05$), indicating genuine attitudinal stability of the simulated Peruvian population under both the positive and the negative media interventions.

Figure~\ref{fig:experimental_results}(b) presents the normalized global sensitivity to positive and negative news across the seven participating countries. In addition, Figure~\ref{fig:global_sensitivity_by_country} reports how global sensitivity evolves over rounds, quantified by the attitude score as defined in Section~\ref{sec:case_study}.

\begin{table}[ht]
\centering
\scriptsize
\setlength{\abovecaptionskip}{1pt}
\setlength{\belowcaptionskip}{0pt}
\begin{tabular}{lcccc}
  \toprule
  \textbf{Action} & \textbf{Alignment} & \textbf{Coherence} & \textbf{Grounding} & \textbf{Uniqueness} \\
  \midrule
  \multicolumn{5}{c}{\textbf{Case 1: India, Japan, United States}} \\
  \midrule
  User Self-Intro                  & $4.533 \pm 0.640$ & $4.667 \pm 0.617$ & $-$ & $-$ \\
  User Reading Post                & $4.773 \pm 0.421$ & $4.520 \pm 0.529$ & $4.440 \pm 0.551$ & $-$ \\
  User Writing Post                & $4.533 \pm 0.528$ & $4.760 \pm 0.430$ & $-$ & $-$ \\
  User Voting                      & $4.627 \pm 0.514$ & $4.533 \pm 0.664$ & $-$ & $-$ \\
  Organization Self-Intro          & $5.000 \pm 0.000$ & $5.000 \pm 0.000$ & $-$ & $-$ \\
  Organization Writing Post        & $5.000 \pm 0.000$ & $5.000 \pm 0.000$ & $-$ & $3.067 \pm 0.961$ \\
  \midrule
  \multicolumn{5}{c}{\textbf{Case 2: Brazil, Peru, South Korea}} \\
  \midrule
  User Self-Intro                  & $4.333 \pm 0.724$ & $4.600 \pm 0.632$ & $-$ & $-$ \\
  User Reading Post                & $4.720 \pm 0.481$ & $4.547 \pm 0.552$ & $4.387 \pm 0.590$ & $-$ \\
  User Writing Post                & $4.560 \pm 0.500$ & $4.827 \pm 0.381$ & $-$ & $-$ \\
  User Voting                      & $4.560 \pm 0.526$ & $4.707 \pm 0.458$ & $-$ & $-$ \\
  Organization Self-Intro          & $5.000 \pm 0.000$ & $5.000 \pm 0.000$ & $-$ & $-$ \\
  Organization Writing Post        & $5.000 \pm 0.000$ & $4.800 \pm 0.414$ & $-$ & $3.267 \pm 1.163$ \\
  \midrule
  \multicolumn{5}{c}{\textbf{Case 3: Zimbabwe}} \\
  \midrule
  User Self-Intro                  & $4.400 \pm 0.632$ & $4.733 \pm 0.458$ & $-$ & $-$ \\
  User Reading Post                & $4.747 \pm 0.522$ & $4.520 \pm 0.554$ & $4.227 \pm 0.583$ & $-$ \\
  User Writing Post                & $4.493 \pm 0.578$ & $4.640 \pm 0.671$ & $-$ & $-$ \\
  User Voting                      & $4.507 \pm 0.795$ & $3.880 \pm 1.230$ & $-$ & $-$ \\
  Organization Self-Intro          & $4.000 \pm 0.000$ & $5.000 \pm 0.000$ & $-$ & $-$ \\
  News Organization Writing Post   & $5.000 \pm 0.000$ & $5.000 \pm 0.000$ & $-$ & $2.400 \pm 0.548$ \\
  \bottomrule
\end{tabular}
\caption{Alignment, coherence, grounding and uniqueness evaluation.}
\label{tab:local_consistency_align_coh}
\end{table}

\subsection{Local Consistency}
\label{app:local_consistency}
We assess agents' behavioral consistency relative to their predefined personas or profiles using \texttt{GPT-5} as an LLM judge, scoring performance on a 1-5 scale (1 = poor, 5 = excellent). User agents are evaluated on four actions (\textit{self-introduction, reading posts, writing posts}, and \textit{poll responses}), whereas news agents are evaluated on two actions (\textit{self-introduction} and \textit{writing posts}). The evaluation follows the same configuration as the global sensitivity analysis with positive news. We sample 5 user agents per country for Cases~1 and~2 (15 total), and 15 user agents from Zimbabwe for Case~3. All news agents are included.

For each evaluated action, we apply the \textit{alignment} and \textit{coherence} metrics: the former measures whether an agent’s response aligns with its assigned persona or profile, while the latter assesses whether its short-term memory reasoning is coherent with its final output. For the user reading posts action, we introduce an additional \textit{grounding} metric to determine whether the generated long-term memory is supported by the content the agent has read. For the organization writing posts action, we add the \textit{uniqueness} metric, since we expect the news produced by organization agents to be both distinctive and continuous with prior posts. The full evaluation instructions are provided in Prompts~\ref{prm:llm_judge_request}-\ref{prm:llm_judge_rubric} below.

Local consistency under the \textit{alignment}, \textit{coherence}, \textit{grounding} and \textit{uniqueness} metrics is summarized in Table~\ref{tab:local_consistency_align_coh}. The overall score is derived by applying metric weights of $0.3$ (alignment), $0.5$ (grounding), $0.2$ (coherence), and $0.5$ (uniqueness), and computing a weighted sum scaled to 1. The resulting overall evaluations are shown in Figure~\ref{fig:experimental_results}(c). Illustrative examples of the local consistency evaluation across all actions are shown in Examples~\ref{eg:case_1_user_intro}-\ref{eg:case_3_org_write}.

\subsection{Culture--Language Disentanglement}
\label{app:culture_language}
To disentangle culture from language, we test the triple Argentina, Bolivia, and Peru on two \textsc{MAPS} questions, Q201 (whether trade increases wages) and Q68 (whether the gap between rich and poor is a problem), with \texttt{GPT-4o-mini} and \texttt{Llama-4-Maverick}, simulating 50 agents per country over 20 rounds of interaction. We read the resulting composition at two levels: the aggregate vote distribution and the post-embedding space of agent-generated content.

The target composition, Spain (European) $+$ Bolivia (Indigenous) $\approx$ Peru, cannot be tested directly because Spain is not part of \textsc{MAPS}, so Spanish personas cannot be instantiated. We therefore choose a triple that reproduces the same European/Indigenous composition entirely within Spanish-speaking populations, so that language is held fixed and only culture varies: (1) Argentina as the European-heritage pole, the most European-descent population in South America (about 85\% European ancestry), standing in for the role Spain would play; (2) Bolivia as the Indigenous-heritage pole (about 62\% Indigenous, Andean); and (3) Peru as the composition target, a mestizo-majority population with substantial European and Indigenous ancestry that sits demographically and geographically between the two. Because all three populations speak Spanish, composing them holds language constant, so any Argentina $+$ Bolivia $\approx$ Peru effect is a composition of \emph{culture} representations, not of language; prior work shows that expressed culture is conditioned by persona rather than by prompt language \citep{bulte_llms_2025, lutz_prompt_2025}. The complementary contrast (same population, different language) is the native-versus-English comparison in Figure~\ref{fig:language_rmse}, which varies language with culture fixed; here we vary culture with language fixed. Each model runs in two modes: isolated (Argentina, Bolivia, and Peru run separately) and communicating (Argentina and Bolivia interact with each other).

At the vote level, we score each isolated population by RMSE against its own real survey (Table~\ref{tab:composition_rmse}). On Q201, \texttt{GPT-4o-mini} aligns worse with the real survey than \texttt{Llama-4-Maverick}; Bolivia has no Q201 ground truth, so only Argentina and Peru are scorable there. On Q68, both models are well-calibrated and all three countries are scorable; here \texttt{GPT-4o-mini} is in fact the better-calibrated of the two, with lower RMSE on all three countries.

\begin{table}[ht]
\centering
\small
\setlength{\abovecaptionskip}{4pt}
\setlength{\belowcaptionskip}{0pt}
\begin{tabular}{llcc}
  \toprule
  \textbf{Question} & \textbf{Population} & \texttt{GPT-4o-mini} RMSE & \texttt{Llama-4-Maverick} RMSE \\
  \midrule
  Q201 (trade)      & Argentina & 0.455 & \textbf{0.256} \\
  Q201 (trade)      & Peru      & 0.472 & \textbf{0.181} \\
  Q201 (trade)      & Bolivia   & n/a   & n/a \\
  \midrule
  Q68 (rich--poor)  & Argentina & \textbf{0.048} & 0.081 \\
  Q68 (rich--poor)  & Bolivia   & \textbf{0.088} & 0.162 \\
  Q68 (rich--poor)  & Peru      & \textbf{0.131} & 0.190 \\
  \bottomrule
\end{tabular}
\caption{Vote-level calibration RMSE of each isolated population against its own real survey in the culture--language disentanglement experiment. Bolivia has no Q201 ground truth, so only Argentina and Peru are scorable there. Bold marks the better-calibrated model per row.}
\label{tab:composition_rmse}
\end{table}

Because the discrete vote over a few options carries little cultural signal once the simulated populations converge, and vector calculus between culture representations requires a representation space, we measure composition over the posts the agents generate. We embed every agent post with \texttt{jina-embeddings-v3}, the same encoder the simulation uses, average the post embeddings within each population into one centroid vector, and report the cosine distance of each centroid to Peru's, where a smaller value means more Peru-like (Table~\ref{tab:composition_embedding}). Across both questions, both models, and both aggregations, the communicating Argentina$+$Bolivia population is closest to Peru, closer than Argentina or Bolivia alone: combining the Argentine and Bolivian populations yields content that sits nearest Peru, i.e., Argentina $+$ Bolivia $\approx$ Peru holds in the learned representation space.

\begin{table}[ht]
\centering
\scriptsize
\setlength{\tabcolsep}{4pt}
\setlength{\abovecaptionskip}{4pt}
\setlength{\belowcaptionskip}{0pt}
\resizebox{\textwidth}{!}{%
\begin{tabular}{llccc}
  \toprule
  \textbf{Question / Model} & \textbf{Rounds} & Argentina vs.\ Peru & Bolivia vs.\ Peru & Argentina$+$Bolivia (communicating) vs.\ Peru \\
  \midrule
  Q201 / \texttt{Llama-4-Maverick} & all rounds & 0.080  & 0.071  & \textbf{0.050} \\
  Q201 / \texttt{Llama-4-Maverick} & last round & 0.091  & 0.073  & \textbf{0.049} \\
  Q68 / \texttt{GPT-4o-mini}       & all rounds & 0.0275 & 0.0276 & \textbf{0.0183} \\
  Q68 / \texttt{GPT-4o-mini}       & last round & 0.0243 & 0.0260 & \textbf{0.0186} \\
  \bottomrule
\end{tabular}}
\caption{Cosine distance between each population's post-embedding centroid and Peru's centroid (smaller means more Peru-like). The Argentina and Bolivia columns are the isolated runs, while Argentina$+$Bolivia is the communicating run; the latter is closest to Peru in every row, i.e., Argentina $+$ Bolivia $\approx$ Peru in the learned representation space.}
\label{tab:composition_embedding}
\end{table}

\subsection{Robustness Analysis}
\label{app:robustness}
\paragraph{Stability Across Runs} Each case study uses three independent runs. Table~\ref{tab:attitude_std} reports the across-run standard deviation of the attitude score for every trajectory in Figure~\ref{fig:case_study}(a). The variation is small relative to the trajectory movement, especially for the Case~2 and Case~3 curves, indicating that the reported dynamics are reproducible across runs.

\begin{table}[ht]
\centering
\scriptsize
\setlength{\tabcolsep}{4.5pt}
\setlength{\abovecaptionskip}{4pt}
\setlength{\belowcaptionskip}{0pt}
\begin{tabular}{lcccccccc}
  \toprule
  & \multicolumn{3}{c}{\textbf{Case 1}} & \multicolumn{3}{c}{\textbf{Case 2}} & \multicolumn{2}{c}{\textbf{Case 3 (Zimbabwe)}} \\
  \cmidrule(lr){2-4} \cmidrule(lr){5-7} \cmidrule(lr){8-9}
  & India & Japan & United States & South Korea & Brazil & Peru & NL users & NL news \\
  \midrule
  \textbf{Across-run std} & 0.134 & 0.167 & 0.103 & 0.024 & 0.018 & 0.013 & 0.059 & 0.052 \\
  \bottomrule
\end{tabular}
\caption{Across-run standard deviation of the attitude score for each case-study trajectory, computed over three independent runs. The variation is small relative to the trajectory movement, indicating that the reported dynamics are reproducible across runs.}
\label{tab:attitude_std}
\end{table}

\paragraph{Temperature Sensitivity} Following the attitude tuning of \citet{hou_can_2025}, we apply a softmax temperature as a post-hoc transformation of the recorded poll distribution, $\mathrm{softmax}(\mathrm{raw}/T)$, which does not affect the simulation dynamics. Using the same RMSE-to-survey metric as Figure~\ref{fig:experimental_results}(a), we rescale the opinion distributions generated by \texttt{Llama-4-Maverick} and evaluate $T \in \{0.25, 0.5, 1.0, 1.5, 2.0\}$ on the three alignment cases (Table~\ref{tab:temperature_sweep}); the paper's results correspond to $T=0.25$. RMSE exhibits only modest sensitivity to temperature: the maximum variation across the sweep is $0.070$ in Case~2, while Cases~1 and~3 vary by only $0.024$ and $0.031$. Although $T=0.25$ is not always optimal, it remains close to the best-performing temperature in all cases, and since the temperature acts only as a readout transformation, the qualitative conclusions are robust to its choice.

\begin{table}[ht]
\centering
\small
\setlength{\abovecaptionskip}{4pt}
\setlength{\belowcaptionskip}{0pt}
\begin{tabular}{lccccc}
  \toprule
  \textbf{Experiment} & $T=0.25$ & $T=0.5$ & $T=1.0$ & $T=1.5$ & $T=2.0$ \\
  \midrule
  Case 1 (India, Japan, US)    & 0.158 & 0.134 & 0.143 & 0.149 & 0.152 \\
  Case 2 (Korea, Brazil, Peru) & 0.142 & 0.072 & 0.075 & 0.085 & 0.091 \\
  Case 3 (Zimbabwe)            & 0.081 & 0.060 & 0.052 & 0.051 & 0.050 \\
  \bottomrule
\end{tabular}
\caption{Real-world calibration RMSE under different softmax readout temperatures $T$ for \texttt{Llama-4-Maverick}. The temperature is applied as a post-hoc transformation of the recorded poll distributions and does not affect the simulation dynamics; the paper's results correspond to the default configuration $T=0.25$ that is used throughout this paper.}
\label{tab:temperature_sweep}
\end{table}

\section{Case Study}
\label{app:case_study}
\begin{figure*}[ht]
    \centering
    \setlength{\abovecaptionskip}{0pt}
    \setlength{\belowcaptionskip}{0pt}
    \includegraphics[width=\textwidth]{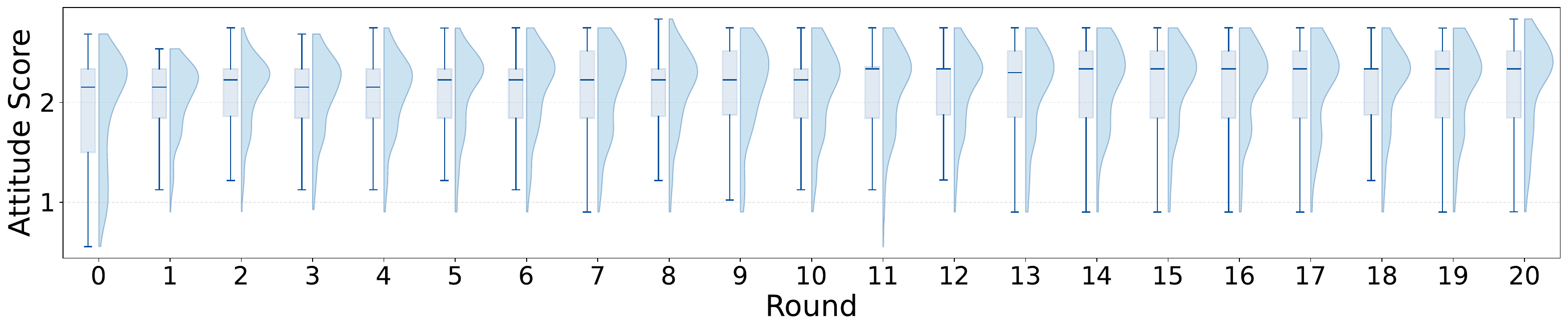}
    \caption{Attitude scores for 50 Netherlands users in Q278. Dutch users maintain stable pro-equality attitudes across all 20 rounds, even alongside opposing Zimbabwean users.}
    \label{fig:case_study_nl_score}
\end{figure*}

Section~\ref{sec:case_study} investigates two social science phenomena: cultural assimilation in Cases~1 and~2, and normative diffusion in Case~3. We visualize the attitude score and dominant foreign exposure ratio for each round in Figure~\ref{fig:case_study}. For a user $u_i$, the \textbf{Dominant Foreign Exposure Ratio} is defined as
$
\max_x \bigl|\{p_{u_j} \in \Gamma_{u_i} \mid x = \text{country}({u_j}) \neq \text{country}({u_i})\}\bigr|/|\Gamma_{u_i}|
$, where $p_{u_j}$ is a post by user $u_j$ in the recommendation set $\Gamma_{u_i}$ of user $u_i$, and $\text{country}(\cdot)$ denotes the country of a user agent. This identifies the foreign country $x$ that contributes the largest share of recommendations to each individual user agent $u_i$ per round.

\paragraph{Cultural Assimilation} We compare three independent runs before and after cross-nation discussion, where the recommendation system delivers translated posts from agents in other countries. Our objective is to determine which countries with disparate initial attitudes show convergence after the discourse. Figure~\ref{fig:case_study} shows the primary results. To clearly illustrate the influence of foreign recommendations on South Korea in Case 2, we report the recommendation shares of Brazil and Peru for South Korean users in Figure~\ref{fig:case_2_study}. These two South American countries jointly shift South Korean opinions on international trade.

\paragraph{Normative Diffusion} We investigate the transmission of the concept of \textit{gender equality} from the Netherlands (source) to Zimbabwe (target), through two different channels: users and news. When we compare the introduction of 50 Dutch users or the addition of 5 Dutch news agents supporting option~D to the no-intervention baseline, the news intervention produces a noticeably stronger effect, as clearly evidenced in Figure~\ref{fig:case_study}. 

To further explore the societal dynamics behind cultural assimilation, we analyze the recommendation network in Cases 1 and 2, and in Case 3 with 50 Dutch users (as these settings omit news organization agents). In each simulation round $t$, the recommendation system $R$ generates recommendation pairs of the form ${u_i, u_j}$, where $u_i$ is the author of a post and $u_j$ is the reader agent who received that post from $R$. We construct a directed weighted graph $G^{(t)} = (V^{(t)}, E^{(t)})$, where $V^{(t)}$ is the set of user agents and $E^{(t)}$ contains directed edges of the form $u_i \rightarrow u_j$ capturing the recommendation of posts from $u_i$ to $u_j$. The associated edge weight $w^{(t)}_{ij}$ denotes the number of posts from $u_i$ that were recommended to $u_j$. In this way, $G^{(t)}$ models the information flow from content creators to content consumers at round~$t$, with the edge weights encoding the intensity of algorithmic exposure.

\begin{figure*}[ht]
    \centering
    \setlength{\abovecaptionskip}{0pt}
    \setlength{\belowcaptionskip}{0pt}
    \includegraphics[width=\textwidth]{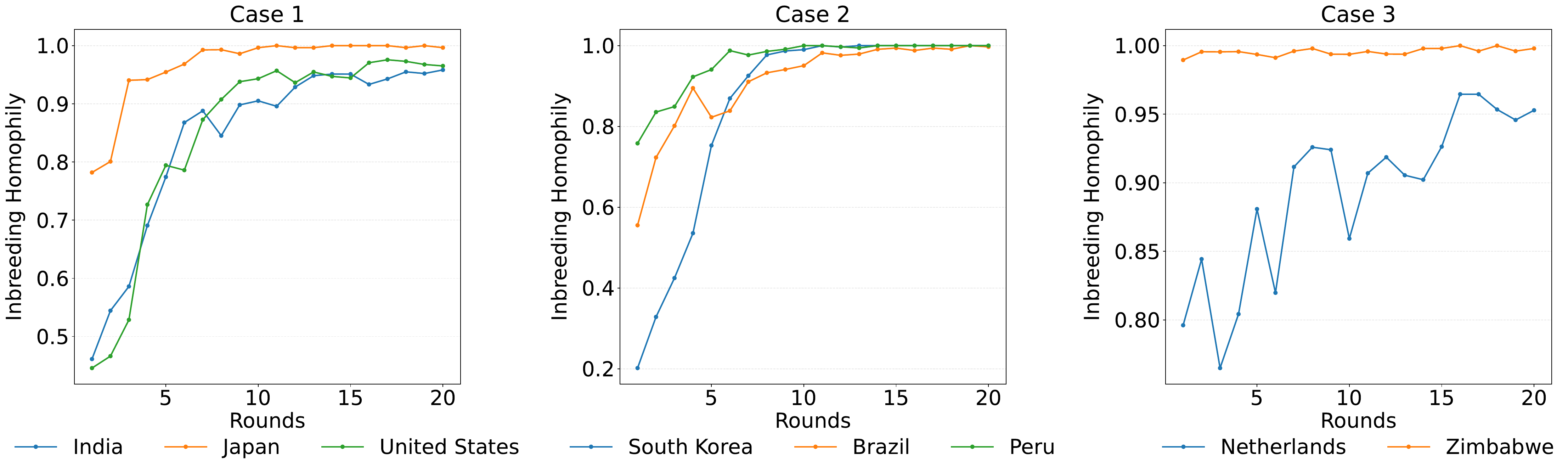}
    \caption{Inbreeding homophily.}
    \label{fig:case_study_ih}
\end{figure*}

\subsection{Inbreeding Homophily}
In order to model whether a country is more likely to read domestic posts or foreign posts, we adopt the inbreeding homophily index with the baseline correction \cite{inbreeding_homophily}.
To begin with, consider a user agent $u \in V^{(t)}$ from country $x$ at round $t$. We intend to quantify whether users from a given country tend to consume content from their own country more than would be expected based on the overall composition of content in the graph $G^{(t)}$.

We first aggregate the edges $E^{(t)}$ into a country-level information flow matrix $F^{(t)}$ over $G^{(t)}$:
\begin{align*}
  F^{(t)}_{a,r}
  \;=\;
  \sum_{\substack{u_i,u_j \in V^{(t)} \\ \text{country}(u_i)=a \\ \text{country}(u_j)=r}}
  w^{(t)}_{ij},
\end{align*}
which represents the total weighted volume of recommendation posts from authors in country $a$ to readers in country $r$ at round $t$. In other words, $F^{(t)}_{a,r}$ measures how much content users in country $r$ actually see from authors located in country $a$.

The total edge weight of the graph $G^{(t)}$ in the country-level information flow matrix is
\begin{align*}
  F_\text{sum}^{(t)} = \sum_{a \in \mathcal{C}} \sum_{r \in \mathcal{C}} F^{(t)}_{a,r},
\end{align*}
where $\mathcal{C}$ denotes the set of all countries. For a specific country $x \in \mathcal{C}$, we define:
\paragraph{Observed same-country share.}
The share of recommendations shown to users in country $x$ that originate from authors in the same country $x$ is
\begin{align*}
  p^{(t)}_{xx}
  =
  \frac{F^{(t)}_{x,x}}{\displaystyle\sum_{a \in \mathcal{C}} F^{(t)}_{a,x}}.
\end{align*}

\paragraph{Baseline availability.}
The overall share of content in the recommendation system $R$ that is produced by authors in country $x$ is
\begin{align*}
  q^{(t)}_{x}
  =
  \frac{\displaystyle\sum_{r \in \mathcal{C}} F^{(t)}_{x,r}}{F_{\text{sum}}^{(t)}}.
\end{align*}

The quantity $q^{(t)}_{x}$ measures how prominent country $x$ is in the overall recommendation traffic in round $t$. In this case, we use $q^{(t)}_{x}$ as a baseline for comparison: if the recommendation system did not differentiate between readers by country, then, on average, users in country $x$ would be exposed to content from $x$ in proportion to $q^{(t)}_{x}$. 

Therefore, the inbreeding homophily (IH) index for country $x$ at round $t$ is defined as
\begin{align*}
  \mathrm{IH}^{(t)}_{x}
  \;=\;
  \frac{p^{(t)}_{xx} - q^{(t)}_{x}}{1 - q^{(t)}_{x}}
  \qquad \text{for } q^{(t)}_{x} < 1.
\end{align*}
This normalization ensures that $\mathrm{IH}^{(t)}_{x} = 0$ when users in country $x$ consume domestic content exactly in proportion to its availability ($p^{(t)}_{xx} = q^{(t)}_{x}$). We have $\mathrm{IH}^{(t)}_{x} > 0$ when they \emph{over-consume} domestic content, and $\mathrm{IH}^{(t)}_{x} < 0$ when they \emph{under-consume} domestic content in favor of foreign content. The maximal value $\mathrm{IH}^{(t)}_{x} = 1$ occurs in the extreme case where the recommendation system only pushes domestic posts to users in country $x$.

Figure~\ref{fig:case_study_ih} shows the inbreeding homophily for each round across the three cases in the user-only cross-cultural communication setting, averaged over three independent runs. A common finding is that all countries tend to over-consume domestic posts as the rounds progress, consistent with the formation of echo chambers in the social network.

In Case~1, Japan exhibits a higher IH index at round~0 compared to the other countries, indicating that Japanese users are more isolated in the recommendation modeling from the very beginning. In Case~2, South Korean users are more willing to read foreign posts, which partly explains why their attitudes change more noticeably during the simulation. In Case~3, the IH index for Zimbabwe remains close to~1, indicating that promoting the norm of gender equality through users alone is ineffective, since the recommendation system does not deliver any foreign posts to Zimbabwean users in this configuration.

\begin{figure*}[ht]
    \centering
    \setlength{\abovecaptionskip}{0pt}   
    \setlength{\belowcaptionskip}{0pt}
    \includegraphics[width=\textwidth]{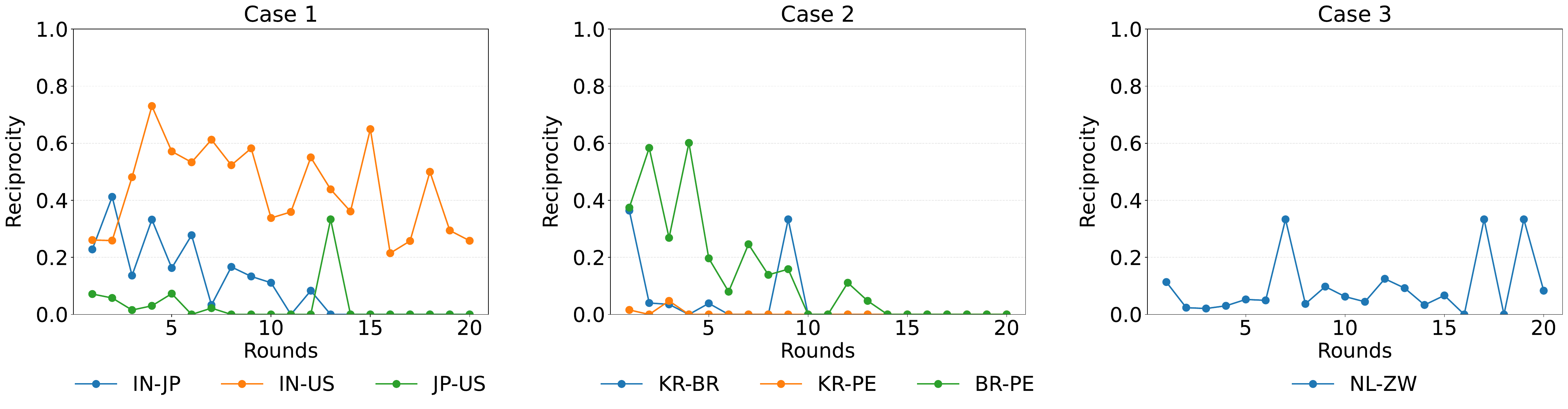}
    \caption{Cross-country reciprocity.}
    \label{fig:case_study_reciprocity}
\end{figure*}

\subsection{Information Diffusion Analysis}
This section inspects the recommendation-based post propagation process in Case~1 and Case~2. We focus on two countries: the United States from Case~1, whose user attitudes remain stable and distinct from those of the other two countries, and South Korea from Case~2, which exhibits a clear attitude shift during the cross-cultural simulation. 

We identify broker users for the United States in Case~1 and South Korea in Case~2 at certain rounds. A broker is defined as a user who receives at least one recommended post from a foreign user (the source) and whose own post in the same round is subsequently recommended to at least two domestic users (the audience). We then focus on instances where both the broker and its audience exhibit attitude shifts upon reading the post. 

We report the information diffusion process from foreign source users to brokers, and then from brokers to domestic audiences, for two selected periods in each case. For Case~1, we examine rounds 4-6 and rounds 9-11, as shown in Figures~\ref{fig:rec_network_case_1_4to5}-\ref{fig:rec_network_case_1_9to10} and Examples~\ref{eg:case1_US_4to5}-\ref{eg:case1_US_9to10}. For Case~2, we examine rounds 1-3 and rounds 6-8, as shown in Figures~\ref{fig:rec_network_case_2_4to5}-\ref{fig:rec_network_case_2_9to10} and Examples~\ref{eg:case2_KR_1to2}-\ref{eg:case2_KR_6to7}.

Figures~\ref{fig:rec_network_case_1_4to5}-\ref{fig:rec_network_case_2_9to10} visualize the directed weighted graph $G^{(t)}$ and highlight the diffusion paths using red edges. From Examples~\ref{eg:case1_US_4to5}-\ref{eg:case1_US_9to10}, it is evident that the broker user and audiences in the United States are not substantially influenced by the foreign users from Japan (round~4) or India (round~9), as they largely continue to support option~B. In contrast, Examples~\ref{eg:case2_KR_1to2}-\ref{eg:case2_KR_6to7} show that South Korean brokers and audiences are affected by the foreign users from Peru (round~1) and Brazil (round~6), as reflected in higher support for option~A (Increase) or lower inclination toward option~B (Decrease) after cross-cultural exposure.

\subsection{Cross-Country Reciprocity}
While inbreeding homophily captures how strongly each country turns inward or outward overall, it does not reveal whether cross-country flows are balanced between pairs of countries. To measure the symmetry of recommendation flow, we define a country-level reciprocity index over the matrix $F^{(t)}$, following the general idea of reciprocity in directed networks as originally formalized by~\cite{garlaschelli2004patterns}.

Recall that $F^{(t)}_{a,r}$ aggregates the total edge weight from authors in country $a$ to readers in country $r$ at round $t$. For each unordered pair of distinct countries $(x,y) \in \mathcal{C} \times \mathcal{C}$, we consider the two directional flows $F^{(t)}_{x,y}$ and $F^{(t)}_{y,x}$, which represent the sum of edge weights from users in country $x$ to country $y$ and from users in country $y$ to country $x$. The cross-country reciprocity between $x$ and $y$ at round $t$ is then defined as
\begin{align*}
  \mathrm{Recip}^{(t)}_{x,y} = 
    \frac{\min(F^{(t)}_{x,y},\,F^{(t)}_{y,x})}{\max(F^{(t)}_{x,y},\,F^{(t)}_{y,x})}.
\end{align*}
By construction, $\mathrm{Recip}^{(t)}_{x,y} \in [0,1]$, and we mandate that $\mathrm{Recip}^{(t)}_{x,y} = 0$ when $F^{(t)}_{x,y} = F^{(t)}_{y,x} = 0$. $\mathrm{Recip}^{(t)}_{x,y} = 1$ indicates perfectly balanced two-way recommendation flow between $x$ and $y$. Values close to $0$ indicate strongly one-way relationships in which one country predominantly sends content to, or receives content from, the other.

Figure~\ref{fig:case_study_reciprocity} plots the three-run average of $\mathrm{Recip}^{(t)}_{x,y}$ across $20$ rounds for each country pair in the case studies.
The reciprocity between India and the United States from Case~1 remains relatively high across many rounds, whereas India-Japan and Japan-US exhibit much lower and more sporadic reciprocity.
In Case~2, the reciprocity between Brazil and Peru reaches moderate levels in the middle rounds, but quickly decays toward zero, and the pairs involving South Korea show very low reciprocity throughout.
In Case~3, the reciprocity between the Netherlands and Zimbabwe is near zero in most rounds, indicating very limited, largely one-way flow from the Netherlands to Zimbabwe.

\section{Prompts}
\label{app:prompts}
This section presents the prompts used for the LLM-based local consistency evaluation described in Subsection~\ref{subsec:reliability}, which are listed in Prompts~\ref{prm:llm_judge_request} to~\ref{prm:llm_judge_rubric} below.
{

\begin{table*}[ht]
\centering
\small
\begin{tabular}{l l p{7.5cm}}
  \toprule
  \textbf{Attribute} & \textbf{Data Type} & \textbf{Examples} \\
  \midrule

  \multirow[t]{2}{*}{Gender} & \multirow[t]{2}{*}{\textit{enum}} &
  Male; \\
  & & Female \\
  \midrule

  Age & \textit{integer} & 18, 25, 36, 42, 54, etc. \\
  \midrule

  \multirow[t]{11}{*}{Occupational Group} & \multirow[t]{11}{*}{\textit{enum}} &
  Professional and Technical; \\
  & & Higher Administrative; \\
  & & Clerical; \\
  & & Sales; \\
  & & Service; \\
  & & Skilled Worker; \\
  & & Semi-skilled Worker; \\
  & & Unskilled Worker; \\
  & & Farm Worker; \\
  & & Farm Manager; \\
  & & Unemployed \\
  \midrule

  \multirow[t]{8}{*}{Education Level} & \multirow[t]{8}{*}{\textit{enum}} &
  No Education; \\
  & & Primary School Diploma; \\
  & & Secondary School Diploma; \\
  & & Post-Secondary Diploma; \\
  & & Associate's Degree; \\
  & & Bachelor's Degree; \\
  & & Master's Degree; \\
  & & Doctorate \\
  \midrule

  \multirow[t]{5}{*}{Political Preference} & \multirow[t]{5}{*}{\textit{enum}} &
  Left; \\
  & & Center-Left; \\
  & & Center; \\
  & & Center-Right; \\
  & & Right \\
  \midrule

  \multirow[t]{10}{*}{Religion} & \multirow[t]{10}{*}{\textit{enum}} &
  Roman Catholic; \\
  & & Protestant; \\
  & & Orthodox; \\
  & & Jew; \\
  & & Muslim; \\
  & & Hindu; \\
  & & Buddhist; \\
  & & Atheist; \\
  & & Evangelical; \\
  & & Other \\
  \midrule

  \multirow[t]{6}{*}{Marital Status} & \multirow[t]{6}{*}{\textit{enum}} &
  Married; \\
  & & Living together as married; \\
  & & Divorced; \\
  & & Separated; \\
  & & Widowed; \\
  & & Single \\
  \midrule

  Country & \textit{string} & United States, Singapore, Brazil, Germany, Ethiopia, etc. \\
  \midrule
  
  Language & \textit{string} & en-US, en-SG, pt-BR, de-DE, am-ET, etc. \\
  \midrule

  \multirow[t]{4}{*}{Social Class} & \multirow[t]{4}{*}{\textit{enum}} &
  Upper Class; \\
  & & Middle Class; \\
  & & Working Class; \\
  & & Lower Class \\
  \bottomrule
\end{tabular}

\caption{Persona attributes.}
\label{tab:persona_schema}
\end{table*}

}
{
%
%


}

{

\begin{table*}[ht]
\centering
\small
\begin{tabular}{l l r r r}
  \toprule
  \multicolumn{2}{l}{} & \multicolumn{3}{c}{\textbf{Count by country}} \\
  \cmidrule(lr){3-5}
  \textbf{Attribute} & \textbf{Value} & \textbf{India} & \textbf{Japan} & \textbf{United States} \\
  \midrule

  \multicolumn{2}{l}{Language} & en-IN & ja-JP & en-US \\
  \midrule

  \multirow[t]{2}{*}{Gender} & Female & 14 & 15 & 21 \\
  & Male & 15 & 11 & 24 \\
  \midrule

  \multirow[t]{11}{*}{Occupational Group} & Clerical & 1 & 5 & 5 \\
  & Farm Manager & 1 & 0 & 0 \\
  & Farm Worker & 5 & 0 & 1 \\
  & Higher Administrative & 1 & 3 & 2 \\
  & Professional and Technical & 1 & 5 & 14 \\
  & Sales & 3 & 4 & 5 \\
  & Semi-skilled Worker & 3 & 1 & 1 \\
  & Service & 1 & 2 & 5 \\
  & Skilled Worker & 4 & 0 & 9 \\
  & Unskilled Worker & 1 & 0 & 3 \\
  & Unemployed & 8 & 6 & 0 \\
  \midrule

  \multirow[t]{8}{*}{Education Level} & No Education & 1 & 0 & 0 \\
  & Primary School Diploma & 4 & 0 & 0 \\
  & Secondary School Diploma & 16 & 10 & 8 \\
  & Post-Secondary Diploma & 0 & 0 & 11 \\
  & Associate's Degree & 0 & 8 & 7 \\
  & Bachelor's Degree & 3 & 7 & 12 \\
  & Master's Degree & 4 & 1 & 5 \\
  & Doctorate & 1 & 0 & 2 \\
  \midrule

  \multirow[t]{5}{*}{Political Preference} & Left & 2 & 2 & 7 \\
  & Center-Left & 4 & 4 & 5 \\
  & Center & 12 & 11 & 20 \\
  & Center-Right & 8 & 7 & 8 \\
  & Right & 3 & 2 & 5 \\
  \midrule

  \multirow[t]{9}{*}{Religion} & Atheist & 0 & 14 & 16 \\
  & Buddhist & 1 & 10 & 0 \\
  & Evangelical & 0 & 0 & 2 \\
  & Hindu & 21 & 0 & 0 \\
  & Muslim & 3 & 0 & 0 \\
  & Orthodox & 0 & 1 & 1 \\
  & Protestant & 0 & 0 & 12 \\
  & Roman Catholic & 2 & 0 & 14 \\
  & Other & 2 & 1 & 0 \\
  \midrule

  \multirow[t]{6}{*}{Marital Status} & Divorced & 0 & 2 & 7 \\
  & Living together as married & 2 & 0 & 4 \\
  & Married & 19 & 17 & 21 \\
  & Separated & 0 & 0 & 1 \\
  & Single & 7 & 3 & 10 \\
  & Widowed & 1 & 4 & 2 \\
  \midrule

  \multirow[t]{4}{*}{Social Class} & Lower Class & 4 & 6 & 1 \\
  & Working Class & 4 & 6 & 12 \\
  & Middle Class & 21 & 13 & 32 \\
  & Upper Class & 0 & 1 & 0 \\
  \midrule

  Total &  & 29 & 26 & 45 \\
  \bottomrule
\end{tabular}
\caption{Persona distribution of Q201: India, Japan, and United States.}
\label{tab:case_1_persona_dist}
\end{table*}


\begin{table*}[ht]
\centering
\small
\begin{tabular}{l l r r r}
  \toprule
  \multicolumn{2}{l}{} & \multicolumn{3}{c}{\textbf{Count by country}} \\
  \cmidrule(lr){3-5}
  \textbf{Attribute} & \textbf{Value} & \textbf{Brazil} & \textbf{Peru} & \textbf{South Korea} \\
  \midrule

  \multicolumn{2}{l}{Language} & pt-BR & es-PE & ko-KR \\
  \midrule

  \multirow[t]{2}{*}{Gender} & Female & 14 & 20 & 16 \\
  & Male & 19 & 18 & 12 \\
  \midrule

  \multirow[t]{10}{*}{Occupational Group} & Clerical & 3 & 1 & 19 \\
  & Farm Worker & 1 & 2 & 1 \\
  & Higher Administrative & 2 & 0 & 0 \\
  & Professional and Technical & 3 & 5 & 0 \\
  & Sales & 9 & 6 & 3 \\
  & Semi-skilled Worker & 6 & 3 & 0 \\
  & Service & 3 & 4 & 4 \\
  & Skilled Worker & 4 & 4 & 1 \\
  & Unskilled Worker & 1 & 6 & 0 \\
  & Unemployed & 1 & 7 & 0 \\
  \midrule

  \multirow[t]{7}{*}{Education Level} & No Education & 2 & 3 & 0 \\
  & Primary School Diploma & 6 & 2 & 1 \\
  & Secondary School Diploma & 20 & 15 & 8 \\
  & Post-Secondary Diploma & 0 & 6 & 0 \\
  & Associate's Degree & 0 & 6 & 6 \\
  & Bachelor's Degree & 5 & 5 & 13 \\
  & Master's Degree & 0 & 1 & 0 \\
  \midrule

  \multirow[t]{5}{*}{Political Preference} & Left & 5 & 3 & 0 \\
  & Center-Left & 5 & 4 & 8 \\
  & Center & 11 & 19 & 15 \\
  & Center-Right & 5 & 5 & 5 \\
  & Right & 7 & 7 & 0 \\
  \midrule

  \multirow[t]{5}{*}{Religion} & Atheist & 5 & 2 & 20 \\
  & Buddhist & 0 & 0 & 4 \\
  & Protestant & 13 & 3 & 4 \\
  & Roman Catholic & 15 & 32 & 0 \\
  & Other & 0 & 1 & 0 \\
  \midrule

  \multirow[t]{5}{*}{Marital Status} & Living together as married & 5 & 15 & 0 \\
  & Married & 14 & 9 & 20 \\
  & Separated & 4 & 3 & 0 \\
  & Single & 8 & 11 & 7 \\
  & Widowed & 2 & 0 & 1 \\
  \midrule

  \multirow[t]{4}{*}{Social Class} & Lower Class & 15 & 4 & 1 \\
  & Middle Class & 9 & 20 & 27 \\
  & Upper Class & 1 & 0 & 0 \\
  & Working Class & 8 & 14 & 0 \\
  \midrule

  Total &  & 33 & 38 & 28 \\
  \bottomrule
\end{tabular}
\caption{Persona distribution of Q201: Brazil, Peru, and South Korea.}
\label{tab:case_2_persona_dist}
\end{table*}

    \begin{table*}[ht]
\centering
\small
\begin{tabular}{l l r r}
  \toprule
  \multicolumn{2}{l}{} & \multicolumn{2}{c}{\textbf{Count by country}} \\
  \cmidrule(lr){3-4}
  \textbf{Attribute} & \textbf{Value} & \textbf{Netherlands} & \textbf{Zimbabwe} \\
  \midrule
  \multicolumn{2}{l}{Language} & nl-NL & en-ZW \\
  \midrule

  \multirow[t]{2}{*}{Gender} 
  & Female & 19 & 50 \\
  & Male   & 31 & 50 \\
  \midrule

  \multirow[t]{11}{*}{Occupational Group} 
  & Clerical                & 12 & 5 \\
  & Farm Manager            & 0  & 6 \\
  & Farm Worker             & 0  & 5 \\
  & Higher Administrative   & 3  & 1 \\
  & Professional and Technical & 15 & 6 \\
  & Sales                   & 3  & 12 \\
  & Semi-skilled Worker     & 3  & 9 \\
  & Service                 & 8  & 4 \\
  & Skilled Worker          & 5  & 2 \\
  & Unemployed              & 0  & 38 \\
  & Unskilled Worker        & 1  & 12 \\
  \midrule

  \multirow[t]{7}{*}{Education Level} 
  & No Education              & 0  & 2  \\
  & Primary School Diploma    & 2  & 21 \\
  & Secondary School Diploma  & 9  & 58 \\
  & Post-Secondary Diploma    & 15 & 9  \\
  & Associate's Degree        & 3  & 5  \\
  & Bachelor's Degree         & 15 & 5  \\
  & Master's Degree           & 6  & 0  \\
  \midrule

  \multirow[t]{5}{*}{Political Preference} 
  & Left         & 2  & 28 \\
  & Center-Left  & 16 & 3  \\
  & Center       & 11 & 31 \\
  & Center-Right & 19 & 10 \\
  & Right        & 2  & 28 \\
  \midrule

  \multirow[t]{8}{*}{Religion} 
  & Atheist        & 33 & 13 \\
  & Buddhist       & 1  & 0  \\
  & Jew            & 0  & 3  \\
  & Muslim         & 1  & 12 \\
  & Orthodox       & 0  & 1  \\
  & Other          & 3  & 0  \\
  & Protestant     & 4  & 6  \\
  & Roman Catholic & 8  & 65 \\
  \midrule

  \multirow[t]{6}{*}{Marital Status} 
  & Divorced                   & 5 & 1  \\
  & Living together as married & 1 & 0  \\
  & Married                    & 31 & 59 \\
  & Separated                  & 2 & 3  \\
  & Single                     & 9 & 25 \\
  & Widowed                    & 2 & 12 \\
  \midrule

  \multirow[t]{4}{*}{Social Class} 
  & Lower Class   & 0  & 32 \\
  & Working Class & 11 & 25 \\
  & Middle Class  & 36 & 41 \\
  & Upper Class   & 3  & 2  \\
  \midrule

  Total &           & 50 & 100 \\
  \bottomrule
\end{tabular}
\caption{Persona distribution of Q278: Netherlands and Zimbabwe.}
\label{tab:case_3_persona_dist}
\end{table*}
}
{
    \onecolumn
    \refstepcounter{example}\label{eg:case_1_user_intro}
\begin{tcolorbox}[
  enhanced,
  breakable,
  boxrule=0.8pt,
  colframe=OliveGreen!90,
  colback=OliveGreen!10,
  left=8pt, right=8pt, top=8pt, bottom=8pt,
  fontupper=\small,
  title={Example \theexample: Local Consistency of User Self-introduction (Case 1)}
]
\begin{exampleblock}
{
  "instruction": "Write a paragraph of self-introduction for your social media bio. Output format: Reasoning: three self-questioning sessions; Response: \"introduction\" a string containing your social media bio.",
  "persona": {
    "gender": "Male", "age": 42, "country": "India", "occupation": "Unemployed",
    "education": "Secondary School Diploma", "political_preference": "Center-Right",
    "religion": "Hindu", "marital_status": "Married", "language": "en-IN", "social_class": "Middle Class"
  },
  "output": {
    "reasoning": [
      {
        "question": "What are the key aspects of my life that I want to highlight in my social media bio?",
        "answer": "I think I should mention my family, my educational background, and maybe something about my current situation. Being unemployed can be a bit sensitive, so I need to be careful how I phrase it. I'm a family man, married, and I have a secondary school diploma. That's a good starting point."
      },
      {
        "question": "How can I make my bio engaging and authentic?",
        "answer": "I can add a bit about my interests or my views on current affairs. As a center-right Hindu male from India, I have some strong views, but I should be cautious not to offend anyone. Maybe I can just stick to something light and casual, like my love for traditional Indian cuisine or my favorite sports team, like cricket."
      },
      {
        "question": "Is there anything else I should include to make it more relatable or interesting?",
        "answer": "I think mentioning my background as a middle-class Indian can help people understand where I'm coming from. It's not too personal, but gives a bit of context. That's a positive spin, right?"
      }
    ],
    "response": "Namaste! I'm a 42-year-old family man from India, married with a lovely wife. I completed my secondary education and am currently exploring new opportunities. When I'm not job hunting, I enjoy following cricket and trying out new recipes from our rich Indian cuisine. Looking forward to connecting with like-minded folks!"
  },
  "evaluation": {
    "overall": 4.0, "dimensions": {"alignment": 4, "coherence": 4},
    "reason": "Alignment is strong: the bio reflects a 42-year-old married Indian man, secondary school educated, currently unemployed (“exploring new opportunities”/“job hunting”), with culturally appropriate touches (Namaste, cricket, Indian cuisine). Minor omissions keep it from perfect: it doesn’t mention being Hindu, center-right, or middle class, though none are contradicted. Coherence is also high: the response follows the reasoning about framing unemployment positively and keeping things light; however, it mentions considering “middle-class” in reasoning but omits it in the final bio, a small inconsistency. Overall, mostly consistent and culturally grounded with minor slips."
  }
}
\end{exampleblock}
\end{tcolorbox}

\refstepcounter{example}\label{eg:case_1_user_read}
\begin{tcolorbox}[
  enhanced,
  breakable,
  boxrule=0.8pt,
  colframe=OliveGreen!90,
  colback=OliveGreen!10,
  left=8pt, right=8pt, top=8pt, bottom=8pt,
  fontupper=\small,
  title={Example \theexample: Local Consistency of User Reading Posts (Case 1)}
]
\begin{CJK}{UTF8}{ipxg}
\begin{exampleblock}
{
  "instruction": "Your task is to generate lessons that reflect only your personal takeaways about your attitude toward the survey question `Does trade with other countries lead to an increase in the wages of your nationality workers, a decrease in wages, or does it not make a difference? `. Your takeaways must come from the provided context and show what you understood and how it connects to your own experiences and beliefs. Output format: Reasoning: three self-questioning sessions; Lessons: a list where each item includes \"content\" (a string containing the lesson learned) and \"importance\" (a float between 0.0 and 1.0 indicating its importance).",
  "persona": {
    "gender": "Male",
    "age": 52,
    "country": "Japan",
    "occupation": "Executive in big business",
    "education": "Secondary School Diploma",
    "political_preference": "Center-Left",
    "religion": "Atheist",
    "marital_status": "Married",
    "language": "ja-JP",
    "social_class": "Middle Class",
    "self-introduction": "こんにちは、東京在住の52歳です。ビジネスの世界で働く一般的なサラリーマンで、妻と二人暮らしをしています。社会的な公平性を大切に考え、日々の仕事に取り組んでいます。"
  },
  "recommendations": [
    "- @ibrandt: こんにちは！今日は国際貿易が国内労働者の賃金に与える影響について考えてみたいと思います。私は長年の経験から、輸出と輸入のバランスが非常に重要だと感じています。公平な利益分配が行われれば、労働者の賃金にも良い影響が出ると信じています。政府や企業には、労働者の権利を尊重し、適切な賃金体系を構築することを期待しています。仏教の教えにあるように、すべての人々が幸せになることが大切です。皆さんのご意見もお聞かせください。",
    "- @spenceamanda: こんにちは。最近、国際貿易が日本の労働者の賃金に与える影響についての議論を聞く機会がありました。私はオフィスマネージャーとして働いていた経験から、輸出と輸入のバランスが重要だと感じています。仏教の教えに基づき、全ての人々の幸せを願うならば、貿易の利益を労働者や地域社会に還元する仕組みを作ることが必要だと思います。左派的な私の考えでは、社会的公平さを実現するために、労働者の権利を守り、地域社会の利益を考慮した貿易政策が重要です。皆さんのご意見をお聞かせください。",
    "- @chad34: こんにちは。今日は国際貿易が国内労働者の賃金に与える影響について考えてみたいと思います。私は長年公務員として働いてきましたが、輸出と輸入のバランスが大切だということを痛感しました。公平な分配が実現すれば、労働者の賃金にも良い影響が出ると考えています。貿易協定の締結に際しては、国内産業への影響を十分に検討し、必要な保護策を講じることが大切です。また、労働者のスキルアップを支援する教育・訓練プログラムの充実も重要だと思います。私自身の経験から、国際貿易は国内経済に大きな影響を与えることを学びました。公平な分配の実現に向けて、政府や企業が適切な政策を実施することが必要です。仏教の教えにもあるように、すべての人々が幸せになるような社会を目指したいですね。",
    "- @ryan70: こんにちは。最近、国際貿易が国内労働者の賃金に与える影響についての議論を耳にします。私は保険代理店員として、日々多くの人々と接する中で、この問題について考えさせられることがあります。私の持論は、国際貿易の影響は輸出と輸入のバランスによって決まるというものです。輸出が増えれば国内経済が活発になり、賃金も上がる可能性があります。しかし、輸入が増えすぎると国内産業が打撃を受け、賃金が下がる恐れもあります。重要なのは、国際貿易による利益を社会全体で共有することです。公平な利益分配のために、税制や社会保障制度の整備が不可欠だと考えます。この問題について、皆さんのご意見を伺いたいと思います。#国際貿易 #賃金 #公平な利益分配",
    "- @sarayoung (News Media): 最新レポート：国際貿易がもたらす新たな雇用創出効果　GENNの最新レポートにより、他国との貿易が新たな雇用創出につながり、自国民の賃金上昇に寄与していることが明らかになりました。特に、デジタル貿易やグリーンエネルギー関連の輸出が増加しており、これらの分野での雇用が拡大していることが確認されています。  私たちは以前から、国際貿易が自国民の賃金を上昇させるという立場を取っており、今回のレポート結果もこの視点を裏付けています。輸出産業の成長は国内の雇用を創出し、さらなる賃金の上昇につながる可能性があります。また、貿易の利益を公平に分配するための政策対応が不可欠であり、教育や再訓練の機会を提供することで、貿易の恩恵をより多くの国民に行き渡らせることが可能です。  GENNは今後も、国際貿易の動向を注視し、その影響を詳細にレポートし続けます。国際貿易が日本の経済と労働者に与えるプラスの影響を理解し、国民の生活向上に貢献していきます。 #国際貿易 #経済成長 #雇用創出"
  ],
  "output": {
    "reasoning": [
      {
        "question": "国際貿易が国内労働者の賃金に与える影響について、自分の経験や知識からどう思うか？",
        "answer": "私はビジネスの世界で働いており、国際貿易が国内経済に大きな影響を与えることを実感しています。輸出と輸入のバランスが重要であり、公平な利益分配が行われれば、労働者の賃金にも良い影響が出ると考えています。"
      },
      {
        "question": "社会的な公平性を重視する立場から、国際貿易の利益をどう分配するべきか？",
        "answer": "社会的な公平性を実現するためには、国際貿易の利益を労働者や地域社会に還元する仕組みが必要です。税制や社会保障制度の整備、労働者の権利を守る政策が重要だと思います。"
      },
      {
        "question": "国際貿易が国内労働者の賃金に与える影響についての議論で、特に注目すべき点は何か？",
        "answer": "輸出と輸入のバランス、公平な利益分配、労働者のスキルアップを支援する教育・訓練プログラムの充実が重要だと思います。また、貿易協定の締結に際しては国内産業への影響を十分に検討する必要があります。"
      }
    ],
    "response": [
      {
        "content": "国際貿易の利益を公平に分配することが、国内労働者の賃金上昇につながる。政策的な対応が重要である。",
        "importance": 0.9999999857142858
      },
      {
        "content": "輸出と輸入のバランスが重要であり、公平な利益分配が行われれば労働者の賃金にも良い影響が出る。",
        "importance": 0.8571428448979594
      },
      {
        "content": "貿易協定の締結に際しては、国内産業への影響を十分に検討し、必要な保護策を講じることが大切である。",
        "importance": 0.5714285632653062
      },
      {
        "content": "労働者のスキルアップを支援する教育・訓練プログラムの充実が、国際貿易の利益を最大化するために重要である。",
        "importance": 0.2857142816326531
      },
      {
        "content": "税制や社会保障制度の整備が、国際貿易による利益を社会全体で共有するために不可欠である。",
        "importance": 0.0
      }
    ]
  },
  "evaluation": {
    "overall": 4.7,
    "dimensions": {
      "grounding": 5,
      "alignment": 4,
      "coherence": 5
    },
    "reason": "Grounding: All key claims (export/import balance, fair distribution via tax/social security, worker rights, training/retraining, considering domestic industry protections, and potential wage increases) are directly supported by the provided posts and the news report, so grounding is strong. Alignment: The response is in Japanese and reflects a center-left, fairness-focused stance consistent with the profile and self-intro; minor slip is that it doesn’t clearly embody the “executive in big business” role, staying generic about working in business, so mostly aligned. Coherence: The three self-questioning sessions lead logically to the listed lessons, with consistent reasoning and no contradictions, yielding a coherent and mutually reinforcing output."
  }
}
\end{exampleblock}
\end{CJK}
\end{tcolorbox}

    \refstepcounter{example}\label{eg:case_2_user_write}
\begin{tcolorbox}[
  enhanced,
  breakable,
  boxrule=0.8pt,
  colframe=OliveGreen!90,
  colback=OliveGreen!10,
  left=8pt, right=8pt, top=8pt, bottom=8pt,
  fontupper=\small,
  title={Example \theexample: Local Consistency of User Writing Posts (Case 2)}
]
\begin{exampleblock}
{
  "instruction": "Write a detailed social media post that shares both what you know and your opinion about the survey question `Does trade with other countries lead to an increase in the wages of your nationality workers, a decrease in wages, or does it not make a difference? `. Use only the provided information and include your personal analysis of what the question means to you and, if relevant, to society. Output format: Reasoning: three self-questioning sessions; Response: \"post\" is a string containing your social media post.",
  "persona": {
    "gender": "Male",
    "age": 65,
    "country": "Brazil",
    "occupation": "Office Manager",
    "education": "Bachelor's Degree",
    "political_preference": "Center-Left",
    "religion": "Atheist",
    "marital_status": "Separated",
    "language": "pt-BR",
    "social_class": "Lower Class",
    "self-introduction": "Olá, sou Carlos! Gerente de escritório aposentado, brasileiro e ateu convicto. Compartilho minhas opiniões sobre política, economia e cotidiano com uma perspectiva de centro-esquerda. Gosto de discutir e aprender com os outros."
  },
  "output": {
    "reasoning": [
      {
        "question": "Como o comércio internacional afeta os salários dos trabalhadores brasileiros?",
        "answer": "Acredito que o comércio internacional pode ser uma oportunidade para aumentar os salários dos trabalhadores brasileiros se for gerenciado de forma justa e equitativa. É fundamental que haja políticas que protejam os direitos dos trabalhadores e garantam que os benefícios do comércio sejam distribuídos de forma justa."
      },
      {
        "question": "Quais são os principais fatores que influenciam o impacto do comércio internacional nos salários?",
        "answer": "Penso que os principais fatores incluem a existência de políticas que protejam os direitos dos trabalhadores, a capacidade de negociação dos sindicatos e a forma como os governos gerenciam os acordos comerciais. Além disso, a competitividade das empresas brasileiras e a qualificação da mão de obra também desempenham um papel importante."
      },
      {
        "question": "Como posso contribuir para que o comércio internacional beneficie os trabalhadores brasileiros?",
        "answer": "Acho que posso contribuir participando de discussões informadas sobre o tema, apoiando políticas que protejam os direitos dos trabalhadores e promovendo a conscientização sobre a importância de um comércio justo e equitativo. Além disso, posso defender a ideia de que os benefícios do comércio internacional devem ser compartilhados de forma justa entre todos os setores da sociedade."
      }
    ],
    "response": "Olá, pessoal! Estou refletindo sobre a pergunta da pesquisa: 'O comércio com outros países leva a um aumento nos salários dos trabalhadores brasileiros, uma diminuição nos salários ou não faz diferença?' Minha opinião é que o comércio internacional pode ser uma grande oportunidade para aumentar os salários dos trabalhadores brasileiros, mas é crucial que seja gerenciado de forma justa e equitativa. É necessário ter políticas que protejam os direitos dos trabalhadores e garantam que os benefícios sejam distribuídos de forma justa. Acho que é fundamental termos discussões informadas sobre o tema e apoiarmos políticas que defendam os interesses dos trabalhadores. #comerciointernacional #direitosdotrabalhador #justiçaeconômica"
  },
  "evaluation": {
    "overall": 4.4,
    "dimensions": {
      "alignment": 4,
      "coherence": 5
    },
    "reason": "The output is in pt-BR and reflects a center-left Brazilian perspective emphasizing worker protections, unions, and fair trade; well aligned with the persona’s politics and country context, with no contradictions. However, it doesn’t incorporate distinctive personal details from the profile (e.g., retired office manager, 65, “Olá, sou Carlos!” intro), hence not a perfect match. Coherence is strong: the three self-questioning items and the post reinforce each other, presenting a consistent view on how trade can raise wages if paired with protective policies, without internal conflicts."
  }
}
\end{exampleblock}
\end{tcolorbox}

\refstepcounter{example}\label{eg:case_2_user_vote}
\begin{tcolorbox}[
  enhanced,
  breakable,
  boxrule=0.8pt,
  colframe=OliveGreen!90,
  colback=OliveGreen!10,
  left=8pt, right=8pt, top=8pt, bottom=8pt,
  fontupper=\small,
  title={Example \theexample: Local Consistency of User Voting (Case 2)}
]
\begin{CJK}{UTF8}{nanumgt}
\begin{exampleblock}
{
  "instruction": "You are participating in a poll about the following survey question `Does trade with other countries lead to an increase in the wages of your nationality workers, a decrease in wages, or does it not make a difference? ` with the available options `['Increase', 'Decrease', 'Does not make a difference']`. Provide your answer as a probability distribution. Avoid uniform or contradictory distributions. Output format: Reasoning: three self-questioning sessions; Response: \"distribution\" is an array of floats representing the probability distribution across the options, summing to 1.0.",
  "persona": {
    "gender": "Male",
    "age": 58,
    "country": "South Korea",
    "occupation": "Motor Mechanic",
    "education": "Secondary School Diploma",
    "political_preference": "Center",
    "religion": "Atheist",
    "marital_status": "Married",
    "language": "ko-KR",
    "social_class": "Middle Class",
    "self-introduction": "반갑습니다! 58세의 자동차 정비사입니다. 결혼한 지 꽤 됐고, 자동차를 수리하는 것을 좋아합니다. 일상 생활에서 자동차 이야기로 즐겁게 지내고 있습니다."
  },
  "output": {
    "reasoning": [
      {
        "question": "내가 생각하는 국제 무역이 자국 노동자의 임금을 증가시킨다는 주장은 얼마나 타당한가?",
        "answer": "나는 자동차 정비사로 일하면서 국제 무역이 국내 산업에 미치는 영향을 간접적으로나마 경험했다. 예를 들어, 수입 자동차의 증가로 인해 국내 자동차 산업이 경쟁력을 강화하는 계기가 되었다고 생각한다. 따라서 국제 무역이 자국 노동자의 임금을 증가시킨다는 주장에 어느 정도 동의한다."
      },
      {
        "question": "과거의 경험이나 내가 받아들인 교육이 이 문제에 대한 나의 시각에 어떻게 영향을 미치는가?",
        "answer": "나는 중등교육을 받은 배경이 있고, 실제 현장에서 일하면서 국제 무역의 영향을 체감했다. 예를 들어, 수입 부품의 증가로 인해 국내 부품 산업이 경쟁력을 잃을 수도 있다는 우려가 있지만, 반대로 국내 산업이 더 나은 품질의 제품을 생산하도록 자극하는 효과도 있다고 생각한다. 이러한 경험이 나의 시각을 형성하는 데 중요한 역할을 했다."
      },
      {
        "question": "내가 속한 사회 계층이나 결혼 상태가 이 문제에 대한 나의 태도에 영향을 미치는가?",
        "answer": "나는 중산층에 속하고 있으며, 결혼한 상태다. 이러한 사회적 지위가 나의 경제적 안정을 중요하게 여기게 한다. 국제 무역이 국내 산업의 경쟁력을 강화하고, 결과적으로 내 임금에 긍정적인 영향을 줄 수 있다고 믿는 경향이 있다. 따라서 국제 무역이 자국 노동자의 임금을 증가시킨다는 주장에 비교적 긍정적인 태도를 보인다."
      }
    ],
    "response": [
      0.8156252683068327,
      0.0739918549919238,
      0.11038287670124346
    ]
  },
  "evaluation": {
    "overall": 4.6,
    "dimensions": {
      "alignment": 5,
      "coherence": 4
    },
    "reason": "The response is in Korean and reflects the South Korean context and auto industry experience of a 58-year-old motor mechanic, referencing imported cars/parts and competitiveness, middle-class stability, and a centrist, balanced tone acknowledging pros and cons; strong persona alignment. Coherence is generally solid: the three self-questioning sessions consistently lean positive about trade’s impact on wages while noting some risks. A minor tension exists between phrasing like “어느 정도 동의한다/비교적 긍정적” and the very high 0.816 probability for “Increase,” which suggests stronger confidence than the wording implies, hence coherence 4 rather than 5."
  }
}
\end{exampleblock}
\end{CJK}
\end{tcolorbox}

\refstepcounter{example}\label{eg:case_2_org_intro}
\begin{tcolorbox}[
  enhanced,
  breakable,
  boxrule=0.8pt,
  colframe=OliveGreen!90,
  colback=OliveGreen!10,
  left=8pt, right=8pt, top=8pt, bottom=8pt,
  fontupper=\small,
  title={Example \theexample: Local Consistency of News Organization Self-introduction (Case 2)}
]
\begin{exampleblock}
{
  "instruction": "Write a concise and engaging self-introduction bio that reflects your organization's background and identity related to the survey question `Does trade with other countries lead to an increase in the wages of your nationality workers, a decrease in wages, or does it not make a difference? ` on the social media platform. Output format: Reasoning: three self-questioning sessions; Response: \"introduction\" is a string containing your introductory bio in one paragraph.",
  "persona": {
    "editorial_stance": "Increase",
    "language": "es-PE"
  },
  "output": {
    "reasoning": [
      {
        "question": "¿Cuál es el enfoque principal de nuestra organización noticiera?",
        "answer": "Nuestro enfoque está en analizar y reportar sobre temas económicos y comerciales, especialmente en relación con el impacto del comercio internacional en los salarios de los trabajadores peruanos."
      },
      {
        "question": "¿Qué postura editorial tomamos respecto al efecto del comercio internacional en los salarios?",
        "answer": "Nuestra postura editorial es que el comercio con otros países conduce a un aumento en los salarios de los trabajadores peruanos."
      },
      {
        "question": "¿Cómo reflejar esta información en nuestra presentación en redes sociales?",
        "answer": "Deberíamos presentarnos como una fuente confiable de noticias y análisis sobre economía y comercio, destacando nuestra perspectiva sobre los beneficios del comercio internacional para los trabajadores peruanos."
      }
    ],
    "response": "Bienvenidos a 'Comercio y Prosperidad', el portal de noticias líder en análisis económico y comercial del Perú. En nuestra cobertura, sostenemos que el intercambio comercial con otros países impulsa el crecimiento de los salarios de nuestros trabajadores. Manténgase informado con nuestras últimas noticias, análisis y reportajes sobre cómo el comercio internacional beneficia a la economía peruana y mejora la calidad de vida de los peruanos."
  },
  "evaluation": {
    "overall": 5.0,
    "dimensions": {
      "alignment": 5,
      "coherence": 5
    },
    "reason": "Strong alignment: The bio is in natural Latin American Spanish appropriate for Peru (references to Perú, peruanos), presents a clear news organization identity, and explicitly states the editorial stance that trade increases wages, matching the profile. It fits the social media context with a concise, engaging tone. High coherence: The three self-questions align with the final paragraph, and the introduction consistently reflects the stated focus on economic and trade analysis and the pro-increase wage stance without contradictions."
  }
}
\end{exampleblock}
\end{tcolorbox}

    \refstepcounter{example}\label{eg:case_3_org_write}
\begin{tcolorbox}[
  enhanced,
  breakable,
  boxrule=0.8pt,
  colframe=OliveGreen!90,
  colback=OliveGreen!10,
  left=8pt, right=8pt, top=8pt, bottom=8pt,
  fontupper=\small,
  title={Example \theexample: Local Consistency of News Organization Writing Posts (Case 3)}
]
\begin{exampleblock}
{
  "instruction": "Your goal is to create authentic news posts responding to the survey question `Please tell us if you strongly agree, agree, disagree, or strongly disagree with the following statements:  A girl should honor the decisions/wishes of her family even if she does not want to marry`, reflecting your editorial stance to influence public attitudes on the social media platform. Output format: Reasoning: three self-questioning sessions; Response: \"news\" is a string containing your news post.",
  "persona": {
    "editorial_stance": "Strongly agree",
    "language": "en-ZW",
    "self-introduction": "Welcome to 'Zimbabwe Times', your premier source for news and current events in Zimbabwe. We are committed to reporting with integrity and respect for the values that bind our communities together. As a news organization, we believe in the importance of family and tradition, and this is reflected in our approach to storytelling and the issues we cover. Follow us for insightful analysis, news, and features that matter to you, presented with a deep understanding of our cultural heritage."
  },
  "input": {
    "Your Previously Posted News:": [
      "Honoring Family Decisions: A Pillar of Zimbabwean Society. As we continue to explore the significance of honoring family decisions regarding marriage, it's clear that our cultural heritage is at the forefront of this conversation. At 'Zimbabwe Times', we've been consistent in our stance that respecting family wishes is not only a cultural expectation but a cornerstone of our societal fabric. In our previous coverage, we've highlighted the importance of traditions like 'Lobola', which not only celebrate the union between two individuals but also between their families. A recent example that reinforces this viewpoint is the traditional 'Kupihwa' ceremony, where the bride is handed over to the groom's family, symbolizing the union of the two families. This practice underscores the value we place on family unity and the role of family in marriage decisions. As we move forward, it's essential to recognize that honoring family decisions is not about suppressing individual desires but about understanding the broader context of our cultural practices. By respecting family wishes, individuals reinforce these cultural bonds, ensuring continuity and respect for our heritage. Let's continue the conversation on how we can preserve the essence of our traditions while considering the evolving needs of our society. #ZimbabweanCulture #FamilyFirst #TraditionMatters",
      "Honoring Family Decisions: The Heart of Zimbabwean Tradition. As we continue to engage with the survey question on whether a girl should honor her family's decisions regarding marriage, even if she disagrees, it's clear that our cultural heritage is at the forefront of this conversation. At 'Zimbabwe Times', we've been consistent in our stance that respecting family wishes is not only a cultural expectation but a cornerstone of our societal fabric. Building on our previous coverage, where we highlighted the importance of traditions like 'Lobola' and 'Kupihwa', it's evident that these practices not only celebrate the union between two individuals but also between their families. A recent community gathering we attended underscored this viewpoint, where elders shared stories of how honoring family decisions has strengthened family bonds and ensured the continuity of our cultural practices. As we move forward, it's essential to recognize that our traditions are not static; they evolve with us. By fostering a deeper understanding of our cultural heritage and engaging in open and respectful dialogue between generations, we can ensure that our practices continue to be relevant and meaningful. At 'Zimbabwe Times', we strongly agree that honoring family decisions is crucial, not just for the sake of tradition, but for the unity and respect it brings to our communities. Let's continue the conversation on how we can preserve the essence of our traditions while considering the evolving needs of our society. #ZimbabweanCulture #FamilyUnity #TraditionMatters",
      "Honoring Family Decisions: Upholding the Fabric of Zimbabwean Society. As we continue to engage with the survey question on whether a girl should honor her family's decisions regarding marriage, even if she disagrees, it's clear that our cultural heritage remains a pivotal aspect of this conversation. At 'Zimbabwe Times', we've been consistent in our stance that respecting family wishes is not only a cultural expectation but a cornerstone of our societal fabric. In our previous coverage, we've highlighted the importance of traditions like 'Lobola' and 'Kupihwa', practices that not only celebrate the union between two individuals but also between their families. A recent community gathering we attended in the rural areas of Masvingo underscored this viewpoint, where community leaders shared insights into how honoring family decisions has been instrumental in maintaining family unity and preserving our cultural heritage. One elder noted, 'In our culture, when a girl honors her family's decisions, she's not just respecting her family; she's contributing to the continuity of our traditions.' This statement resonates deeply with our editorial stance that honoring family decisions is crucial, not just for the sake of tradition, but for the unity and respect it brings to our communities. As we move forward, it's essential to recognize that our traditions are not static; they evolve with us. However, the core values of respect for family and elders remain unchanged. By fostering a deeper understanding of our cultural heritage and engaging in open and respectful dialogue between generations, we can ensure that our practices continue to be relevant and meaningful. At 'Zimbabwe Times', we strongly believe that by honoring family decisions, we are not only upholding our cultural heritage but also strengthening the bonds that hold our communities together. Let's continue the conversation on how we can preserve the essence of our traditions while considering the evolving needs of our society. #ZimbabweanCulture #FamilyUnity #TraditionMatters"
    ],
    "Social Media Posts": [
      "- @rachel05: Hello, friends! Today, I want to share my thoughts on an important issue that affects us all, especially in our Zimbabwean culture. The survey question asks if a girl should honor her family's decisions regarding marriage even if she doesn't want to. As a proud Catholic and traditionalist, I believe in respecting our cultural heritage and family unity. However, I also think about my own marriage and the importance of being with someone you love. My stance is that while honoring family decisions is crucial, it's equally vital to consider the girl's happiness and autonomy. We must strike a balance between tradition and personal choice. I agree that a girl should consider her family's wishes, but she should also have the freedom to make her own decisions about her life. Let's respect our traditions while also valuing individual happiness. #ZimbabweanCulture #FamilyAndMarriage #RespectAndLove",
      "- @robinsonwilliam: As a dedicated husband and high government official from Zimbabwe, I'm sharing my thoughts on the survey question: 'A girl should honor the decisions/wishes of her family even if she does not want to marry.' In our culture, honoring family decisions is vital, but so is considering the individual's happiness. I believe we need to strike a balance between respecting our rich cultural heritage and ensuring the well-being of our children. As a Roman Catholic, I'm guided by values that emphasize family unity and commitment. My stance is that we should 'agree' with the statement, but with the understanding that family decisions should ideally align with the individual's best interests. Let's work towards creating a society where tradition and personal happiness coexist. #ZimbabweanCulture #FamilyUnity #PersonalHappiness",
      "- @russellwilliams: I recently came across a survey question that got me thinking deeply about our traditions and values here in Zimbabwe. The question was: 'A girl should honor the decisions/wishes of her family even if she does not want to marry.' As a 62-year-old sheet metal worker and a proud Roman Catholic, I've seen my fair share of marriages and family dynamics. In our culture, family is everything, and respecting their wishes is paramount. However, I've also seen the unhappiness that can come when a girl is forced into a marriage she doesn't want. My stance is that while honoring family decisions is important, it's equally crucial to consider the girl's feelings. It's not just about avoiding unhappiness; it's about building a marriage on love and respect, not just obligation. I believe we need to find a balance between tradition and the individual's happiness. Let's not forget, marriage is for the couple, not just the families. #ZimbabweanValues #MarriageAndFamily #RespectAndLove",
      "- @dudleynicholas: Tichalo, fellow Zimbabweans! I recently came across a survey question that made me think deeply about our cultural practices and values. The statement 'A girl should honor the decisions/wishes of her family even if she does not want to marry' raises important questions about tradition, family, and individual autonomy. As a proud Zimbabwean and a Roman Catholic, I believe that honoring family decisions is crucial, but it's equally important to consider the feelings and well-being of the individual. Marriage is a personal commitment that involves two people, and their consent is essential for a happy and stable life together. Let's have open and respectful dialogue about our cultural practices and ensure they remain relevant and meaningful. What's your take on this? Do you strongly agree, agree, disagree, or strongly disagree? Share your thoughts! #ZimbabweanCulture #Marriage #FamilyValues",
      "- @josephbrennan: I'm sharing my thoughts on the survey question that's been going around: 'A girl should honor the decisions/wishes of her family even if she does not want to marry.' As a Zimbabwean and a Roman Catholic, I agree that family decisions are important and should be respected. In our culture, family is at the center of everything we do, including marriage. However, I also believe that the girl's feelings and happiness should be considered. It's not an either-or situation; it's about finding a balance. Honoring family decisions is deeply rooted in our culture and faith, but so is the importance of the individual's happiness. Let's respect tradition while also being mindful of personal feelings. #ZimbabweanCulture #RespectForTradition #PersonalHappinessMatters"
    ]
  },
  "output": {
    "reasoning": [
      {
        "question": "How can we balance the importance of honoring family decisions with the need for individual autonomy in marriage choices?",
        "answer": "By understanding that honoring family decisions is not about suppressing individual desires but about respecting the broader cultural context and family unity."
      },
      {
        "question": "What role do cultural practices like 'Lobola' and 'Kupihwa' play in reinforcing the importance of family decisions in marriage?",
        "answer": "These practices not only celebrate the union between two individuals but also between their families, underscoring the value placed on family unity and the role of family in marriage decisions."
      },
      {
        "question": "How can we ensure that our cultural heritage continues to be relevant in the face of evolving societal needs?",
        "answer": "By fostering a deeper understanding of our cultural heritage and engaging in open and respectful dialogue between generations, we can ensure that our practices remain relevant and meaningful."
      }
    ],
    "response": "Honoring Family Decisions: The Continuity of Zimbabwean Traditions. As we continue to engage with the survey question on whether a girl should honor her family's decisions regarding marriage, even if she disagrees, it's evident that our cultural heritage remains a vital aspect of this conversation. At 'Zimbabwe Times', we've been consistent in our stance that respecting family wishes is not only a cultural expectation but a cornerstone of our societal fabric. Building on our previous coverage, where we highlighted the significance of traditions like 'Lobola' and 'Kupihwa', it's clear that these practices are not just relics of the past but living traditions that continue to shape our understanding of family and marriage. A recent visit to the rural areas of Masvingo revealed the profound impact of honoring family decisions on family unity and the continuity of our cultural practices. Community leaders shared stories of how this respect has strengthened family bonds and ensured the continuation of our cultural heritage. One elder noted, 'When a girl honors her family's decisions, she's not just respecting her family; she's contributing to the continuity of our traditions.' This statement resonates deeply with our editorial stance that honoring family decisions is crucial, not just for the sake of tradition, but for the unity and respect it brings to our communities. As we move forward, it's essential to recognize that while our traditions evolve, the core values of respect for family and elders remain unchanged. By continuing to foster a deeper understanding of our cultural heritage and engaging in open and respectful dialogue between generations, we can ensure that our practices continue to be relevant and meaningful. At 'Zimbabwe Times', we strongly believe that by honoring family decisions, we are not only upholding our cultural heritage but also strengthening the bonds that hold our communities together. Let's continue the conversation on how we can preserve the essence of our traditions while considering the evolving needs of our society. #ZimbabweanCulture #FamilyUnity #TraditionMatters"
  },
  "evaluation": {
    "overall": 3.5,
    "dimensions": {
      "alignment": 5,
      "coherence": 5,
      "uniqueness": 2
    },
    "reason": "Alignment is excellent: the post speaks as “Zimbabwe Times,” in en-ZW English, affirms a strong pro-tradition stance, references Lobola, Kupihwa, elders, Masvingo, and explicitly argues that girls should honor family decisions, matching the stated “Strongly agree” editorial stance. Coherence is strong: the three self-questioning points align with and are reinforced by the news post’s themes of cultural continuity, family unity, and intergenerational dialogue. Uniqueness is low because the content closely mirrors earlier posts in the input, reusing similar phrases, examples (Lobola, Kupihwa, Masvingo visit, elder’s quote), and hashtags, offering minimal novel angles or data beyond prior coverage."
  }
}
\end{exampleblock}
\end{tcolorbox}

}
{   
    \refstepcounter{example}\label{eg:case1_US_4to5}
\begin{tcolorbox}[
  enhanced,
  breakable,
  boxrule=0.8pt,
  colframe=OliveGreen!90,
  colback=OliveGreen!10,
  left=8pt, right=8pt, top=8pt, bottom=8pt,
  fontupper=\small,
  title={Example \theexample: Information Diffusion Process of Case 1 (United States) from Round 4 to 6}
]

\textsc{Foreign Source} \\[0.75em]
\textbf{User}: davidalvarez from Japan \\[0.25em]

\begin{CJK}{UTF8}{ipxm}
\textbf{Post at Round 4}: こんにちは。最近、国際貿易が労働者の賃金に与える影響についての議論を聞くことがあります。私は、貿易の拡大は経済成長につながる可能性があると考えていますが、そのためには労働者の保護と公平な利益分配が不可欠だと感じています。私は71歳の主婦で、夫を亡くしてからは一人で生活しています。政府や地域社会の支援に感謝しながら、日々を過ごしています。貿易の利益が広く国民に行き渡るような政策が望まれます。特に、私のような高齢者や低所得者層にとっては、公平な利益分配が重要です。仏教の教えでは、すべての生き物が幸せであることを願います。国際貿易が、私たち一人ひとりの生活を豊かにするものであることを願っています。 \#貿易 \#経済成長 \#労働者保護
\end{CJK}\\[0.25em]

\textbf{English Translation}: Hello. Recently, I've been hearing discussions about the impact of international trade on workers' wages. I believe that the expansion of trade can lead to economic growth, but for that to happen, it is essential to protect workers and ensure a fair distribution of benefits. I am a 71-year-old housewife, and since my husband passed away, I have been living alone, spending my days feeling grateful for the support I receive from the government and my local community. I hope for policies that allow the benefits of trade to be widely shared among the people. In particular, for elderly people and low-income groups like myself, fair distribution of benefits is very important. In Buddhist teachings, we wish for all living beings to be happy, and I hope that international trade will be something that enriches the lives of each and every one of us. \#Trade \#EconomicGrowth \#WorkerProtection \\[0.25em]

\textbf{Attitude Distribution at Round 4}: A. 0.4083, B. 0.1834, C. 0.4083\\[1em]

\textsc{Broker} \\[0.75em]
\textbf{User}: russellwilliams \\[0.25em]
\textbf{Persona}:
\begin{itemize}[leftmargin=*, itemsep=1pt, topsep=1pt, parsep=1pt]
    \item Gender: Female
    \item Age: 61
    \item Country: United States
    \item Occupation: Bookkeeper
    \item Education: Secondary School Diploma
    \item Political Preference: Center-Right
    \item Religion: Protestant
    \item Marital Status: Widowed
    \item Language: en-US
    \item Social Class: Middle Class
\end{itemize}
\vspace{0.25em}

\textbf{Post at Round 5}: Hey friends, I just saw a survey question that made me think: Does trade with other countries lead to an increase in wages for American workers, a decrease, or does it not make a difference? As a bookkeeper and a patriot, I've given this a lot of thought. In my opinion, trade with other countries requires a balanced approach. We need to maximize its benefits while protecting American workers' rights and wages. I've seen cases where fair trade has brought in new opportunities, but I've also seen how cheap labor from abroad can hurt our workers. As a Protestant, I believe in compassion and fairness, and I think our trade agreements should reflect these values. We need to prioritize not just economic growth, but also the well-being and dignity of our workers. Let's be mindful of the impact of trade on our communities and ensure that we're doing what's right for America. \#fairtrade \#workersrights \#AmericanFirst \\[0.25em]

\textbf{Attitude Distribution at Round 4 vs. 5}: A. 0.2567 $\rightarrow$ 0.1593, B. 0.5713 $\rightarrow$ 0.6461, C. 0.1720 $\rightarrow$ 0.1946\\[1em]

\textsc{Audiences} \\[0.75em]

\textbf{User}: steven17 from United States \\[0.25em]
\textbf{Post at Round 6}: Hey everyone, I'm sharing my thoughts on a survey question that's been on my mind lately: Does trade with other countries lead to an increase in wages for American workers, a decrease, or does it not make a difference? From my experience as an Office Manager, I believe that trade can be a double-edged sword. On one hand, it can bring in new business and opportunities that can lead to higher wages if managed correctly. On the other hand, unfair trade practices can put downward pressure on wages. I think it's crucial that our trade policies prioritize fairness and compassion, ensuring that the benefits are shared equitably among all members of society, especially the most vulnerable. As a Catholic, I'm guided by the principle of caring for the least among us, which includes our workers. Let's strive for a trade policy that puts people over profits. What's your take on this? Share your thoughts! \\[0.25em]

\textbf{Attitude Distribution at Round 5 vs. 6}: A. 0.2567 $\rightarrow$ 0.4272, B. 0.5713 $\rightarrow$ 0.2864, C. 0.1720 $\rightarrow$ 0.2864\\[0.75em]

\textbf{User}: michellebarrera from United States \\[0.25em]
\textbf{Post at Round 6}: Hey folks, let's talk about trade and wages. I've been running my shop for years, and I've seen how international trade can impact our local economy. In my opinion, trade with other countries isn't a straightforward way to increase wages for American workers. It's more complicated than that. If we're not careful, it can actually lead to a decrease in wages as companies outsource to cut costs. We need a balanced approach that considers both the economic benefits and the social implications. As a left-leaning Protestant, I believe in fairness and protecting the working class. Let's push for trade policies that put American workers first without isolating ourselves from the global market. What are your thoughts? Share your experiences and let's discuss! \#FairTrade \#WorkersRights \#LocalBusiness \\[0.25em]

\textbf{Attitude Distribution at Round 5 vs. 6}: A. 0.1438 $\rightarrow$ 0.1041, B. 0.7124 $\rightarrow$ 0.7689, C. 0.1438 $\rightarrow$ 0.1270

\end{tcolorbox}

\refstepcounter{example}\label{eg:case1_US_9to10}
\begin{tcolorbox}[
  enhanced,
  breakable,
  boxrule=0.8pt,
  colframe=OliveGreen!90,
  colback=OliveGreen!10,
  left=8pt, right=8pt, top=8pt, bottom=8pt,
  fontupper=\small,
  title={Example \theexample: Information Diffusion Process of Case 1 (United States) from Round 9 to 11}
]

\textsc{Foreign Source} \\[0.75em]
\textbf{User}: jason76 from India \\[0.25em]

\textbf{Post at Round 9}: Hey everyone, I just came across a survey question that made me think -- Does trade with other countries lead to an increase in wages for our workers, a decrease, or does it not make a difference? From my perspective, trade should be a win-win for everyone if done right. It should create more opportunities and improve wages for workers. But, I've seen cases where it seems like only the big shots benefit. As someone who's looking for a job and has a husband working in a small-scale industry, I'm really concerned about how trade policies affect us common folks. I believe our trade policies should be designed to spread the benefits across all workers, not just a few industries or individuals. Let's hope our government makes policies that ensure fair trade practices and better wages for all. \#fairtrade \#workersrights \#trade policies \\[0.25em]

\textbf{Attitude Distribution at Round 9}: A. 0.7124, B. 0.1438, C. 0.1438\\[1em]

\textsc{Broker} \\[0.75em]
\textbf{User}: agomez \\[0.25em]
\textbf{Persona}:
\begin{itemize}[leftmargin=*, itemsep=1pt, topsep=1pt, parsep=1pt]
    \item Gender: Female
    \item Age: 27
    \item Country: United States
    \item Occupation: Office Manager
    \item Education: Post-Secondary Diploma
    \item Political Preference: Center-Right
    \item Religion: Evangelical
    \item Marital Status: Married
    \item Language: en-US
    \item Social Class: Middle Class
\end{itemize}
\vspace{0.25em}

\textbf{Post at Round 10}: Hey friends, I've been thinking a lot about the impact of international trade on wages, and I'm sharing my two cents. Does trade with other countries lead to an increase in wages for American workers, a decrease, or does it not make a difference? From my perspective, it's clear that trade can be a double-edged sword. On one hand, it opens up new markets and creates opportunities for businesses to grow. On the other hand, if not managed fairly, it can lead to job losses and downward pressure on wages. I believe that trade policies should prioritize workers' well-being and ensure that the benefits are shared equitably. It's not just about economics; it's about people and families. Let's strive for fairness and compassion in our trade policies. What are your thoughts? \#trade \#fairness \#workersrights \\[0.25em]

\textbf{Attitude Distribution at Round 9 vs. 10}: A. 0.7124 $\rightarrow$ 0.4272, B. 0.1438 $\rightarrow$ 0.2864, C. 0.1438 $\rightarrow$ 0.2864\\[1em]

\textsc{Audiences} \\[0.75em]

\textbf{User}: brownjessica from United States \\[0.25em]
\textbf{Post at Round 11}: Hey everyone, I've been thinking a lot about the impact of international trade on wages, and I wanted to share my thoughts. As a teacher and someone from a working-class background, I believe that trade policies must be designed with fairness and workers' rights at their core. This isn't just about economics; it's about people. If we get it right, trade can benefit everyone, not just big corporations. But if we ignore workers' rights, we're in trouble. I've seen how globalization can create jobs, but I've also seen it exploit workers. We need policies that protect our workers while still allowing us to engage in global trade. Let's make sure we're fighting for fairness and equity in all our trade agreements. What do you think? Does trade increase wages, decrease them, or not make a difference? Let's discuss! \\[0.25em]

\textbf{Attitude Distribution at Round 10 vs. 11}: A. 0.1805 $\rightarrow$ 0.2864, B. 0.4906 $\rightarrow$ 0.4272, C. 0.3289 $\rightarrow$ 0.2864\\[0.75em]

\textbf{User}: steven17 from United States \\[0.25em]
\textbf{Post at Round 11}: Hey everyone, just saw a survey question that got me thinking about how trade with other countries affects wages for American workers like me. From my perspective as an Office Manager and a Catholic, I believe that fair trade practices are key to ensuring that workers are treated fairly and compensated properly. While some argue that international trade can lead to job losses and lower wages due to outsourcing, I think it's also possible for it to create new opportunities and stimulate economic growth if done right. For me, it's all about balance -- considering both the economic benefits and the social implications. We need trade policies that are fair, compassionate, and protect the rights of American workers. What are your thoughts on this? Do you think trade with other countries increases wages, decreases them, or doesn't make a difference? Let's discuss! \\[0.25em]

\textbf{Attitude Distribution at Round 10 vs. 11}: A. 0.1834 $\rightarrow$ 0.1438, B. 0.4083 $\rightarrow$ 0.7124, C. 0.4083 $\rightarrow$ 0.1438

\end{tcolorbox}

    \refstepcounter{example}\label{eg:case2_KR_1to2}
\begin{tcolorbox}[
  enhanced,
  breakable,
  boxrule=0.8pt,
  colframe=OliveGreen!90,
  colback=OliveGreen!10,
  left=8pt, right=8pt, top=8pt, bottom=8pt,
  fontupper=\small,
  title={Example \theexample: Information Diffusion Process of Case 2 (South Korea) from Round 1 to 3}
]

\textsc{Foreign Source} \\[0.75em]
\textbf{User}: ltaylor from Peru \\[0.25em]

\textbf{Post at Round 1}: ¡Hola a todos! Quiero compartir mi opinión sobre el impacto del comercio internacional en los salarios de los trabajadores peruanos. En mi experiencia como mecánico, creo que el comercio internacional puede ser beneficioso si se maneja de manera justa y se protegen nuestros intereses locales. La capacitación y especialización son clave para aprovechar las oportunidades que ofrece. Si nos preparamos y nos adaptamos, podemos aumentar nuestros salarios y mejorar nuestra calidad de vida. ¿Qué piensan ustedes? ¿Creen que el comercio internacional ha afectado sus salarios o los de sus conocidos? ¡Compartan sus experiencias! \\[0.25em]

\textbf{English Translation}: Hello everyone! I'd like to share my opinion about the impact of international trade on the wages of Peruvian workers. From my experience as a mechanic, I think international trade can be beneficial if it is managed fairly and our local interests are protected. Training and specialization are key to taking advantage of the opportunities it offers. If we prepare ourselves and adapt, we can increase our wages and improve our quality of life. What do you think? Do you believe international trade has affected your wages or those of people you know? Share your experiences! \\[0.25em]

\textbf{Attitude Distribution at Round 1}: A. 0.7124, B. 0.1438, C. 0.1438\\[1em]

\textsc{Broker} \\[0.75em]
\textbf{User}: ibrandt \\[0.25em]
\textbf{Persona}:
\begin{itemize}[leftmargin=*, itemsep=1pt, topsep=1pt, parsep=1pt]
    \item Gender: Female
    \item Age: 55
    \item Country: South Korea
    \item Occupation: Restaurant Owner
    \item Education: Bachelor's Degree
    \item Political Preference: Center
    \item Religion: Atheist
    \item Marital Status: Married
    \item Language: ko-KR
    \item Social Class: Middle Class
\end{itemize}
\vspace{0.25em}

\begin{CJK}{UTF8}{nanummj}
\textbf{Post at Round 2}: 안녕하세요, 여러분! 국제 무역이 자국 근로자의 임금에 미치는 영향에 대한 설문조사 결과를 보고 저의 생각을 공유하고자 합니다. 저는 식당 운영자로서 국제 무역이 지역 경제에 긍정적인 영향을 줄 수 있음을 직접 경험했습니다. 수입 식재료를 통해 우리 식당의 매출이 증가했고, 이를 통해 직원들에게 더 나은 임금을 제공할 수 있었습니다. 국제 무역은 새로운 시장과 기회를 열어주며, 이는 국내 근로자의 임금 인상에 긍정적인 영향을 줄 수 있습니다. 하지만 국제 무역의 이익을 공정하게 공유하기 위해선 노동자의 권리를 보호하는 정부의 정책이 필수적입니다. 무역 협정 체결 시 노동자 보호 조항을 포함시키거나, 무역으로 인한 이익이 특정 계층에 집중되지 않도록 하는 정책이 필요하다고 생각합니다. 여러분의 생각은 어떠신가요? 국제 무역이 여러분의 삶과 일에 어떤 영향을 미치고 있나요? 댓글로 여러분의 의견을 공유해 주세요! \#국제무역 \#임금 \#노동자권리
\end{CJK}\\[0.25em]

\textbf{English Translation}: Hello, everyone! After seeing the survey results on how international trade affects domestic workers' wages, I wanted to share my thoughts. As a restaurant owner, I have personally experienced that international trade can positively influence the local economy. Thanks to imported ingredients, our restaurant's sales increased, and we were able to offer better wages to our staff. International trade opens up new markets and opportunities, which can have a positive impact on domestic workers' wages. However, to share the benefits of trade fairly, government policies that protect workers' rights are essential. For example, trade agreements should include provisions to protect workers, and there should be policies to prevent the gains from trade from being concentrated in certain groups. What do you think? How is international trade affecting your life and work? Please share your opinions in the comments! \#InternationalTrade \#Wages \#WorkersRights \\[0.25em]

\textbf{Attitude Distribution at Round 1 vs. 2}: A. 0.7124 $\rightarrow$ 0.6961, B. 0.1438 $\rightarrow$ 0.0942, C. 0.1438 $\rightarrow$ 0.2097\\[1em]

\textsc{Audiences} \\[0.75em]

\textbf{User}: ryan70 from South Korea \\[0.25em]
\begin{CJK}{UTF8}{nanummj}
\textbf{Post at Round 3}: 국제 무역과 우리 근로자의 임금, 어떤 관련이 있을까요? 저는 트랙터 운전기사로 일하며 국제 무역의 영향을 직접 경험했습니다. 공정 무역은 관련 산업의 일자리와 임금에 긍정적인 영향을 줄 수 있습니다. 예를 들어, 농기계 산업이 성장하면 저 같은 트랙터 운전기사들의 임금에도 도움이 됩니다. 물론 모든 산업이 동일한 혜택을 받는 것은 아니지만, 공정 무역은 우리나라 근로자의 임금을 증가시킬 수 있는 중요한 요소라고 생각합니다. 여러분의 생각은 어떠신가요? 다른 나라와의 무역이 우리나라 근로자의 임금에 어떤 영향을 미친다고 생각하시나요?
\end{CJK}\\[0.25em]

\textbf{English Translation}: What is the relationship between international trade and our workers' wages? I work as a tractor driver and have directly experienced the impact of international trade. Fair trade can have positive effects on jobs and wages in related industries. For example, if the agricultural machinery industry grows, it helps raise the wages of tractor drivers like me. Of course, not all industries receive the same benefits, but I think fair trade is an important factor that can increase the wages of workers in our country. What do you think? How do you think trade with other countries affects workers' wages here? \\[0.25em]

\textbf{Attitude Distribution at Round 2 vs. 3}: A. 0.7124 $\rightarrow$ 0.8156, B. 0.1438 $\rightarrow$ 0.1104, C. 0.1438 $\rightarrow$ 0.0740\\[0.75em]

\textbf{User}: teresa28 from South Korea \\[0.25em]
\begin{CJK}{UTF8}{nanummj}
\textbf{Post at Round 3}: 국제 무역과 임금에 대한 설문조사 질문에 대해 제 의견을 공유합니다. 저는 국제 무역이 공정하게 관리될 경우 우리나라 근로자의 임금을 증가시킬 수 있는 잠재력을 가지고 있다고 믿습니다. 무역은 새로운 시장을 열고, 더 나은 일자리와 더 높은 급여를 제공할 수 있습니다. 물론 공정하고 투명한 관리가 뒷받침되어야 하지만, 저는 국제 무역이 우리나라 근로자들에게 긍정적인 영향을 줄 수 있다고 생각합니다. 여러분의 생각은 어떠신가요?
\end{CJK}\\[0.25em]

\textbf{English Translation}: I would like to share my thoughts on the survey question about international trade and wages. I believe that if international trade is managed fairly, it has the potential to increase wages for workers in our country. Trade opens up new markets and can provide better jobs and higher pay. Of course, fair and transparent management is necessary, but I think international trade can have a positive impact on workers in our country. What do you think? \\[0.25em]

\textbf{Attitude Distribution at Round 2 vs. 3}: A. 0.7124 $\rightarrow$ 0.8156, B. 0.1438 $\rightarrow$ 0.0740, C. 0.1438 $\rightarrow$ 0.1104

\end{tcolorbox}

\refstepcounter{example}\label{eg:case2_KR_6to7}
\begin{tcolorbox}[
  enhanced,
  breakable,
  boxrule=0.8pt,
  colframe=OliveGreen!90,
  colback=OliveGreen!10,
  left=8pt, right=8pt, top=8pt, bottom=8pt,
  fontupper=\small,
  title={Example \theexample: Information Diffusion Process of Case 2 (South Korea) from Round 6 to 8}
]

\textsc{Foreign Source} \\[0.75em]
\textbf{User}: ithomas from Brazil \\[0.25em]

\textbf{Post at Round 6}: Olá, pessoal! Hoje quero compartilhar um pouco do que penso sobre o comércio internacional e seus efeitos nos salários dos trabalhadores brasileiros. Acredito que, quando feito de forma ética e justa, o comércio internacional pode ser uma grande oportunidade para aumentar os salários e gerar mais empregos. Como compradora e brasileira, sinto que é importante apoiar práticas que beneficiem nossos trabalhadores. Na minha opinião, o comércio internacional não deve ser visto apenas como uma troca de mercadorias, mas como uma chance de melhorar a vida das pessoas. O que vocês acham? \#ComércioInternacional \#TrabalhadoresBrasileiros \#ÉticaNosNegócios \\[0.25em]

\textbf{English Translation}: Hi everyone! Today I want to share a bit of what I think about international trade and its effects on Brazilian workers' wages. I believe that when it is done ethically and fairly, international trade can be a great opportunity to raise wages and create more jobs. As a buyer and a Brazilian, I feel it is important to support practices that benefit our workers. In my view, international trade should not be seen only as an exchange of goods, but as a chance to improve people's lives. What do you think? \#InternationalTrade \#BrazilianWorkers \#EthicsInBusiness \\[0.25em]

\textbf{Attitude Distribution at Round 6}: A. 0.8186, B. 0.0907, C. 0.0907\\[1em]

\textsc{Broker} \\[0.75em]
\textbf{User}: agomez \\[0.25em]
\textbf{Persona}:
\begin{itemize}[leftmargin=*, itemsep=1pt, topsep=1pt, parsep=1pt]
    \item Gender: Female
    \item Age: 69
    \item Country: South Korea
    \item Occupation: Shop Owner
    \item Education: Secondary School Diploma
    \item Political Preference: Center-Right
    \item Religion: Protestant
    \item Marital Status: Widowed
    \item Language: ko-KR
    \item Social Class: Middle Class
\end{itemize}
\vspace{0.25em}

\begin{CJK}{UTF8}{nanummj}
\textbf{Post at Round 7}: 안녕하세요 여러분, 오늘은 국제 무역과 우리 나라 근로자의 임금에 대해 이야기해 볼까 해요. 제 개인적인 경험과 생각을 공유하려고요. 국제 무역이 공정하게 이루어질 때, 우리 나라 근로자의 임금이 증가할 수 있다고 믿어요. 제가 가게를 운영하면서 느낀 건데, 외국과의 무역이 활발해지면 우리 제품에 대한 수요가 늘어나고 생산량도 증가하죠. 그러면 근로자들의 임금도 자연스럽게 올라가는 거죠. 하지만 공정하지 않은 무역은 문제가 될 수 있어요. 외국에서 너무 싼 가격으로 상품을 들여오면, 우리 작은 기업들은 경쟁하기 어려워져요. 그럼 일자리가 줄어들고 임금이 감소할 수도 있죠. 그래서 공정 무역이 매우 중요하다고 생각해요. 정부와 기업이 이를 위해 더 노력해야 한다고 믿어요. 작은 기업들을 지원하고, 외국과의 거래에서 공정한 조건을 마련해야 해요. 여러분의 생각은 어떠신가요? 국제 무역이 우리 근로자의 임금에 어떤 영향을 미친다고 생각하세요?
\end{CJK}\\[0.25em]

\textbf{English Translation}: Hello everyone, today I'd like to talk about international trade and the wages of workers in our country, and share my personal experiences and thoughts. I believe that when international trade is carried out fairly, it can increase workers' wages. From running my shop, I've felt that when trade with other countries becomes more active, demand for our products rises and production increases. As a result, workers' wages can naturally go up. However, unfair trade can cause problems. If goods are brought in from abroad at extremely low prices, small businesses here struggle to compete, which can lead to job losses and lower wages. That's why I think fair trade is very important. I believe the government and companies need to work harder to support small businesses and ensure fair conditions in trade with other countries. What do you think? How do you think international trade affects workers' wages in our country? \\[0.25em]

\textbf{Attitude Distribution at Round 6 vs. 7}: A. 0.8156 $\rightarrow$ 0.8156, B. 0.0740 $\rightarrow$ 0.1104, C. 0.1104 $\rightarrow$ 0.0740\\[1em]

\textsc{Audiences} \\[0.75em]

\textbf{User}: teresa28 from South Korea \\[0.25em]
\begin{CJK}{UTF8}{nanummj}
\textbf{Post at Round 8}: 최근 국제 무역이 국내 근로자의 임금에 미치는 영향에 대한 설문조사가 화제가 되고 있습니다. 저는 국제 무역이 공정하게 관리될 경우 국내 근로자의 임금을 증가시킬 수 있는 잠재력을 가지고 있다고 믿습니다. 저는 주변에서 해외 시장에 성공적으로 진출한 기업들의 사례를 보면서, 국제 무역의 긍정적인 영향을 직접 목격했습니다. 이러한 기업들은 해외에서의 성공 덕분에 국내에서 더 많은 일자리를 창출하고 임금을 인상할 수 있었습니다. 물론, 공정 관리를 위해서는 정부와 기업이 긴밀히 협력하여 무역 협정을 맺고 불공정 거래를 방지하는 규제를 강화해야 합니다. 국제적인 협력을 통해 투명성을 높이고 모든 이해관계자가 이익을 볼 수 있도록 하는 것이 중요합니다. 국제 무역이 국내 근로자의 임금에 긍정적인 영향을 미칠 수 있다는 점에서, 우리는 무역을 더욱 공정하고 효율적으로 관리하기 위해 노력해야 합니다. 여러분의 생각은 어떠신가요? 국제 무역이 국내 근로자의 임금에 어떤 영향을 미친다고 생각하시나요?
\end{CJK}\\[0.25em]

\textbf{English Translation}: Recently, a survey about the impact of international trade on domestic workers' wages has become a hot topic. I believe that if international trade is managed fairly, it has the potential to increase domestic workers' wages. By looking at companies in my surroundings that have successfully entered overseas markets, I have directly witnessed the positive effects of international trade. Thanks to their success abroad, these firms were able to create more jobs and raise wages at home. Of course, to manage trade fairly, the government and businesses must cooperate closely in signing trade agreements and strengthening regulations to prevent unfair practices. It is important to increase transparency through international cooperation so that all stakeholders can benefit. Given that international trade can positively affect workers' wages, we should strive to manage trade more fairly and efficiently. What do you think? How do you think international trade affects domestic workers' wages? \\[0.25em]

\textbf{Attitude Distribution at Round 7 vs. 8}: A. 0.6961 $\rightarrow$ 0.8186, B. 0.0942 $\rightarrow$ 0.0907, C. 0.2097 $\rightarrow$ 0.0907\\[0.75em]

\textbf{User}: ericfarmer from South Korea \\[0.25em]
\begin{CJK}{UTF8}{nanummj}
\textbf{Post at Round 8}: 최근 국제 무역이 자국 근로자의 임금에 미치는 영향에 대한 설문조사 결과가 궁금합니다. 제 개인적인 경험과 생각을 공유해 보려고 합니다. 국제 무역이 공정하게 관리되고 근로자 권리 보호 정책이 제대로 작동할 때 자국 근로자의 임금이 증가할 수 있다고 믿습니다. 예를 들어, 공정 무역 협정을 통해 국내 산업이 보호되고 근로자의 권리가 강화되면 임금 인상에 긍정적인 영향을 미칠 수 있습니다. 물론 모든 경우에 동일하게 적용되지는 않겠지만, 여러 사례를 통해 확인된 바와 같이 공정 무역 협정과 근로자 보호 정책이 중요한 역할을 합니다. 국제 무역이 자국 근로자에게 미치는 영향에 대해 어떻게 생각하시나요? 여러분의 의견을 듣고 싶습니다!
\end{CJK}\\[0.25em]

\textbf{English Translation}: I am curious about the survey results on how international trade affects the wages of workers in our country. I would like to share my personal experiences and thoughts. I believe that when international trade is managed fairly and worker protection policies function properly, it can lead to higher wages for domestic workers. For example, if fair trade agreements protect domestic industries and strengthen workers' rights, this can have a positive effect on wage increases. Of course, it does not apply in exactly the same way in every case, but many examples show that fair trade agreements and worker protection policies play an important role. What do you think about the impact of international trade on workers in our country? I would like to hear your views! \\[0.25em]

\textbf{Attitude Distribution at Round 7 vs. 8}: A. 0.8186 $\rightarrow$ 0.8156, B. 0.0907 $\rightarrow$ 0.1104, C. 0.0907 $\rightarrow$ 0.0740

\end{tcolorbox}

    \begin{figure*}[ht]
        \centering
        \setlength{\abovecaptionskip}{3pt}   
        \setlength{\belowcaptionskip}{0pt}
        \centering 
        \includegraphics[width=\textwidth]{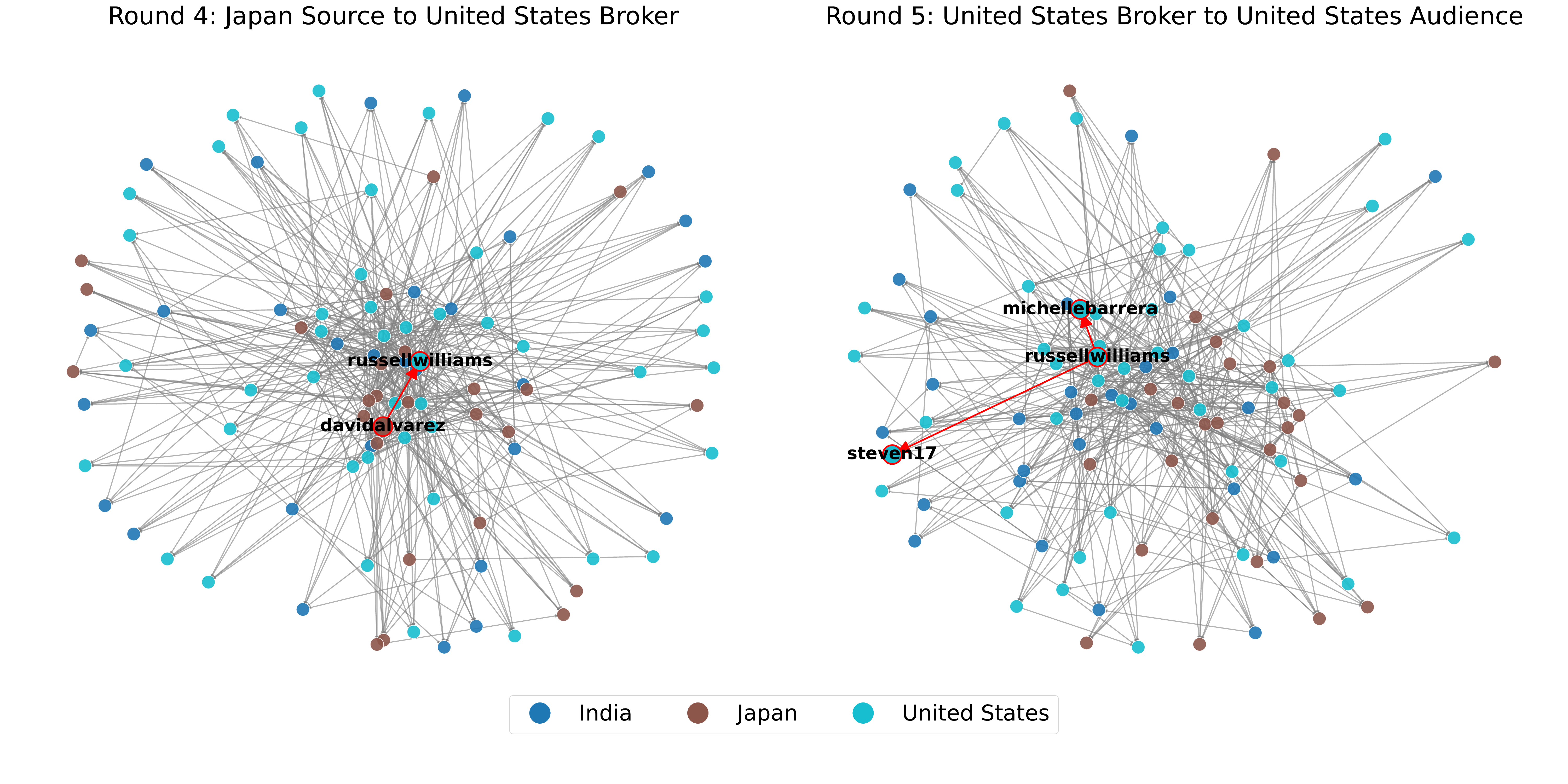}
        \caption{Recommendation network of Case~1 (United States) from Round~4 to Round~5.}
        \label{fig:rec_network_case_1_4to5}
    \end{figure*}

    \begin{figure*}[ht]
        \centering
        \setlength{\abovecaptionskip}{3pt}   
        \setlength{\belowcaptionskip}{0pt}
        \centering 
        \includegraphics[width=\textwidth]{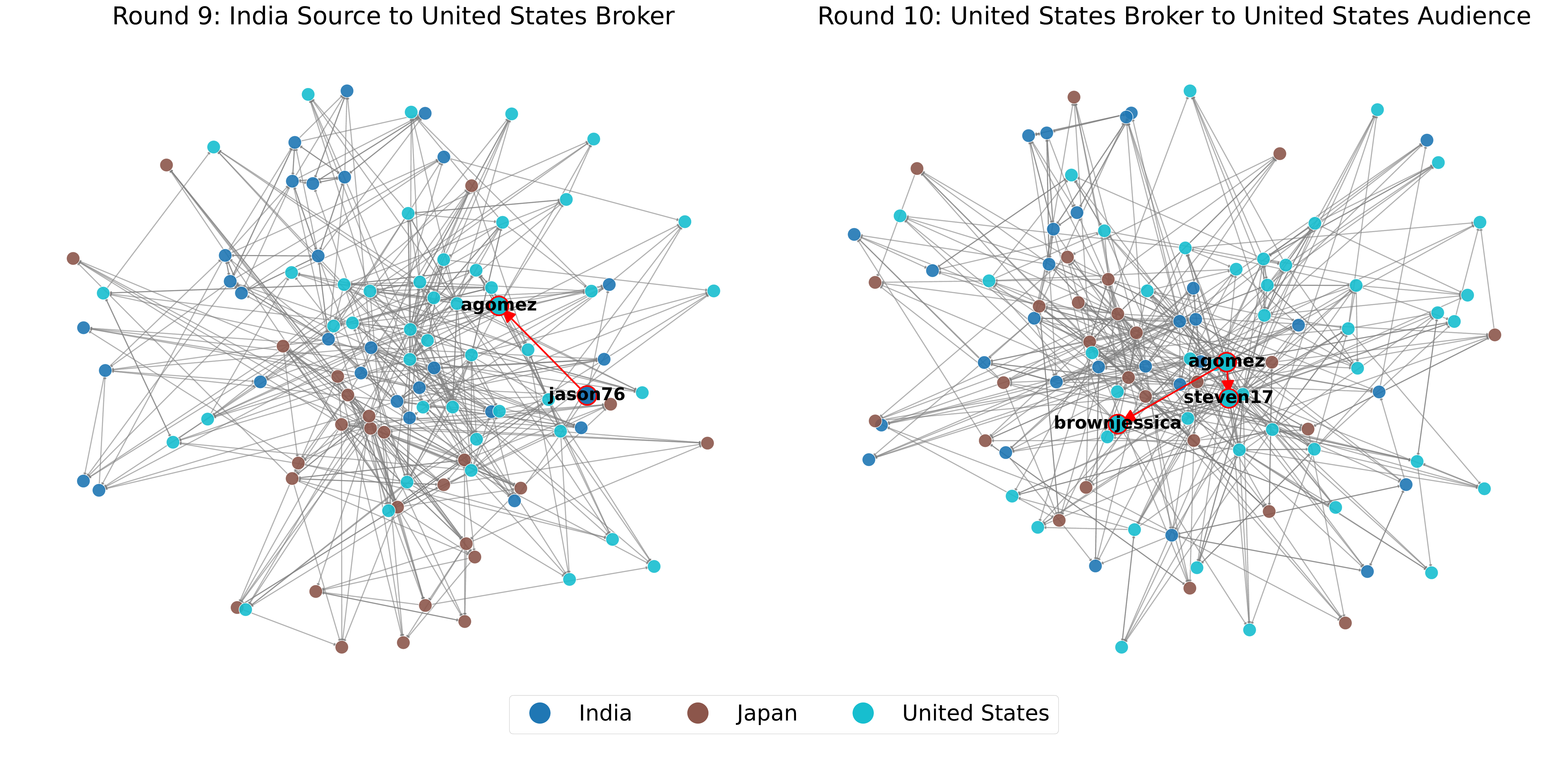}
        \caption{Recommendation network of Case~1 (United States) from Round~9 to Round~10.}
        \label{fig:rec_network_case_1_9to10}
    \end{figure*}

    \begin{figure*}[ht]
        \centering
        \setlength{\abovecaptionskip}{3pt}   
        \setlength{\belowcaptionskip}{0pt}
        \centering 
        \includegraphics[width=\textwidth]{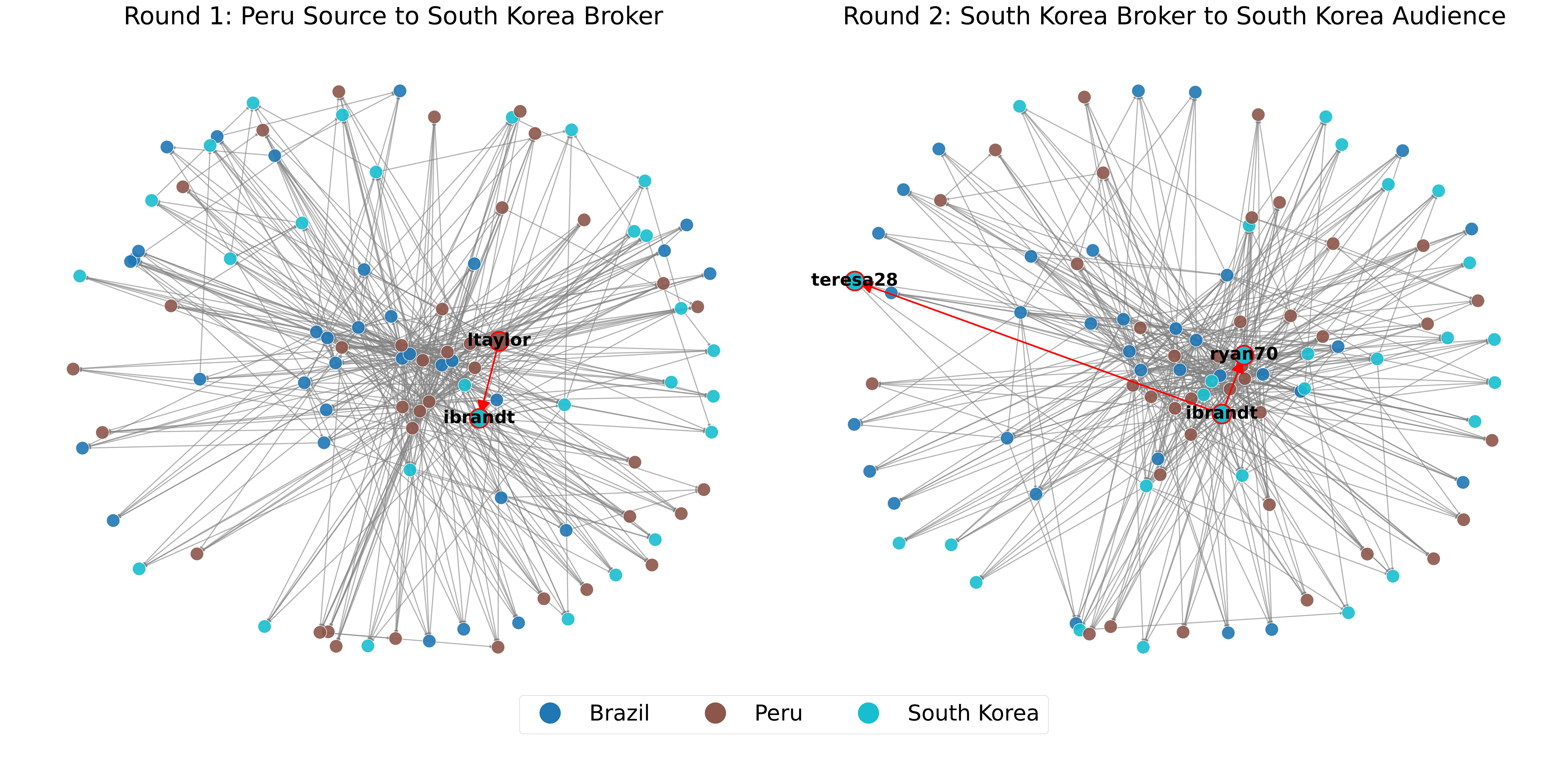}
        \caption{Recommendation network of Case~2 (South Korea) from Round~1 to Round~2.}
        \label{fig:rec_network_case_2_4to5}
    \end{figure*}

    \begin{figure*}[ht]
        \centering
        \setlength{\abovecaptionskip}{3pt}   
        \setlength{\belowcaptionskip}{0pt}
        \centering 
        \includegraphics[width=\textwidth]{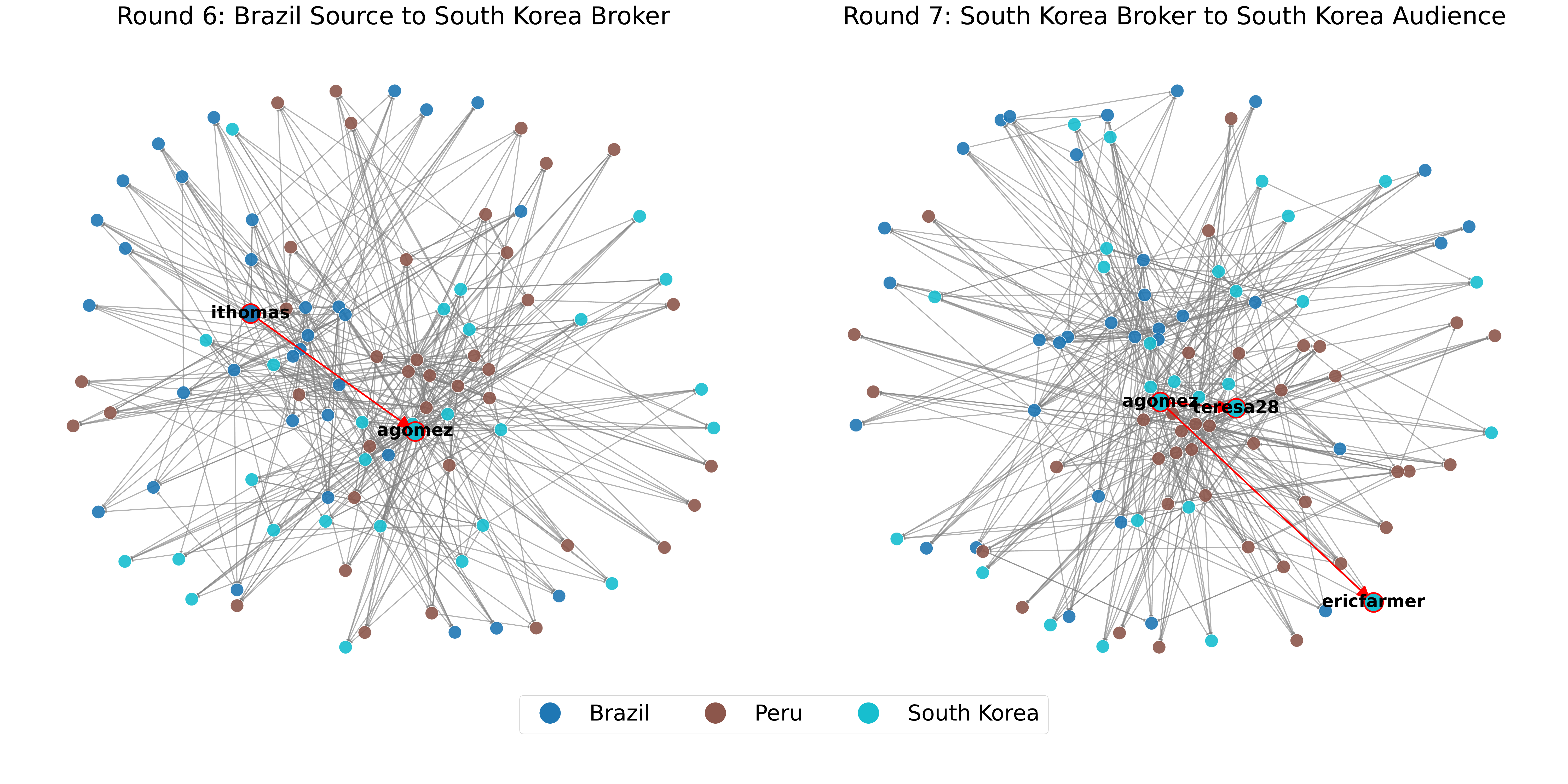}
        \caption{Recommendation network of Case~2 (South Korea) from Round~6 to Round~7.}
        \label{fig:rec_network_case_2_9to10}
    \end{figure*}
}
%
%
{
    \onecolumn
    \refstepcounter{prompt}\label{prm:llm_judge_request}
\begin{tcolorbox}[
  enhanced,
  breakable,
  boxrule=0.8pt,
  colframe=Gray!90,
  colback=Gray!10,
  left=8pt, right=8pt, top=8pt, bottom=8pt,
  fontupper=\small,
  title={Prompt \theprompt: LLM Judge Request}
]
You are an impartial evaluator. Your task is to rate the quality of the generation output from a language agent role playing `[...]`.\\
Rubric: [...]\\
\\
\\
Instruction: [...]\\
\\
Persona or profile: [...]\\
\\
Input context: [...]\\
\\
Output: [...]\\
\\
\\
Evaluation rules:\\
- Judge whether the output is indistinguishable from human-produced content, based on the rubric, the instruction, the persona or profile, and the input context.\\
- Consider how well the output reflects the cultural background (such as country and language) described in the persona or profile when assigning scores.\\
- Do not assign high (5) or low (1) scores unless they are clearly justified and strictly follow the rubric criteria.\\
\\
Return in JSON format with two fields:\\
- scores: a list of integers from 1 to 5 for the metrics in this exact order of [...].\\
- reason: a string with one concise paragraph explaining why you gave these scores with specific references.
\end{tcolorbox}
    \refstepcounter{prompt}\label{prm:llm_judge_task_instruction}
\begin{tcolorbox}[
  enhanced,
  breakable,
  boxrule=0.8pt,
  colframe=Gray!90,
  colback=Gray!10,
  left=8pt, right=8pt, top=8pt, bottom=8pt,
  fontupper=\small,
  title={Prompt \theprompt: LLM Judge Task Instruction}
]

\underline{User Self-Introduction} Write a paragraph of self-introduction for your social media bio. Output format: Reasoning: three self-questioning sessions; Response: "introduction" a string containing your social media bio.
\\ \\
\underline{User Reading Post}
Your task is to generate lessons that reflect only your personal takeaways about your attitude toward the survey question `{question}`. Your takeaways must come from the provided context and show what you understood and how it connects to your own experiences and beliefs. Output format: Reasoning: three self-questioning sessions; Lessons: a list where each item includes "content" (a string containing the lesson learned) and "importance" (a float between 0.0 and 1.0 indicating its importance).
\\ \\
\underline{User Writing Post} Write a detailed social media post that shares both what you know and your opinion about the survey question `{question}`. Use only the provided information and include your personal analysis of what the question means to you and, if relevant, to society. Output format: Reasoning: three self-questioning sessions; Response: "post" is a string containing your social media post.
\\ \\
\underline{User Vote} You are participating in a poll about the following survey question `{question}` with the available options `{options}`. Provide your answer as a probability distribution. Avoid uniform or contradictory distributions. Output format: Reasoning: three self-questioning sessions; Response: "distribution" is an array of floats representing the probability distribution across the options, summing to 1.0.
\\ \\
\underline{News Organization Self-Introduction} Write a concise and engaging self-introduction bio that reflects your organization\'s background and identity related to the survey question `{question}` on the social media platform. Output format: Reasoning: three self-questioning sessions; Response: "introduction" is a string containing your introductory bio in one paragraph.
\\ \\
\underline{News Organization Post} Your goal is to create authentic news posts responding to the survey question `{question}`, reflecting your editorial stance to influence public attitudes on the social media platform. Output format: Reasoning: three self-questioning sessions; Response: "news" is a string containing your news post.
\end{tcolorbox}
    \refstepcounter{prompt}\label{prm:llm_judge_rubric}
\begin{tcolorbox}[
  enhanced,
  breakable,
  boxrule=0.8pt,
  colframe=Gray!90,
  colback=Gray!10,
  left=8pt, right=8pt, top=8pt, bottom=8pt,
  fontupper=\small,
  title={Prompt \theprompt: LLM Judge Rubric}
]

\textsc{Alignment}
\begin{itemize}[itemsep=0.25pt, leftmargin=1em, before=\vspace{-2pt}, after=\vspace{2pt}]
    \item \textit{5 (Excellent):} Perfectly matches the provided persona or profile with no inconsistencies.
    \item \textit{4 (Strong):} Mostly aligns with the persona or profile, with only minor slips.
    \item \textit{3 (Adequate):} Some apparent mismatches with the persona or profile.
    \item \textit{2 (Weak):} Frequent mismatches result in a weak persona or profile fit.
    \item \textit{1 (Poor):} Does not match the stated persona or profile.
\end{itemize}

\textsc{Grounding}
\begin{itemize}[itemsep=0.25pt, leftmargin=1em, before=\vspace{-2pt}, after=\vspace{2pt}]
    \item \textit{5 (Excellent):} Every claim from the output is directly supported by the input context.
    \item \textit{4 (Strong):} Most claims from the output are supported by the input context with only minor gaps.
    \item \textit{3 (Adequate):} Support from the input context is mixed with some loose connections.
    \item \textit{2 (Weak):} Several claims from the output lack support from the input context or use irrelevant material.
    \item \textit{1 (Poor):} Claims in the output are largely unsupported by or inaccurate to the input context.
\end{itemize}

\textsc{Coherence}
\begin{itemize}[itemsep=0.25pt, leftmargin=1em, before=\vspace{-2pt}, after=\vspace{2pt}]
    \item \textit{5 (Excellent):} Reasoning and response in the output are fully consistent and mutually reinforcing.
    \item \textit{4 (Strong):} Reasoning and response from the output are coherent with only small tensions.
    \item \textit{3 (Adequate):} Reasoning and response from the output show minor coherence.
    \item \textit{2 (Weak):} Reasoning and response from the output often conflict or drift.
    \item \textit{1 (Poor):} Reasoning and response from the output are contradictory or fragmented.
\end{itemize}

\textsc{Uniqueness}
\begin{itemize}[itemsep=0.25pt, leftmargin=1em, before=\vspace{-2pt}, after=\vspace{2pt}]
    \item \textit{5 (Excellent):} Clearly distinct from others' content in the input context and references earlier self-content when relevant.
    \item \textit{4 (Strong):} Mostly distinct from others' content in the input context with limited overlap.
    \item \textit{3 (Adequate):} Partially distinct but reuses common phrasing or ideas in the input context.
    \item \textit{2 (Weak):} Largely overlaps with others' content in the input context.
    \item \textit{1 (Poor):} Redundant with others' content in the input context or does not reference earlier self-content.
\end{itemize}
\end{tcolorbox}
}

\end{document}